\documentclass[11pt]{article}
\usepackage{amsmath}
\usepackage{comment}
\usepackage{graphicx,psfrag,epsf,subcaption,multirow}
\usepackage{enumerate}
\usepackage[backend=biber, 
natbib=true, 
style=authoryear-comp, 
maxcitenames=2, 
uniquelist=false, 
giveninits=true, 
maxbibnames=6,  
minbibnames=6, 
sortcites=false, 
uniquename=false 
]{biblatex}
\DeclareNameAlias{author}{family-given} 
\DeclareDelimFormat[parencite]{nameyeardelim}{\addspace} 
\uspunctuation 
\renewbibmacro{in:}{} 
\usepackage{url} 
\newcommand{\blind}{0}

\usepackage{float}
\usepackage{indentfirst}
\usepackage{caption}
\usepackage{doi} 
\usepackage{amsfonts} 
\usepackage{amssymb}
\usepackage{mathtools}
\usepackage{amsthm}
\usepackage{algorithm}
\usepackage{algorithmic}
\theoremstyle{plain}
\newtheorem{theorem}{Theorem}[section]
\newtheorem{corollary}[theorem]{Corollary}
\newtheorem{proposition}[theorem]{Proposition}

\newtheorem{remark}{Remark}[section]
\usepackage{xcolor}
\allowdisplaybreaks
\usepackage{titlesec}
\titlelabel{\thetitle.\quad}
\usepackage{sidecap,changepage} 

\DeclareMathOperator{\E}{\mathbb{E}}

\begin{document}

\def\spacingset#1{\renewcommand{\baselinestretch}%
{#1}\small\normalsize} \spacingset{1}


\if0\blind
{
  \title{\bf 
  Latent Utility Q-Learning for Preference-Adaptive Dynamic Treatment Regimes
  }
  \author{Joshua P. Zitovsky\footnotemark[2]\\
    Department of Biostatistics \\ UNC Chapel Hill \\
    \and 
    Yating Zou\footnotemark[2]\\
    Department of Biostatistics \\ UNC Chapel Hill \\
    \and
    Leslie Wilson\\
    Department of Clinical Pharmacy \\
    University of California, San Francisco \\
    \and 
    Michael R. Kosorok\thanks{Correspondence to: Michael R. Kosorok $<$kosorok@bios.unc.edu$>$} \\
     Department of Biostatistics \\ UNC Chapel Hill}
  \date{}
  \maketitle
  {\renewcommand{\thefootnote}{\fnsymbol{footnote}}%
 \footnotetext[2]{These authors contributed equally and are listed as co-first authors.}}
} \fi

\if1\blind
{
  \title{\bf 
  Latent Utility Q-Learning for Preference-Adaptive Dynamic Treatment Regimes
  }
  \author{}
  \date{}
  \maketitle
  {\renewcommand{\thefootnote}{\fnsymbol{footnote}}%
 \footnotetext[2]{Equal contribution as first author.}}
} \fi

\begin{abstract}
Optimizing individualized treatment sequences for patients who weigh multiple, competing outcomes differently poses a challenge for dynamic treatment regime (DTR) methods, which typically assume a single univariate outcome. We propose Latent Utility Q-Learning (LUQ-Learning), which estimates DTRs optimizing patient-specific preference-weighted combinations of multivariate outcomes $\mathbf{Y}\in\mathbb{R}^d$ across $K$ decision points. A conditional mean factorization decouples preference estimation from outcome regression, enabling flexible, modular learning under imperfectly observed and heterogeneous preferences without requiring explicit outcome ranking by patients. We establish consistency of the estimated value function and derive unified $\epsilon$-optimality guarantees that bound policy value loss in terms of posterior preference uncertainty, yielding interpretable criteria for data-driven policy selection. Simulations calibrated to Sequential Multiple Assignment Randomized Trials (SMARTs) demonstrate that LUQ-Learning outperforms Q-learning with naive outcome aggregation, last-reported satisfaction optimization, and existing preference-based methods.
\end{abstract}

\noindent%
\vfill

\newpage
\spacingset{1} 

\section{Introduction}
\label{sec:introduction}

Many chronic conditions require treatment decisions made sequentially over time, where the goal is not merely to improve a single endpoint but to balance multiple, possibly competing outcomes according to each patient's individual priorities. Consider the Biomarkers for Evaluating Spine Treatments (BEST) trial (ClinicalTrials.gov: NCT05396014), a two-stage SMART targeting chronic low back pain (cLBP) across twelve U.S.\ sites. Although the primary endpoint is the Pain, Enjoyment of Life, and General Activity (PEG) scale; cLBP affects function, mood, and daily activity in ways that differ substantially across patients, and no single fixed summary measure can capture these trade-offs. A patient who prioritizes physical function over pain intensity should, in principle, receive a different treatment sequence than one with the opposite priorities. Standard dynamic treatment regime (DTR) methods \citep{Kosorok2019, Tsiatis2019}, which optimize a single pre-specified scalar outcome, offer no principled mechanism for incorporating such heterogeneous preferences.

This gap motivates development of a framework for learning \textit{preference-adaptive} DTRs, where the policy is optimized for patient-specific weighted combinations of multiple outcomes. Incorporating more than a fixed scalar reward relates closely to the reward engineering problem in reinforcement learning (RL): how to define a reward function that effectively targets the desired scientific objective. Related work includes minimax optimization over outcome combinations \citep{Jiang2021}, distributional RL \citep{zhang_distributional_2021, lee_off-policy_2024}, and inverse RL learning from expert trajectories \citep{hejna_inverse_2023, hassani_towards_2024}. However, deep RL methods typically require large sample sizes and are difficult to apply in moderate-sample clinical studies. Inverse RL is also incompatible with SMARTs, where treatments are randomized, and no single expert policy governs observed treatment assignments.

The incorporation of subject outcome preferences into DTR estimation was introduced by \citet{Butler2018}, who used item response theory to link questionnaire responses to latent preference parameters and derived optimal single-stage treatment rules that maximize patient-specific convex combinations of competing outcomes. \citet{Zhong2021} extended this framework to two-stage settings with right-censored survival outcomes through SAPP-Q-Learning. From a trial design perspective, \citet{Wank2024PRPP} proposed PRPP-SMARTs, which combine SMARTs with partially randomized patient preference designs. Existing preference-informed DTR methods, however, remain largely limited to two-stage settings, rely on relatively restrictive parametric utility and outcome models, and offer limited treatment of uncertainty in preference estimation. 

Outcome preference information need not come from a single instrument or questionnaire. Useful signals include direct preference elicitation questions in discrete choice experiments \citep{Johnson2013, Janssen2017ImprovingTQ}, conjoint analysis \citep{Bridges2011, Leeper2019MeasuringSP, Liu2023MultipleHT}, satisfaction ratings, and quality-of-life subscales. Many such instruments can be interpreted through discrete choice models that recover latent utilities driving respondent choices \citep{McFadden1974, Hauber2016}, providing a natural probabilistic link between observable responses and unobserved preferences. This stands in contrast to preference-based reinforcement learning (PbRL) 
\citep{wirth_survey_2017, advances_in_pbrl_2022}, where preference is encoded as an ordering over experienced trajectories and used to learn a population-level reward function. Despite success in games and robotics where trajectory generation is cheap \citep{Christiano2017, Ibarz2018}, PbRL is incompatible with clinical trial 
settings as each patient experiences only one treatment trajectory, making trajectory comparisons unavailable and the resulting reward function incapable of capturing individual-level preferences.

We propose \textit{Latent Utility Q-Learning (LUQ-Learning)}, a latent-variable extension of Q-learning that estimates DTRs optimizing preference-weighted combinations of $\mathbf{Y} \in \mathbb{R}^d$ over $K < \infty$ decision points. 
We assume patients' preferences are not directly observed but manifested through indirect measurements.
The central methodological insight is a conditional mean factorization that, under a mild sequential preference independence condition, decouples preference estimation from outcome regression. This reduces multi-outcome, preference-adaptive regime learning to a sequence of scalar Q-learning problems while preserving individualized trade-offs. LUQ-Learning accommodates heterogeneous preference data formats without requiring scale alignment across instruments, avoids explicit action ranking by patients, and propagates preference uncertainty to both treatment recommendations and value estimation through a Bayesian framework.

Our theoretical contributions are twofold. We establish  consistency of the estimated value function under weak regularity conditions on the preference and outcome estimators. We introduce two uncertainty-aware policy variants (modal and robust policy) and derive a unified $\epsilon$-optimality result that bounds policy value loss in terms of posterior preference uncertainty. These bounds yield interpretable, data-driven criteria for selecting among intervention options in practice, when regulatory or resource constraints restrict the action space.

The remainder of the paper is organized as follows. Section~\ref{sec:setup} introduces the setup and notation. Section~\ref{sec:assumptions} presents the assumptions. Section~\ref{sec:LUQ-Learning} presents the LUQ-Learning algorithm and its theoretical properties. Section~\ref{sec:uncertainty-aware} presents uncertainty-aware variants. Section~\ref{sec:policy-selection} discusses implementation. Section~\ref{sec:application} presents simulations tailored to the BEST trial, and Section~\ref{sec:discussions} concludes. All proofs and additional simulation results are provided in the Supplementary Material.

\section{Notation and Setup}
\label{sec:setup}

\begin{figure}[h]
    \centering
    \includegraphics[width=0.9\textwidth]{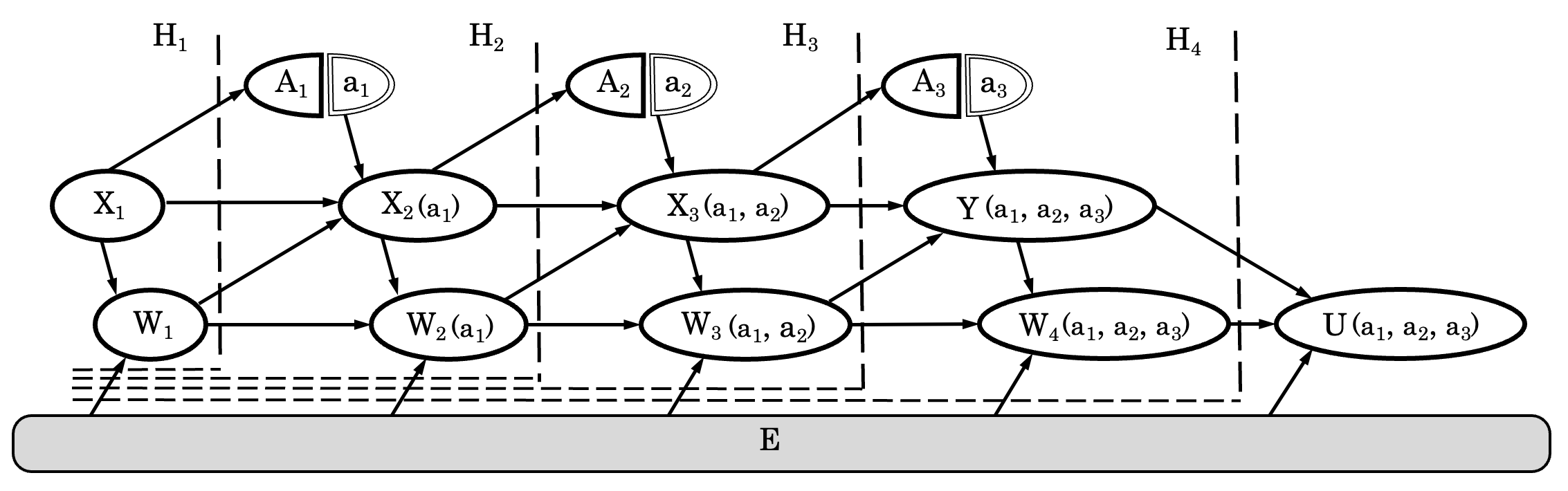}
    \caption{Illustration of a scenario satisfying assumptions related to conditional independence using a Single World Intervention Graph (SWIG) with $K=3$ decision points. $\mathbf{E}$ in gray is the unobserved latent preference. }
    \label{setup_plot}
\end{figure}

We adopt Rubin's potential outcome framework \citep{imbens_causal_2015} to formalize our optimality criterion and adopt the classic notation of $Q$-learning in the precision medicine setting. Consider the problem of sequential decision optimization over $K < \infty$ decision points. We assume that the observed data consist of $n$ i.i.d. trajectories of the form
\[
\mathcal D=\{(\mathbf{Z}_{i,1},\mathbf{Z}_{i,2},\dots,\mathbf{Z}_{i,K}, \mathbf{Y}_i, \mathbf{W}_{i,K+1})\}_{i=1}^n,
\text{ where }
\mathbf{Z}_{i,k} = (\mathbf{X}_{i,k},\mathbf{W}_{i,k},A_{i,k}),
\]
$\mathbf{X}_k \in \mathcal{X}$ are patient covariates which can include baseline or summary statistics of patient status before action $A_k\in{\cal A}_{k}$; and where $\mathbf{Y} \in \mathcal{Y} \subset \mathbb{R}^d$ is a vector of outcomes of interest measured at the end of the study, for example, the primary and secondary endpoints of the study. One of the main differences in our setting is the introduction of $\mathbf{W}_{k} \in \mathcal{W}$, the preference elicitation instrument regarding the final outcome of interest $\mathbf{Y}$. Each $\mathbf{W}_k$ is collected after $A_{k-1}$ but prior to $A_k$, which we expect to reflect individual patient preference. Based on patient trajectories, we define history $\mathbf{H}_k$ for $k = 1, \dots, K$ as random variables taking values in a measurable space $\mathcal{H}_k$ and denote all the information available before action $A_k$ is taken. That is, $\mathbf{H}_1 = (\mathbf{X}_1,\mathbf{W}_{1})$, $\mathbf{H}_2 = (\mathbf{H}_1, A_1, \mathbf{X}_2,\mathbf{W}_{2})$, $\dots$, $\mathbf{H}_{K} = (\mathbf{H}_{K-1}, A_{K-1}, \mathbf{X}_K, \mathbf{W}_{K})$. For completeness, define $\mathbf{H}_0 = \emptyset$, $\mathbf{H}_{K+1} = (\mathbf{H}_{K}, \mathbf{Y}, A_K, \mathbf{W}_{K+1})$. 
For each value $h_k \in \mathcal{H}_k$, let $\mathcal{A}_k(h_k)$ denote the finite set of feasible actions given history $h_k$. This allows the incorporation of restrictions on treatment given patient medical history. Accordingly, define $\mathcal{A}_{k} = \cup_{h_k \in \mathcal{H}_k} \mathcal{A}_{k}(h_k)$. Figure \ref{setup_plot} illustrates this overall structure for a setting where $K = 3$. 

Our goal is to find an optimal sequence of decision rules as a function of history, called a dynamic treatment regime (DTR) ${\pi}^{opt} = (\pi^{opt}_1, \dots, \pi^{opt}_K)$, where $\pi^{opt}_k: \mathcal{H}_k \mapsto \mathcal{A}_{k}$, such that 
\begin{equation}
    V_1^{{\pi}^{opt}}(\mathbf{H}_1) \geq V_1^{{\pi}}( \mathbf{H}_1), \quad 
    \forall {\pi} \in \Pi, \quad \text{almost surely,}
    \label{equation:opt_condition}
\end{equation}
where $V_k^{{\pi}}(\mathbf{H}_k) = \mathbb{E}_{a_{k+1}, \dots, a_{K} \sim \pi}[\mathbf{E}^\top \mathbf{Y}^*({\pi}) | \mathbf{H}_k]$, our preference-incorporated \textit{value function}, and $\Pi$ denoting the class of deterministic policies. We let $\mathbf Y^*({\pi})$ be the $d$-dimension vector of outcomes that would be observed if the subject received the treatment sequence ${\pi} = (\pi_1, \dots, \pi_K)$ with $\pi_k(h_k)\in\mathcal A_k(h_k)$ for $h_k\in\mathcal H_k$, and we let $\mathbf{E}\in\mathcal E$ be the unobserved patient preference, where $\mathcal E$ is a $(d-1)$-dimensional probability simplex. We assume that latent preference $\mathbf E$ affects subjective measures $\mathbf{W}_k$ at all stages, and we allow $\mathbf{W}_k$ to also affect $\mathbf{X}_{k+1}$, the patient status at the next stage. This can happen, for example, when patients may respond better if they are more satisfied with their prior treatments. Note that $\mathbf{X}_k$ and $\mathbf{W}_k$ at $k = 2, \dots, K$ are all observed values of potential outcomes $\mathbf{X}_{k}^*$ and $\mathbf{W}_{k}^*$ as depicted in Figure \ref{setup_plot}. 
Let $\mathbf{E}_k$ denote the random variable following  $P(\mathbf{E} \mid \mathbf{H}_k)$.
Denote $U^*(\bar{a}_K) = \mathbf{E}^\top\mathbf{Y}^*(\bar{a}_K)$ the latent utility, defined as the inner product of the unobserved patient preference $\mathbf{E}$ and the vector of potential outcomes under treatment sequence $\bar{a}_K = (a_1, \dots, a_{K})$. Define accordingly the Q function at stage k as $Q^{\pi^{opt}}_k({\mathbf{H}}_k, A_k) = \mathbb{E}_{a_{k+1}, \dots, a_{K} \sim {\pi^{opt}}}[U^* | {\mathbf{H}}_k, A_k]$, the expected utility if all future actions follow the optimal regime, conditioning on current stage history $\mathbf{H}_k$ and action $A_k$.

\section{Assumptions}
\label{sec:assumptions}

We make the following assumptions throughout: 
\begin{enumerate}
\setlength{\itemsep}{-5pt}
    \item[(A1)] Consistency: If treatment sequence $\Bar{a}_K = (a_1, \dots, a_K)$ is the actual treatment sequence received by subject $i$, then ${\mathbf{Y}^{*}}_i(\Bar{{a}}_{K}) = \mathbf{Y}_i$, and ${\mathbf{X}^*}_{i,k}(\Bar{a}_{k-1}) = {\mathbf{X}}_{i,k}$, ${\mathbf{W}^*}_{i,k}(\Bar{{a}}_{k-1}) = \mathbf{W}_{i,k}$ for all $1 \leq k \leq (K-1)$.
    \item[(A2)] SUTVA (Stable Unit Treatment Value Assumption): One version of treatment. Each value of $a_k \in \mathcal{A}_k$ $\forall 1 \leq k \leq K$ is unambiguously defined.
    \item[(A3)] Positivity: $1 > P(A_k = a_k | {\mathbf{H}}_k = h_k) \geq c$, for some $c > 0$, $\forall a_k \in \mathcal{A}_k, {h}_k \in {\mathcal{H}}_k$ and $1 \leq k \leq K$.
    \item[(A4)] Sequential Ignorability: $\mathbf{X}_{k+1}^{*}(\Bar{a}_{k}) \perp \!\!\! \perp A_k | \mathbf{H}_k$ for all $\Bar{a}_{k}, 1 \leq k \leq (K-1)$, and $\mathbf{Y}^{*}(\Bar{a}_{K}) \perp \!\!\! \perp A_K | \mathbf{H}_K$ for all $\Bar{a}_{K}$.
    \item[(A5)] Sequential Preference Independence: $(\mathbf{X}_{k+1}^{*}(\Bar{a}_{k}), A_k)\perp \!\!\! \perp \mathbf{E}|\mathbf{H}_k$ for all $\Bar{a}_{k}, 1 \leq k \leq (K-1)$, and $(\mathbf{Y}^{*}(\Bar{a}_{K}), A_K)\perp \!\!\! \perp \mathbf{E}|\mathbf{H}_K$ for all $\Bar{a}_{K}$.
\end{enumerate}
(A1)-(A4) are standard assumptions when working with sequences of potential outcomes. (A1) and (A2) relate potential outcomes to the observables and can always be satisfied by properly defining the interventions involved. (A3) ensures that given history $\mathbf{H}_k$ at any decision point, data has sufficient variability in the assigned action for the algorithm to learn the value associated with interventions that are not the observed ones. It can be assessed by referring to the study design and examining $\widehat{P}(A_k | {\mathbf{H}}_k)$ fitted using flexible models. (A4) means that all confounding variables between action $A_k$ and the next state $\mathbf{X}^{*}_{k}$ have been captured in the history $\mathbf{H}_k$. While in an observational study this is unverifiable, in a SMART study this assumption is guaranteed by the conditional randomized treatment assignment structure. (A5) is specific to our Latent Utility Q-Learning algorithm. It is equivalent to $A_k \perp \!\!\! \perp \mathbf{E} | \mathbf{H}_k$ and $\mathbf{X}_{k+1}^{*}(\Bar{a}_{k}) \perp \!\!\! \perp \mathbf{E} | \mathbf{H}_k, A_k$ at all $1 \leq k \leq (K-1)$ and $\mathbf{Y}^{*}(\Bar{a}_{K}) \perp \!\!\! \perp \mathbf{E} | \mathbf{H}_K, A_K$ at the last time point. As shown in Figure \ref{setup_plot}, the key in satisfying this assumption is to ensure that the collected preference information $\mathbf{W}_k$ is rich enough so that 1) it captures all influence of preference $\mathbf{E}$ on patient status at the next stage $\mathbf{X}_{k+1}$ and 2) once we control on the history that includes collected preference, latent preference $\mathbf{E}$ does not have additional influence on the observed treatment assignment mechanism. 

\section{Methodology and Theoretical Guarantees}
\label{sec:method}

\subsection{Latent Utility Q-Learning}
\label{sec:LUQ-Learning}

Under the latent variable modeling framework, we accommodate any preference elicitation mechanism. Assumption (A5) does not invalidate the backward sequential optimization scheme to arrive at an optimal policy, yet is sufficient for disentangling the outcome model $\mathbb{E}[\mathbf{Y} | \mathbf{H}_k , A_k]$ and the preference model $P(\mathbf{E} | \mathbf{H}_k)$ sequentially. Specifically, (A1)-(A5) guarantee that $\mathbb{E}[U^*|\mathbf{H}_{k}, A_k] = \mathbb{E}[\mathbf{E}|\mathbf{H}_{k}]^\top \mathbb{E}[\mathbf{Y}|\mathbf{H}_{k}, A_k]$ for all k. We illustrate the following parametric scheme to estimate the preference model, although we note that this can also be done semi- or non-parametrically since the algorithm remains valid as long as we can sample from the posterior $P(\mathbf{E}|\mathbf{H}_k)$ at all stages. 
The Bayesian framework provides a principled mechanism for propagating uncertainty in the unobserved $\mathbf{E}$ through the posterior $P(\mathbf{E}|\mathbf{H}_k)$, which enters directly into treatment recommendations and the $\varepsilon$-optimality bounds of Section~\ref{sec:uncertainty-aware}. Also, the prior $\Lambda_\theta$ acts as regularization on $\theta$, stabilizing estimation in the small-to-moderate sample settings typical of SMARTs.

Denote $\theta$ as the vector of parameters in the models used to define $P(\mathbf{H}_k|\mathbf{E})$. For any $1 \leq k \leq K$, assumptions (A1)--(A5) allow us to express the recursive equation for the observed likelihood as
\[
P(\mathbf{H}_k | \mathbf{E}) = \prod_{i = 1}^n
P(\mathbf{W}_{i,k} | \mathbf{X}_{i,k}, {\mathbf{H}}_{i,k-1}, \mathbf{E}_{i})
P(\mathbf{X}_{i,k}| {\mathbf{H}}_{i,k-1}, A_{i,k-1})
P(A_{i,k-1} | {\mathbf{H}}_{i,k-1})
P(\mathbf{H}_{i,k-1} | \mathbf{E}_{i}),
\]
and similarly for $P(\mathbf{H}_{K+1} | \mathbf{E})$ but with $\mathbf{Y}$ in the place of $\mathbf{X}$. It is then sufficient to estimate the part regarding $\mathbf{W}$ to sample $\mathbf{E}_k$ from $P(\mathbf{E}|\mathbf{H}_k)$. To do so, we parametrize $P(\mathbf{W}_k|\mathbf{X}_k, \mathbf{H}_{k-1}, \mathbf{E})$ with $\theta_k$ and obtain $\widehat{\theta} = (\widehat{\theta}_1, \dots, \widehat{\theta}_k)$, $1 \leq k \leq K+1$ as the maximizer of the data log posterior, with randomness in $\mathbf{E}$ marginalized out:
\begin{align}
\begin{split}
    \log P(\theta|\mathbf{H}_{k}) 
    &\propto 
    \sum_{i=1}^n \log \Bigg\{\int_{\mathcal E}
    P(\mathbf{W}^{i}_{K+1}| \mathbf{Y}^{i}, \mathbf{H}^{i}_K, \mathbf{E}^{i},\theta_{K+1})^{I(k=(K+1))} \\
     &\times \prod_{l = 1}^{\min(k,K)}
    P(\mathbf{W}^{i}_l| \mathbf{X}^{i}_l, \mathbf{H}^{i}_{l-1}, \mathbf{E}^{i},\theta_l) 
     d \Lambda_{E}(\mathbf{E}^{i}) \Bigg\}\\
    &+ I(k = (K+1)) \log(\Lambda_{\theta}(\theta_{K+1})) +
    \sum_{l=1}^{\min(k,K)} \log(\Lambda_{\theta}(\theta_l)).
     \label{eq:posterior}
\end{split}
\end{align}
The full LUQ-Learning algorithm is summarized in Algorithm~\ref{alg:luq_v1}.

\begin{algorithm}[h]
\spacingset{1.3}
\footnotesize
\caption{LUQ-Learning}
\label{alg:luq_v1}
\begin{algorithmic}[1]
    \STATE Specify a parametric model $P(\mathbf{W}|\mathbf{X}, \mathbf{E}, \theta)$. Specify a prior distribution on $\theta$ and $\mathbf E$, denoted $\Lambda_\theta$ and $\Lambda_E$, respectively. Denote $\gamma$ as the parameter for the outcome model, possibly infinite dimensional.
    \STATE \textbf{Input} Observed trajectories $\mathcal D=\{(\mathbf{Z}_{i,1},\mathbf{Z}_{i,2},\dots,\mathbf{Z}_{i,K}, \mathbf{Y}_i, \mathbf{W}_{i,K+1})\}_{i=1}^N$,  
    $\mathbf{Z}_{i,k} = (\mathbf{X}_{i,k},\mathbf{W}_{i,k},A_{i,k})$
    \FOR{k = K}
        \STATE Obtain $\hat{\theta}_n = argmax_\theta \log P(\theta|\mathbf{H}_{K+1})$ in (\ref{eq:posterior}).
        \STATE Fit outcome model using some regression algorithm to obtain $\hat{\mathbb{E}}[\mathbf{Y} | \mathbf{H}_k, A_k; \hat{\gamma}]$.
        \STATE Obtain 
        \[
        \hat{\pi}^{opt}_{K}(\mathbf{H}_K) = argmax_{a_K \in \mathcal{A}_{K}(\mathbf{H}_K)} \widehat{Q}_K(\mathbf{H}_K, a_K) =  argmax_{a_K \in \mathcal{A}_{K}(\mathbf{H}_K)} \widehat{\mathbb{E}} [\mathbf{E}|\mathbf{H}_K; \hat{\theta}_n]^\top \hat{\mathbb{E}}[\mathbf{Y} | \mathbf{H}_K, a_K; \hat{\gamma}],
        \]
        where 
        $\widehat{\mathbb{E}} [\mathbf{E}|\mathbf{H}_k; \hat{\theta}_n]$ is calculated using integration methods such as Monte-Carlo integration.
        \STATE Let $\widehat{V}_K^{\widehat{\pi}^{opt}}(\mathbf{H}_K) \leftarrow \widehat{Q}_K(\mathbf{H}_K, \pi^{opt}_K)$.
    \ENDFOR
    \FOR{$k = K-1$ to $1$}
        \STATE Obtain $\hat{\pi}^{opt}_{k}(\mathbf{H}_k) = \arg\max_{a_k \in \mathcal{A}_{k}(\mathbf{H}_k)} \widehat{\mathbb{E}} [\widehat{V}^{\widehat\pi^{opt}}_{k+1}(\mathbf{H}_{k+1}) | \mathbf{H}_k, a_k]$.
        \STATE Let $\widehat{V}_k^{\widehat\pi^{opt}} (\mathbf{H}_{k}) \leftarrow \widehat{\mathbb{E}} [\widehat{V}^{\widehat{\pi}^{opt}}_{k+1}(\mathbf{H}_{k+1}) | \mathbf{H}_k, \hat{\pi}^{opt}_{k}]$.
    \ENDFOR
    \STATE \textbf{Output} $\hat{\pi}^{opt}_n = (\hat{\pi}^{opt}_1, \dots, \hat{\pi}^{opt}_K)$, each $\hat{\pi}^{opt}_k(\cdot)$ is deterministic.
\end{algorithmic}
\end{algorithm}
\normalsize

We establish LUQ-Learning's theoretical properties in two steps. Proposition~\ref{prop:luq_opt} provides the population-level optimality guarantee, showing that LUQ-Learning under the true outcome and preference distributions recovers the optimal policy defined in \eqref{equation:opt_condition}. Theorem~\ref{thm2} establishes consistency, showing that the value of the estimated policy $\hat{\pi}_n$ converges in probability to that of the true optimal $\pi^{opt}$ under high-level conditions on the preference and outcome model estimators.

\begin{proposition}
\label{prop:luq_opt}
    LUQ-Learning (Algorithm 1) based on the true Q and value function finds $\pi^{opt}$ satisfying the optimal condition: $V_1^{\pi^{\text{opt}}}(\mathbf{H}_1) \geq V_1^{\pi}(\mathbf{H}_1), \quad \forall \pi \in \Pi, \;$ \text{for a.s. $\mathbf{H}_1$}.
\end{proposition}

Let $P$ denote the true data-generating distribution, and let 
$\theta_0$ denote the true parameter governing the $\theta$-dependent component 
of
$P(\mathbf{W}_{K+1} \mid \mathbf{Y}, \mathbf{H}_K, \mathbf{E}; \theta)
\prod_{k=1}^K 
P(\mathbf{W}_k \mid \mathbf{X}_k, \mathbf{H}_{k-1}, \mathbf{E}; \theta).$
We write $\hat{\theta}_n$ for its estimator.
We use $\|\cdot\|_{P}$ to denote the $L^2(P)$ norm, and $\|\cdot\|_{L^\infty(P)}$ the $L^\infty(P)$ norm. We implicitly require $X \in L^2(P)$ whenever we write $\|X\|_{P}$ in the assumption.

\begin{theorem}
\label{thm2}
Assume (A1)-(A5) and $K < \infty$ fixed.
Denote $\widehat{\pi}^{opt}_n$ the estimated policy from LUQ-Learning Algorithm~\ref{alg:luq_v1}.
Let  $r_{E,n}$, $r_{Y,n}$, $r_{V,k,n} \geq 0$ be deterministic sequences such that: \\
(V1)  $||\widehat{\mathbb E}[\mathbf E|\mathbf H_K; \hat\theta_n ]-\mathbb E[\mathbf E|\mathbf H_K; \theta_0]||_{P_{\theta_0}} = O_p(r_{E,n})$; \\
(V2) $\|\widehat{\mathbb E}[\mathbf Y|\mathbf H_K,A_K]\|_{L^\infty(P_{\theta_0})} < \infty$ and \;$\|\widehat{\mathbb E}[\mathbf Y|\mathbf H_K,A_K] - \mathbb E[\mathbf Y|\mathbf H_K,A_K]\|_{P_{\theta_0}} = O_p(r_{Y,n})$; \\
(V3) $\| \widehat{\mathbb E}[\widehat V^{\widehat{\pi}^{opt}}_{n,k}(\mathbf H_k)|\mathbf H_{k-1},A_{k-1}]-{\mathbb E}[\widehat V^{\widehat{\pi}^{opt}}_{n,k}(\mathbf H_{k})|\mathbf H_{k-1},A_{k-1}] \|_{P_{\theta_0}} = O_p(r_{V,k,n})$, for $k = 2, \dots, K$.\\
Then $V_1(\pi^{opt}) - V_1(\widehat{\pi}^{opt}_n) = O_p(r_{E,n} + r_{Y,n} + r_{V,n})$, where $r_{V,n} = \sum_{j=2}^K r_{V,j,n}$. In particular, $V_1(\pi^{opt}) - V_1(\widehat\pi^{opt}_n) \to_p 0$ when $r_{E,n}, r_{Y,n}, r_{V,n} \to 0$ as $n \to \infty$.
\end{theorem}

Conditions (V1)--(V3) are rate conditions that decouple policy
consistency from specific modeling choices. The value gap $V_1(\pi^{opt}) - V_1(\widehat{\pi}^{opt}_n)$ is driven by additive accumulation of three sources of estimation error: preference model error $(r_{E,n})$, outcome regression error $(r_{Y,n})$, and propagated value-function regression error $(r_{V,n} = \sum_{j=2}^K r_{V,j,n})$, which is a finite sum since $K < \infty$ is fixed.

Condition (V1) requires $L^2(P_{\theta_0})$ convergence of the estimated posterior mean preference. Under a correctly specified parametric model for $P(\mathbf{W}_{K+1} \mid \mathbf{Y}, \mathbf{H}_K, \mathbf{E}; \theta)$ and $P(\mathbf{W}_{k} \mid \mathbf{X}_k, \mathbf{H}_{k-1}, \mathbf{E}; \theta)$, consistency and asymptotic normality of $\hat\theta_n$ as maximizer of a penalized M-estimator yield a parametric rate $r_{E,n} = n^{-1/2}$ (Lemma~1.4, Supplementary Material). When posterior expectations are approximated via Monte Carlo (MC) integration, $\big\| \widehat{\mathbb{E}}[\mathbf{E} \mid \mathbf{H}_K; \hat{\theta}_n] - \mathbb{E}[\mathbf{E} \mid \mathbf{H}_K; \hat{\theta}_n] \big\|_{P_{\theta_0}}$ where $\widehat{\mathbb{E}}[\mathbf{E} \mid \mathbf{H}_K; \hat{\theta}_n] = \sum_{b=1}^{N_{sim}} \mathbf{E}^{(b)} P(\mathbf{H}_K \mid \mathbf{E}^{(b)}; \hat{\theta}_n) \big/ \sum_{b=1}^{N_{sim}} P(\mathbf{H}_K \mid \mathbf{E}^{(b)}; \hat{\theta}_n)$ for prior draws $\mathbf{E}^{(b)} \sim \Lambda_E$, can be strictly controlled to be $o_p(n^{-1/2})$ by ensuring the number of MC draws $N_{sim}$ grows sufficiently fast relative to sample size $n$. 

Conditions (V2) and (V3) govern the $L^2(P_{\theta_0})$ convergence of the conditional mean estimators. For a $d$-dimensional history $\mathbf{H}_k$, if the true conditional expectations are $s$-times continuously differentiable, traditional non-parametric estimators such as penalized spline regression achieve the minimax optimal rate $r_{Y,n}, r_{V,k,n} \asymp n^{-\frac{s}{2s+d}}$ \citep{Stone_1982, Peter_2005}. Because this rate suffers from the curse of dimensionality as $d$ increases, adaptive machine learning estimators such as random forests \citep{Wager03072018} and  deep neural networks \citep{SchmidtHieber2017} are preferable in moderate-to-high dimensions if there exists a lower dimension manifold structure.

\subsection{Uncertainty-aware Variations}
\label{sec:uncertainty-aware}
Algorithm~\ref{alg:luq_v1} targets the population optimality condition \eqref{equation:opt_condition} by maximizing expected utility under the posterior mean preference. In practice, however, preference model estimation introduces uncertainty in $\widehat{P}(\mathbf{E}\mid\mathbf{H}_k)$ that propagates to treatment recommendations and, ultimately, to value. Because the full estimated posterior $\widehat{P}(\mathbf{E}\mid\mathbf{H}_k)$ is available, distributional summaries beyond the mean can be exploited to produce uncertainty-aware policies. 

The \textit{modal policy} selects the action that is individually optimal for the
largest proportion of the posterior preference distribution:
\[
\hat{\pi}^{mod}_k(\mathbf{H}_k) = \arg\max_{a_k \in \mathcal{A}_{k}(\mathbf{H}_k)}P\!\left(
\arg\max_{a_k'} \, \mathbf{E}_k ^{\top}\! \widehat{\mathbb{E}}\big[\mathbf{Y}\mid \mathbf{H}_k, a_k'\big]
= a_k
\;\middle|\;
\mathbf{E}_k \sim \widehat{P}(\mathbf{E}\mid \mathbf{H}_k)
\right),
\]
obtained by drawing from $\widehat{P}(\mathbf{E}\mid\mathbf{H}_k)$,
computing the optimal action for each draw, and selecting the most frequent action. The \textit{robust policy} guards against worst-case preferences within a posterior uncertainty region:
\begin{align*}
\hat{\pi}_k^{\mathrm{rob}}(\mathbf H_k)
=
\arg\max_{a_k \in \mathcal{A}_{k}(\mathbf{H}_k)}
\;
\min_{\mathbf E \in \mathcal U_k(\mathbf H_k)}
\;
\mathbf E^{\top}
\widehat{\mathbb{E}}\!\left[\mathbf Y \mid \mathbf H_k, a_k \right], \text{where} \\
\mathcal U_k(\mathbf H_k)
=
\left\{
\mathbf E \in \Delta^{d-1}
:
(\mathbf E - \widehat{\bar{\mathbf E}}_k)^{\top}
\widehat{\Sigma}_k^{-1}
(\mathbf E - \widehat{\bar{\mathbf E}}_k)
\leq \delta^2
\right\},
\label{eq:robust_policy}
\end{align*}
where $\widehat{\bar{\mathbf E}}_k = \widehat{\mathbb{E}}[\mathbf{E} \mid \mathbf{H}_k]$, $\widehat{\Sigma}_k = \widehat{\mathrm{Cov}}(\mathbf{E}\mid\mathbf{H}_k)$ the
posterior mean and covariance estimate, and $\delta > 0$ controls the size of the ellipsoidal uncertainty
region.

While $\widehat{\pi}^{opt}_n$ (via Algorithm~\ref{alg:luq_v1}) is asymptotically optimal under expected utility maximization, $\widehat{\pi}^{\mathrm{mod}}$ and $\widehat{\pi}^{\mathrm{rob}}$ target distinct objectives that may better reflect clinical priorities when preference heterogeneity is large or worst-case guarantees are preferred. 
The following proposition formalizes conditions under which the selected action sequence yields a dynamic treatment regime that is at most $\epsilon$-suboptimal relative to $\pi^{opt}$.

\begin{proposition}[Unified $\epsilon$-Optimality]
\label{prop:unified}
Let $\bar{\mathbf{E}}_k = \mathbb{E}[\mathbf{E}\mid\mathbf{H}_k]$,
$\Sigma_k = \mathrm{Cov}(\mathbf{E}\mid\mathbf{H}_k)$, and
$\mathbf{y}_k(a_k) = \widehat{\mathbb{E}}[\mathbf{Y}\mid\mathbf{H}_k,a_k]$.
For any deterministic policy $\tilde\pi$ and non-negative sequences
$\{\phi_k\}_{k=1}^K$ and vectors $\{\mathbf{c}_k\}_{k=1}^K$, if the single-stage
utility gap
$
\Delta_k(\tilde\pi_k)
:= \bar{\mathbf{E}}_k^\top\mathbf{y}_k(\pi_k^{\mathrm{opt}})
   - \bar{\mathbf{E}}_k^\top\mathbf{y}_k(\tilde\pi_k)
\;\geq\; 0
$
satisfies
\begin{equation}
  \Delta_k(\tilde\pi_k) \;\leq\; \phi_k\,\sqrt{\mathbf{c}_k^\top\Sigma_k\,\mathbf{c}_k}
  \qquad \text{a.s.\ for each } k=1,\ldots,K,
  \label{eq:gap_condition}
\end{equation}
then the multi-stage value gap satisfies
$
  V_1(\pi^{\mathrm{opt}}) - V_1(\tilde\pi)
  \;\leq\;
  \sum_{k=1}^K \mathbb{E}\!\left[
    \phi_k\,\sqrt{\mathbf{c}_k^\top\Sigma_k\,\mathbf{c}_k}
  \right].
  \label{eq:unified_bound}
$
\end{proposition}

The proposition reduces the $\epsilon$-optimality analysis to verifying condition
\eqref{eq:gap_condition} for a specific $(\phi_k, \mathbf{c}_k)$ pair. The modal and
robust policies satisfy this condition through distinct pairs, as shown in the following corollaries.

\begin{corollary}[$\epsilon$-Optimality of the Modal Policy]
\label{cor:modal}
Let $
\pi_k^{\mathrm{ind}}(a_k\mid\mathbf{H}_k)
:=
\mathbb{P}\!( 
\arg \max_{a_k'} \\
\mathbf{E}^{\top}\mathbf{y}_k(a_k') = a_k
\mid
\mathbf{E} \sim \hat{P}(\mathbf{E}\mid \mathbf{H}_k)
)$ denote the action distribution induced
by the estimated posterior.
For any deterministic policy $\tilde\pi \neq \pi^{opt}$ satisfying
$\pi_k^{\mathrm{ind}}(\tilde\pi_k(\mathbf{H}_k)\mid\mathbf{H}_k) \geq \alpha_k > 0$
almost surely, condition~\eqref{eq:gap_condition} holds with
$
  \phi_k = \sqrt{\tfrac{1-\alpha_k}{\alpha_k}},
  \,
  \mathbf{c}_k = \mathbf{d}_k
  := \mathbf{y}_k(\tilde\pi_k) - \mathbf{y}_k(\pi_k^{\mathrm{opt}}).
$
Hence by Proposition~\ref{prop:unified}, the value gap satisfies
\[
  V_1(\pi^{\mathrm{opt}}) - V_1(\tilde\pi)
  \;\leq\;
  \sum_{k=1}^K\mathbb{E}\!\left[
    \sqrt{\tfrac{1-\alpha_k}{\alpha_k}}\,
    \sqrt{\mathbf{d}_k^\top\Sigma_k\,\mathbf{d}_k}
  \right].
\]
Thus if distributing $\epsilon$ uniformly across K stages, $\tilde\pi$ is $\epsilon$-optimal whenever $\alpha_k \geq \alpha_k^* :=
K^2\sigma_k^2/(K^2\sigma_k^2 + \epsilon^2)$ for each $k$, where
$\sigma_k := \E[\sqrt{\mathbf{d}_k^{\top} \Sigma_k \mathbf{d}_k}]$.
In particular, the modal policy $\widehat{\pi}^{\mathrm{mod}}$ maximizes $\alpha_k$ at each
stage.
\end{corollary}

The result from Corollary~\ref{cor:modal} is applicable to but extends beyond the modal policy $\widehat{\pi}^{\mathrm{mod}}$ to any deterministic policy $\tilde\pi$ whose actions receive strictly positive mass under the induced distribution 
$\pi_k^{\mathrm{ind}}(\cdot|\mathbf{H}_k)$. This generality is useful in settings where clinical constraints or regulatory requirements restrict the action space.

\begin{corollary}[$\epsilon$-Optimality of the Robust Policy]
\label{cor:robust}
Let $\widehat{\pi}^{\mathrm{rob}}$ be the robust policy with uncertainty radius $\delta > 0$. Then
condition~\eqref{eq:gap_condition} holds with
$
  \phi_k = \delta,
  \,
  \mathbf{c}_k = \mathbf{y}_k(\widehat{\pi}_k^{\mathrm{rob}}),
$
and by Proposition~\ref{prop:unified}, the value gap satisfies
\[
  V_1(\pi^{\mathrm{opt}}) - V_1(\widehat{\pi}^{\mathrm{rob}})
  \;\leq\;
  \delta\sum_{k=1}^K\mathbb{E}\!\left[
    \sqrt{\mathbf{y}_k(\widehat{\pi}_k^{\mathrm{rob}})^\top\Sigma_k\,
    \mathbf{y}_k(\widehat{\pi}_k^{\mathrm{rob}})}
  \right]
  \;\leq\;
  \delta \, M \, \sum_{j=1}^K \sqrt{\lambda_{\max}(\Sigma_j)},
\]
where $M = \sup_{k,\mathbf{H}_k,a_k}\|\mathbf{y}_k(a_k)\|$ and
$\lambda_{\max}(\Sigma_k)$ the largest eigenvalue of $\Sigma_k$. Consequently, $\widehat{\pi}^{\mathrm{rob}}$
is $\epsilon$-optimal whenever
$\delta \leq \delta^* = \epsilon / (M \sum_{j=1}^K \sqrt{\lambda_{max}(\Sigma_j)})$.
\end{corollary}

Corollary~\ref{cor:robust} provides the threshold $\delta^* = \epsilon/(M\sum_{j=1}^K\sqrt{\lambda_{\max}(\Sigma_j)})$ that guarantees $\epsilon$-optimality for all patients. As $\delta$ serves as a global hyperparameter, one could calibrate it using offline validation data: compute $\widehat{S} = n_{\text{val}}^{-1}\sum_{i=1}^{n_{\text{val}}}\sum_{j=1}^K\sqrt{\lambda_{\max}(\widehat{\Sigma}_j^{(i)})}$ over a held-out set of trajectories, estimate $\widehat{M} = \max_{k,i,a_k}\|\widehat{\mathbf{y}}_k(a_k)\|$, and set $\delta^* = \epsilon/(\widehat{M}\cdot\widehat{S})$. 
Alternatively, one could specify stage-adaptive radii $\{\delta_j\}_{j=1}^K$ that adapt to posterior concentration. For instance, under parametric preference models where $\mathbf{E} \mid \theta \sim \mathrm{Dirichlet}(\alpha_1(\theta), \ldots, \alpha_d(\theta))$ 
and each observation $\mathbf{W}_j$ provides positive Fisher information about $\theta$ at the posterior mean, the posterior precision satisfies $\Sigma_{j+1}^{-1} = \Sigma_j^{-1} + I(\mathbf{W}_j; \theta) \succeq \Sigma_j^{-1}$, implying $\lambda_{\max}(\Sigma_j) \geq \lambda_{\max}(\Sigma_{j+1})$ via the inverse order property of positive definite matrices. When this monotonicity holds, one can adopt stage-adaptive radii $\delta_j = \delta_0\sqrt{\lambda_{\max}(\Sigma_j)/\lambda_{\max}(\Sigma_1)}$, where $\delta_0$ is calibrated on validation data to satisfy $\delta_0 = \epsilon/(M \mathbb{E}[\sum_{k=1}^K \lambda_{\max}(\Sigma_k)/\sqrt{\lambda_{\max}(\Sigma_1)}])$, allocating more uncertainty budget at early stages (where $\Sigma_j$ is large) and less at later stages (where $\Sigma_j$ has contracted). This adaptive allocation achieves the same $\epsilon$-optimality guarantee as constant $\delta$ while permitting a larger base radius $\delta_0$.

The threshold $\delta^*$ and the posterior mass certificate $\alpha^*_k$  are both estimable from data and can be selected via cross-validation due to their role as hyperparameters, providing a concrete basis for policy selection that we formalize in Section~\ref{sec:policy-selection}.

\subsection{Policy Implementation in Practice}
\label{sec:policy-selection}

The three policies derived under LUQ-Learning, mean ($\widehat{\pi}^{opt}_n$), modal ($\widehat{\pi}^{\mathrm{mod}}$), and robust ($\widehat{\pi}^{\mathrm{rob}}$), coincide and are exactly optimal in the sense of condition \eqref{equation:opt_condition} when $\Sigma_k = 0$, i.e., when observed history $\mathbf{H}_k$ perfectly identifies the patient's preferences. When $\Sigma_k \neq 0$, the $\epsilon$-optimality bounds from Corollaries~\ref{cor:modal} and~\ref{cor:robust} provide a data-driven basis for policy selection. 
Specifically, we recommend the following workflow: estimate 
$\widehat{\Sigma}_k$ and compute $\alpha_k^{*}$ and $\delta^*$ on a held-out validation set, then select the policy whose tolerance parameter satisfies the prespecified $\epsilon$-optimality threshold, and choose the policy that best aligns with clinical priorities and the observed distribution of posterior uncertainty.

The three policies occupy distinct positions along the tradeoff between average performance and preference robustness. The mean policy $\widehat{\pi}^{opt}_n$ is preferred when the preference model is well-specified and posterior variance $\Sigma_k$ is small. The modal policy $\widehat{\pi}^{\mathrm{mod}}$ is preferable when $\Sigma_k$ is large and the clinical objective is to recommend a treatment that is individually optimal for the largest patient subgroup, accepting a modest population-average loss in exchange for broader individual coverage. The robust policy $\widehat{\pi}^{\mathrm{rob}}$ is appropriate when worst-case preference realizations carry disproportionate clinical consequences and a conservative guarantee is required.

The following remarks address important practical considerations for deploying LUQ-Learning.

\begin{remark}[Outcome Standardization]
Since LUQ-Learning maximizes $U^* = \mathbf{E}^\top\mathbf{Y}^*$, the coordinates of $\mathbf{Y}$ must be at a comparable scale, and signs assigned so that larger values are uniformly preferable. The coordinates of $\mathbf{Y}$ should span distinct aspects of the clinical objective, ensuring the convex hull of $\mathbf{Y}$ is sufficiently rich to contain the true patient utility. When outcomes are collinear or redundant, the effective dimension of $\Delta^{d-1}$ is reduced and preference heterogeneity may not be recoverable from $\mathbf{W}_k$ alone.
\end{remark}

\begin{remark}[Preference Data Collection]
Preference instruments $\mathbf{W}_k$ need not be collected at every stage; omitting collection at certain stages reduces the precision of 
$\widehat{P}(\mathbf{E}\mid\mathbf{H}_k)$ but does not invalidate the algorithm, as equation~\eqref{eq:posterior} remains well-defined. Moreover, $\mathbf{W}_k$ may be measured either before or after $\mathbf{X}_k$ in the data trajectory, provided it reflects the patient state following action $A_{k-1}$.
\end{remark}

\begin{remark}[Preference Model Selection and Diagnostics]
Condition~(V1) in Theorem~\ref{thm2} depends on the quality of the fitted preference model $P(\mathbf{W}_k\mid\mathbf{X}_k,\mathbf{H}_{k-1},\mathbf{E};\hat\theta_n)$. We recommend selecting among candidate models via cross-validation using held-out log-likelihood or BIC \citep{gelman_bayesian_2014}, with posterior predictive 
checks as a complementary diagnostic for systematic discrepancies between 
observed and model-predicted distributions of $\mathbf{W}_k$. These diagnostic steps are critical in practice. The guarantees of $\widehat{\pi}^{\mathrm{rob}}$ pertain to preference variation within a correctly specified model; if $\widehat{P}(\mathbf{E}\mid\mathbf{H}_k)$ is poorly calibrated, the uncertainty set $\mathcal{U}_k$ may fail to cover the true preference distribution.
\end{remark}

\section{Application to the BEST Study}
\label{sec:application}
\subsection{The BEST Study}
The Biomarkers for Evaluating Spine Treatments (BEST)
Trial is a two-stage SMART that has an expected sample size of at least 600 subjects. The study was motivated by the observation that while many treatments show small-to-moderate average treatment effects, some patients appear to benefit substantially from certain specific treatment plans. Our simulation is designed to reflect the data structure in the BEST study.

The full data trajectory for patients in BEST, denoted as $\mathbf{H}_3$, can be summarized as $(\mathbf{X}_1, \mathbf{W}_1, A_1, \mathbf{X}_2, \mathbf{W}_2, A_2, \mathbf{Y}, \mathbf{W}_3)$, where $\mathbf{Y} \in \mathbb{R}^3$. Three types of questionnaires were used to collect preference information. First, a questionnaire adapted from the CAPER Treatment framework \citep{wilson_preferences_2024, Wilson2023} was administered, with modified attributes to focus on outcome preferences. It contains 12 binary questions, each of which asks patients to choose one over another described scenario. Denote $\mathbf{W}^{B}_k$ responses to this questionnaire. Second, there is a question asking patients to rank the three outcomes. Denote $\mathbf{W}^{R}_k$ their ordinal responses to this question. Third, there is a questionnaire asking how satisfied are they with the most recent treatment received on a scale of one to seven. Denote responses to this question $\mathbf{W}^{Sat}_k$. Following the study design, we have $\mathbf{W}_1 = (\mathbf{W}^B_1, \mathbf{W}^R_1)$, $\mathbf{W}_2 = (\mathbf{W}^B_2, \mathbf{W}^R_2, \mathbf{W}^{Sat}_2)$, $\mathbf{W}_3 = (\mathbf{W}^{Sat}_3)$. In our simulation, we define $\mathbf{X}_1$, $\mathbf{X}_2$, and $\mathbf{Y}$ as 10 minus the PEG score \citep{Krebs2009} so that they remain on a scale of 0-10, but with a higher score corresponding to a better pain experience.

At the first randomization stage, $\mathcal{A}_{1} = \{a_1, \dots, a_4\}$ and all subjects are randomized to one of the four treatments with equal probability. In the second randomization stage, $\mathcal{A}_{2} = \{ \{a_j\}, \{a_j, a_k\}: j,k = 1, \dots, 4 \}$ where the specific subset depends on the observed value of $\mathbf{H}_2$. Denote the response groups after the first treatment $\mathcal{C} = \{c_1, \dots, c_4\}$. If $C = c_1$, it indicates that a patient responds well to the first treatment, so the patient maintains the previously assigned treatment; if $C = c_2$, we randomize subjects to a specific treatment augmenting plan; if $C = c_3$, we randomize subjects to receive a randomly assigned treatment augmentation or to switch to a randomly assigned new treatment; finally, if $C = c_4$, we consider the first treatment non-effective and randomize subjects to a new treatment. The exception is when $A_1= a_1, C\in\{c_3, c_4\}$, in which case a patient will always augment the current treatment instead of switching due to the nature of $a_1$. We refer readers to \cite{sperger_statistical_2025} and \cite{mauck_design_2025} for additional details on the trial design.

We specify the following model for stated preferences $\mathbf{W}$.
\footnotesize
\begin{align*}
& \mathbf{V} \sim \mathcal{N}_{2}(0,\mathbf{I}), \\
& \mathbf{E} = \text{SoftMax}((\mathbf{V}, 1)) = \frac{(\exp(\mathbf{V}), 1)}{ \sum_{j=1}^2 \exp(V_j) + 1}, \\
& \mathbf{W}^B_{1j} | \mathbf{V} \sim \text{Bern}\left(p = \sigma\left(\beta_{1,j,0}+\beta^\top_{1,j,1}\mathbf{V}\right)\right), \quad (1 \leq j \leq 12),\\
& P(\mathbf{W}_1^R = \mathbf{w}^R | \mathbf{E}^R) = \frac{\exp\left(-\lambda_1 \tau(\mathbf{w}^R,\mathbf{E}^R)\right)}{\sum_{\mathbf{v}^R \in \mathcal{\textit{Perm}}} \exp\left(-\lambda_1 \tau(\mathbf{v}^R,\mathbf{E}^R)\right)}, \\
& P(\mathbf{W}^{Sat}_2 \leq j | \mathbf{X}_2, \mathbf{E}) = \sigma\left(\alpha_{2,j,0} - \alpha_{2,\cdot,1} \mathbf{E}^\top \mathbf{X}_2\right), \quad (1 \leq j \leq 6), \\
& \mathbf{W}^{B}_{2j} | \mathbf{V} \sim \text{Bern}\left(p = \sigma\left(\beta_{2,j,0}+\beta^\top_{2,j,1}\mathbf{V}\right)\right), \quad (1 \leq j \leq 12),\\
& P(\mathbf{W}_2^R = \mathbf{w}^R | \mathbf{E}^R) = \frac{\exp\left(-\lambda_2 \tau(\mathbf{w}^R,\mathbf{E}^R)\right)}{\sum_{\mathbf{v}^R \in \mathcal{\textit{Perm}}} \exp\left(-\lambda_2 \tau(\mathbf{v}^R,\mathbf{E}^R)\right)}, \\
& P(\mathbf{W}^{Sat}_3 \leq j | \mathbf{Y}, \mathbf{E}) = \sigma\left(\alpha_{3,j,0} - \alpha_{3,\cdot,1} \mathbf{E}^\top \mathbf{Y}\right), \quad (1 \leq j \leq 6)
\end{align*}
\normalsize
where $\sigma(\cdot)$ denotes the sigmoid function, $\tau(\cdot, \cdot)$ the Kendall's Tau metric with $\mathbf{E}^R$ the rank vector of coordinates of $\mathbf{E}$, and $\textit{Perm}$ the set of all permutations of $\{1,2,3\}$. For computational tractability and ease of comparison, we follow the modeling choice made by \citet{Butler2018} for $\mathbf{E}$ and $\mathbf{W}^B | \mathbf{V}$, assuming binary preference questions related to latent factors $\mathbf{V}$ through independent logistic regression models. Our assumed model for $P(\mathbf{W}_k^R|\mathbf{E}^R)$ is the Mallow's $\phi$ model \citep{Tang2019}. While the BEST study allows for tied ranks, the proposed distribution excludes tied ranks for simplicity. $\mathbf{W}^{Sat}_k$ are assumed to be positively related to the preference-weighted outcomes via the proportional-odds logistic regression model, which translates to parameter space restriction $\alpha_{2,\cdot,1}, \alpha_{3,\cdot,1} > 0$.

Outcomes $\mathbf{Y}$ and $\mathbf{X}_1, \mathbf{X}_2$, and covariates other than preferences, are generated as follows. The action set at the second decision time is $\mathcal{A}_{2} = \{ \{x\}, \{x, y\} : x, y \in \{a_1, \dots, a_4\}, x \neq y\}$, with cardinality $|\mathcal{A}_{2}| = 10$.
\footnotesize
\begin{align*}
& {X}_{1j} \sim \text{Bin}(n = 10, p = 0.5), \quad (1 \leq j \leq 3),\\
& A_1 \sim \text{Unif}(\mathcal{A}_{1}), \quad \mathcal{A}_1 = \{a_1, \dots, a_4\}, \\
& X_{2j} \sim \text{Bin}\left(n = 10, p = \sigma\left[\sum_{a_l \in \mathcal{A}_{1}} \gamma_{2,j,l,0} I(A_1 = a_l) + \frac{X_{1j} - \widehat{\mathbb{E}}[X_{1j}]}{\widehat{\mathrm{SD}}(X_{1j})} \sum_{a_l \in \mathcal{A}_{1}} \gamma_{2,j,k,1} I(A_1 = a_l) \right] \right), \quad (1 \leq j \leq 3), \\
& C = \sum_{m=1}^4 c_m \, I\!\left\{\hat q_{m-1} < \frac{1}{3}\sum_{j=1}^3 X_{1j} \le \hat q_m \right\}, \quad
\mathcal{C}=\{c_1,c_2,c_3,c_4\},\\ 
&\text{with } (\hat q_1,\hat q_2,\hat q_3) \text{ the empirical } (0.25,0.50,0.75)\text{-quantiles of } \{\bar X_{1i}\}_{i=1}^n, \text{ and } \hat q_0=-\infty,\; \hat q_4=\infty,\\
& A_2 = \begin{cases}
\begin{array}{lll}
\{A_1\}, & & C = c_1 \\
\{A_1, \tilde{A}\}, & \tilde{A} \sim \text{Unif}(\mathcal{A}_{1} \setminus A_1) & C = c_2 \text{ or } (A_1 = a_1 \text{ and } C \in \{c_3, c_4\}) \\
B\{A_1, \tilde{A}\} + (1 - B)\{\tilde{A}\}, & B \sim \text{Bern}(0.5), \tilde{A} \sim \text{Unif}(\mathcal{A}_{1} \setminus A_1) & C = c_3 \text{ and } A_1 \neq a_1 \\
\{\tilde{A}\}, & \tilde{A} \sim \text{Unif}(\mathcal{A}_{1} \setminus A_1) & C = c_4 \text{ and } A_1 \neq a_1 \\
\end{array},
\end{cases} \\
& Y_j \sim \text{Bin}\left(n = 10, p = \sigma\left[\sum_{a_l \in \mathcal{A}_{1}} \gamma_{3,j,l,0} I(a_l \in A_2) + \frac{X_{2j} - \widehat{\mathbb{E}}[X_{2j}]}{\widehat{\mathrm{SD}}(X_{2j})} \sum_{a_l \in \mathcal{A}_{1}} \gamma_{3,j,l,1} I(a_l \in A_2) \right] \right), \quad (1 \leq j \leq 3),
\end{align*}
\normalsize 
where $\widehat{\mathbb{E}}[X_{kj}]$ and $\widehat{\mathrm{SD}}(X_{kj})$ denote the sample mean and standard deviation of $X_{kj}$.
Denote ${\theta} = ({\alpha}, {\beta}, {\lambda})$ the unknown true parameters related to the preference model and $\gamma$ parameters for the outcome model, where ${\alpha} = (\alpha_{k,j,0}, \alpha_{k,\cdot,1})^{k=3, j = 6}_{k = 2, j = 1}$, ${\beta} = (\beta_{k,j,0}, \beta_{k,j,1})^{k=2, j = 3}_{k = 1, j = 1}$, ${\lambda} = (\lambda_k)_{k=1}^{k=2}$, ${\gamma} = (\gamma_{k,j,l,0}, \gamma_{k,j,l,1})^{k=3, j = 12, l = 4}_{k = 2, j = 1, l = 1}$. Throughout, we use $k$ to index the decision times, $j$ to index dimensionality; and $l$ to index over the action set. 

We generate the parameters as follows.
\footnotesize
\begin{align*}
&\beta_{1,j,0}=0, \quad \beta_{1,j,1} {\sim} \mathcal N_2(0,\mathbf{I}_2), \quad (1 \leq j \leq 12),\\
&\beta_{2,j,0}=0,\quad \beta_{2,j,1}=\sqrt{0.8}\beta_{1,j,1}+\sqrt{0.2}\epsilon_{\beta},\quad \epsilon_{\beta} {\sim} \mathcal N_2(0,\mathbf{I}), \quad (1\leq j \leq 12), \\
&\alpha_{2,\cdot, 1}=0.5,\quad \alpha_{2,j,0}=0.75 j, \quad (1\leq j \leq 6), \\
&\alpha_{3,\cdot,1}=0.6,\quad \alpha_{3,j,0}=\alpha_{2,j,0} + 0.5, \quad (1\leq j \leq 6), \\
&\lambda_1=0.5,\quad \lambda_2=2, \\
&\gamma_{2,j,l,0} \sim \mathcal{N}(0,0.5^2),\quad  \gamma_{2,j,l,1} \sim \mathcal{N}(0,1), \quad ( 1\leq j \leq 3, \quad 1 \leq l \leq 4), \;\mbox{and}\\
&\gamma_{3,j=1,l,0} = 0, \quad \gamma_{3,j=3,l,0} = -\gamma_{3,j=2,l,0}, \quad (1 \leq l \leq 4), \;\mbox{with}\\
&\gamma_{3,j=2,l,0} = 2 \times (\sqrt{0.8}\gamma_{2,j=2,l,0}+\sqrt{0.2}\epsilon_{\gamma}), \quad \epsilon_{\gamma} {\sim} \mathcal{N}(0,0.5^2),\\
&\gamma_{3,j,l,1}=0, \quad (1 \leq j \leq 3, \quad 1 \leq l \leq 4)
\end{align*}
\normalsize

We set up parameters $\gamma$ so that, at the second decision time, $A_2$ has opposing effects on $Y_2$ and $Y_3$, with $Y_1\sim \text{Binomial}(n=10,p=0.5)$, $Y_2\sim \text{Binomial}(n=10,p=\sigma(g(A_2)))$ and $Y_3\sim \text{Binomial}(n=10,p=\sigma(-g(A_2)))$ where $g(A_2)= \sum_{a_l \in \mathcal{A}_{1}} \gamma_{3,2,l,0}I(a_l \in A_2)$. Now $Y_1$ can be thought of as a random intercept in our utilities $\mathbf{E}^\top\mathbf{Y}$, while $Y_2$ and $Y_3$ can be thought of as conflicting outcomes. 

We conclude this subsection with Corollary~\ref{coro1}, which provides the minimal additional assumptions specific to this BEST-tailored simulation required for LUQ-Learning to achieve consistency. Coupled with standard regularity conditions for bounded outcomes and Random Forests \citep{Scornet2015}, this guarantees $V(\hat{\pi}^{opt}_n) \to_p V(\pi^{opt})$ per Theorem~\ref{thm2}.

\begin{corollary}
\label{coro1}
Under the proposed model for the BEST study described above and the proposed estimation procedure,
$\hat\theta_n \to_p \theta_0$ as $n \to \infty$, and
$
\|\widehat{\mathbb E}[\mathbf{E}\mid \mathbf{H}_2; \hat\theta_n]
-\mathbb E[\mathbf{E}\mid \mathbf{H}_2;\theta_0]\|_{P_{\theta_0}} \to 0
$
as $N_{\mathrm{sim}}, n \to \infty$,
provided that there exists an interior point $\theta_0 \in \Theta$, with $\Theta$ compact, such that
$
P(\mathbf{H}_3 \mid \mathbf{V})
=
M_{\theta_0}(\mathbf{H}_3,\mathbf{E})\, g(\mathbf{H}_3)
$
almost surely,
where
$M_\theta(\mathbf{H},\mathbf{E})=P_\theta(\mathbf{W}\mid \mathbf{H},\mathbf{E})$
collects all $\theta$-dependent components of the proposed preference model and $g(\mathbf{H}_3)$ is a measurable function that does not depend on parameter $\theta$. 
Assume further that the model is identifiable in the sense that
$P(\mathbf{H}_3;M_{\theta_0}) \neq P(\mathbf{H}_3;M_\theta)$ for all $\theta \neq \theta_0$
and that $g(\mathbf{H}_3) > c$ a.s. for some $c>0$.
If, in addition, the Fisher information matrix $I(\theta_0)$ is nonsingular, then
$
\sqrt{n}(\hat\theta_n-\theta_0)\to_d \mathcal N\!\left(0,I(\theta_0)^{-1}\right).
$
\end{corollary}

\subsection{Simulation Result}

We evaluate LUQ-Learning across sample sizes ranging from $n = 300$ to $4800$, conducting 400 independent replicates per scenario. For each replicate, policies are estimated on training data and evaluated on independently generated testing data of the exact same size. Across all compared methods, conditional outcome and Q-function models are estimated using Random Forests to maintain flexibility in moderate sample sizes. For LUQ-Learning specifically, preference model parameters are estimated via L-BFGS optimization with a smooth barrier penalty to enforce domain constraints, and posterior mean preferences are approximated via Monte Carlo integration ($N_{sim} = 2000$). Complete computational details are provided in Section 3.1 of the Supplementary Material.

We consider the following alternative approaches for comparison: 
First, a naive approach using Q-learning \citep{Schulte2014} with the optimization objective set as the simple average of $\mathbf{Y}$, i.e., a ``naive" solution when working with multivariate objective. We denote the estimated DTR as $\hat{\pi}_{\text{Naive}}$. 
Also, we include the case of directly optimizing for a scalar $\mathbf{W}^{Sat}$, a natural alternative that does not incorporate longitudinal preference data. We denote the estimated DTR from which $\hat{\pi}_{\text{Wlast}}$. 
Lastly, for benchmarking, we further include an ``oracle'' setting where one has direct access to the true $\mathbf{E}$. This is done by replacing $\hat{\mathbb{E}}[\mathbf{E}|\mathbf{H}_2; \hat\theta_n]$ with the truth at $k = 2$; replacing $\mathbf{W}$ with $\mathbf{E}$ in the history $\mathbf{H}_1$ in the Q model at $k = 1$, and consequently letting $\pi_{\text{Known}}$ and its estimate be functions of the true $\mathbf E$.

The value improvements of the estimated DTRs relative to the observed regime are summarized in Table~\ref{table:best_n}. LUQ-Learning generally outperforms $\hat{\pi}_{\text{Naive}}$ and $\hat{\pi}_{\text{Wlast}}$ across sample sizes, particularly for small to moderate $N$. $\hat{\pi}_{\text{Wlast}}$ consistently exhibits higher variance and more unstable performances than $\hat{\pi}_{\text{LUQL}}$, with an average value improvement even worse than $\hat{\pi}_{\text{Naive}}$ with competing outcomes, demonstrating the benefit of incorporating longitudinal preference information in improving stability.

As expected, the oracle procedure $\widehat{\pi}_{\mathrm{Known}}$ achieves the largest improvements relative to the observed policy, with a relatively stable gap in the absolute value between $\widehat{\pi}_{\mathrm{Known}}$ and $\widehat{\pi}_{\mathrm{LUQL}}$ across sample sizes induced by preference estimation.
Under mis-specification of $P(\mathbf{E})$, discussed in detail in the Supplementary Material Section~3.4, this gap widens consistently, confirming the importance of model selection and diagnostics.

\begin{table}[ht]
\spacingset{1.3}
\centering
\caption{Mean (SD) of $V(\hat{\pi}) - V(\pi_{obs})$ across Sample Sizes.}
\label{table:best_n}
\begin{tabular}{lllllll}
\hline
DTR & N = 300 & N = 600 & N = 1200 & N = 2400 & N = 4800 \\
\hline
$\widehat{\pi}_{\text{Known}}$ 
& 0.169 (0.15) & 0.190 (0.13) & 0.202 (0.12) & 0.225 (0.13) & 0.303 (0.11) \\

$\widehat{\pi}_{\text{LUQL}}$ 
& 0.155 (0.14) & 0.177 (0.13) & 0.189 (0.12) & 0.213 (0.12) & 0.288 (0.11) \\

$\widehat{\pi}_{\text{Wlast}}$ 
& 0.076 (0.15) & 0.116 (0.14) & 0.153 (0.15) & 0.203 (0.16) & 0.341 (0.18) \\

$\widehat{\pi}_{\text{Naive}}$ 
& 0.148 (0.14) & 0.171 (0.12) & 0.180 (0.12) & 0.203 (0.12) & 0.277 (0.11) \\
\hline
\end{tabular}
\end{table}

\textbf{Effect of trajectory length}\\
To evaluate LUQ-Learning over longer horizons, we generalized the base simulation to accommodate trajectories of length $K \in \{2, 4, 6, 8\}$. In this extended setting, treatment assignments are uniform across all stages, terminal outcomes $\mathbf{Y}$ do not have opposing treatment effects (providing a more conservative benchmark against naive methods), and we focus exclusively on binary questionnaires and post-treatment satisfaction. To simulate evolving, correlated preferences over time, the stage-specific parameters are generated via an autoregressive structure with additive noise. The complete data-generating equations, probability distributions, and parameter initializations for this $K$-stage extension are detailed in Section 3.2 of the Supplementary Material.

\begin{figure}[htbp]
    \centering
    \begin{subfigure}[b]{0.45\textwidth}
        \centering
        \includegraphics[width=\textwidth]{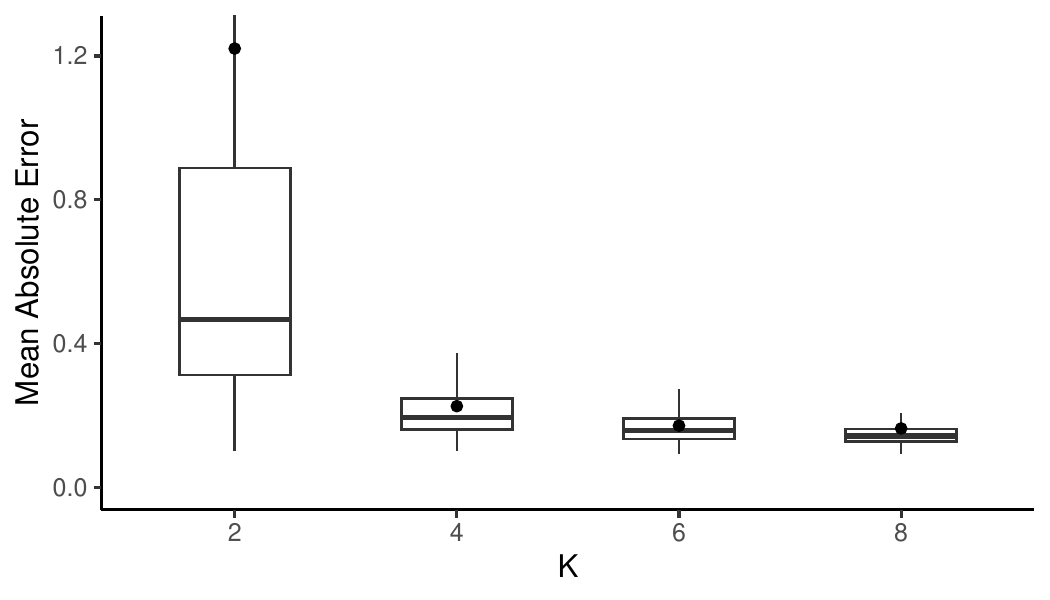}
        \caption{N = 600}
        \label{fig:sub1}
    \end{subfigure}
    \begin{subfigure}[b]{0.45\textwidth}
        \centering
        \includegraphics[width=\textwidth]{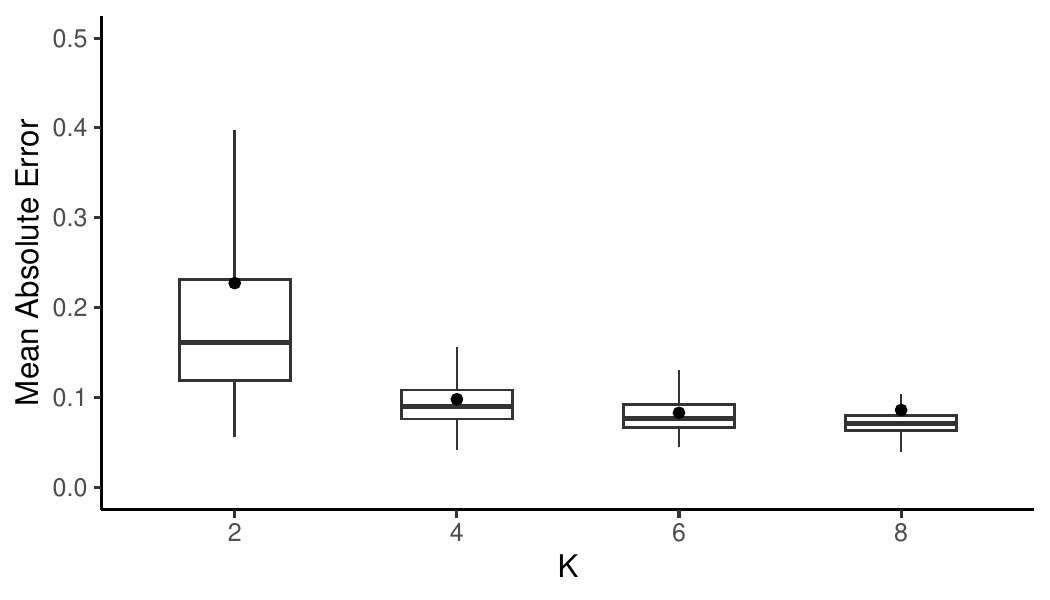}
        \caption{N = 2400}
        \label{fig:sub2}
    \end{subfigure}
    \caption{Boxplots of the Mean Absolute Error of $\hat\theta_n$ by Trajectory Lengths, Summarized over 400 Seeds, with Dots Indicating Sample Average.}
    \label{fig:error}
\end{figure}

Comparison result between LUQ-Learning and the comparators defined the same as before are summarized in Table~\ref{table:best_K}. Across all 
trajectory lengths and sample sizes, the performance ranking remains 
consistent: $\widehat{\pi}_{\text{Known}} \succ \widehat{\pi}_{\text{LUQL}} 
\succ \widehat{\pi}_{\text{Naive}} \succ \widehat{\pi}_{\text{Wlast}}$. 
Notably, the gap between $\widehat{\pi}_{\text{LUQL}}$ and 
$\widehat{\pi}_{\text{Known}}$ decreases as $K$ increases on both absolute 
and relative scales, suggesting that longer trajectories provide richer 
longitudinal preference information that improves estimation of $\theta$ 
despite the linearly expanding parameter space, as supported by 
Figure~\ref{fig:error}. In contrast, $\widehat{\pi}_{\text{Naive}}$ 
remains flat across $K$, while $\widehat{\pi}_{\text{Wlast}}$ deteriorates 
monotonically as $K$ grows, suggesting that terminal-stage proxies alone 
are increasingly inadequate for capturing evolving preference structures 
in longer trajectories.

\begin{table}[ht]
\spacingset{1.3}
\centering
\caption{Mean (SD) of $V(\hat{\pi}) - V(\pi_{obs})$ across Trajectory Lengths and Sample Sizes.}
\label{table:best_K}
\begin{tabular}{llcccc}
\hline
N & DTR & K = 2 & K = 4 & K = 6 & K = 8  \\
\hline
\multirow{4}{*}{600} 
& $\widehat{\pi}_{\text{Known}}$ 
& 1.253 (0.28) & 1.308 (0.27) & 1.318 (0.27) & 1.302 (0.27) \\
& $\hat \pi_{\text{LUQL}}$ 
& 1.166 (0.27) & 1.224 (0.26) & 1.261 (0.26) & 1.253 (0.27) \\
& $\widehat{\pi}_{\text{Wlast}}$ 
& 0.673 (0.31) & 0.602 (0.32) & 0.549 (0.30) & 0.506 (0.32) \\
& $\widehat{\pi}_{\text{Naive}}$ 
& 1.147 (0.28) & 1.190 (0.26) & 1.220 (0.26) & 1.207 (0.26) \\
\hline
\multirow{4}{*}{2400} 
& $\widehat{\pi}_{\text{Known}}$ 
& 1.312 (0.26) & 1.379 (0.28) & 1.382 (0.26) & 1.366 (0.27) \\
& $\hat \pi_{\text{LUQL}}$ 
& 1.207 (0.27) & 1.262 (0.26) & 1.284 (0.25) & 1.281 (0.26) \\
& $\widehat{\pi}_{\text{Wlast}}$ 
& 0.893 (0.27) & 0.849 (0.30) & 0.779 (0.27) & 0.719 (0.28)  \\
& $\widehat{\pi}_{\text{Naive}}$ 
& 1.176 (0.27) & 1.212 (0.25) & 1.217 (0.25) & 1.215 (0.25)  \\
\hline
\end{tabular}
\end{table}

We report simulation results under another design motivated by the Clinical Antipsychotic Trials of Intervention Effectiveness (CATIE) study in Supplementary Material, Section~4. This is a single-stage setting, allowing direct comparison with the approach of \citet{Butler2018}, as well as with $\widehat{\pi}_{\mathrm{Wlast}}$, $\widehat{\pi}_{\mathrm{Naive}}$, and $\widehat{\pi}_{\mathrm{Known}}$, previously defined. LUQ-Learning achieves substantially smaller preference estimation error than Butler’s method across all sample sizes (Table~4, Supplementary Material). Consistent with this improvement, LUQ-Learning yields larger value gains than Butler’s approach and outcome-based Q-learning baselines (Table~5, Supplementary Material). These results suggest that the proposed framework retains advantages over existing methods beyond multi-stage decision problems by flexibly accommodating diverse preference data sources.

All scripts used to create the simulation results can be found on GitHub at: \href{https://github.com/yatingz205/LUQ-Learning.git}{git@github.com:yatingz205/LUQ-Learning.git}.

\section{Discussion}
\label{sec:discussions}

We have introduced Latent Utility Q-Learning (LUQ-Learning), a framework for estimating dynamic treatment regimes that optimize individualized, preference-weighted combinations of possibly competing outcomes over finitely many decision points. The central methodological contribution is a reduction that separates latent preference estimation from outcome regression through a conditional mean factorization, allowing preference-adaptive regime learning to be conducted via a sequence of scalar decision problems. This decomposition allows preference and outcome estimators to be specified independently, accommodates flexible estimation of both preference and outcome components, and enables uncertainty-aware policy selection with explicit $\epsilon$-optimality guarantees relative to the oracle regime. Consistency is established under high-level conditions on nuisance estimators, permitting parametric, semi- or non-parametric implementations.

This work leads to several promising future directions. First, formal identifiability of the latent preference model under semi- or non-parametric specifications of $P(\mathbf{W}_k \mid \mathbf{X}_k, \mathbf{H}_{k-1}, \mathbf{E})$ remains open; progress here could benefit from a broad class of hierarchical latent variable models \citep{Geometry_Lindsay_1983, Allman2009, gustafson_bayesian_2015}. Second, extending utility classes beyond $\mathbf{E}^\top \mathbf{Y}^*$ to allow latent preferences to interact non-linearly with distributional quantities of $\mathbf{Y}^*$ would broaden applicability. Third, jointly optimizing the preference data collection scheme to minimize posterior variance $\Sigma_k$ and improve estimation of $P(\mathbf{E}\mid \mathbf{H}_k)$ connects LUQ-Learning to optimal experimental design \citep{chaloner_bayesian_1995, lopez_fidalgo_optimal_2023} and online learning literature \citep{Chen03042021}.

Sequential decision-making under latent, individual-specific objectives, as considered here, arises broadly beyond clinical medicine, including personalized education, behavioral interventions, and recommendation systems. In online and mobile health settings where preference signals are collected continuously via wearables or ecological momentary assessments, the posterior $P(\mathbf{E} \mid \mathbf{H}_k)$ updates sequentially as observations accumulate, connecting LUQ-Learning naturally to just-in-time adaptive interventions \citep{nahum_2018} 
and the contextual bandit literature \citep{bandit_2007, Foster2021EfficientFC}, 
where real-time personalization is essential. 
More substantively, LUQ-Learning provides a causal-inference-grounded alternative to reinforcement learning from human feedback by treating preference instruments as identifiable proxies for latent utilities, rather than as sources of relative trajectory rankings. The identification structure further parallels proximal causal inference \citep{miao_identifying_2018}, suggesting that tools for handling imperfect proxies from that literature may strengthen the theoretical foundations of preference-incorporated DTRs beyond the parametric conditions established here.

From a practical standpoint, deployment of preference-adaptive regimes requires careful study design. Preferences should be elicited using validated instruments that capture stable, long-term trade-offs rather than transient treatment satisfaction, to avoid myopic optimization at the expense of durable outcomes. Researchers should also ensure alignment between development and target populations, as cultural, regional, and socioeconomic factors can systematically influence reported preferences. More broadly, the framework strengthens coherence among the recommendation algorithm, healthcare providers, and patients; safeguards against preference manipulation; and integrates algorithmic recommendations with clinical judgment, as is crucial in high-stakes settings. When these considerations are formally addressed through rigorous trial design and prospective validation, the proposed framework provides a statistically principled foundation for preference-adaptive decision-making in complex, multi-outcome clinical environments.

\section{Disclosure Statement}
The authors report that there are no competing interests to declare.

\printbibliography

\end{document}


\def\spacingset#1{\renewcommand{\baselinestretch}%
{#1}\small\normalsize} \spacingset{1}


\if0\blind
{
    \title{\bf Supplementary Material for: 
    Latent Utility Q-Learning for Preference-Adaptive Dynamic Treatment Regimes}
  \author{Joshua P. Zitovsky\\
    Department of Biostatistics \\ UNC Chapel Hill \\
    \and 
    Yating Zou\\
    Department of Biostatistics \\ UNC Chapel Hill \\
    \and
    Leslie Wilson\\
    Department of Clinical Pharmacy \\
    University of California, San Francisco \\
    \and 
    Michael R. Kosorok \\
    Department of Biostatistics \\ UNC Chapel Hill}
  \date{}
  \maketitle
} \fi

\if1\blind
{
  \title{\bf Supplementary Material for: 
  Latent Utility Q-Learning for Preference-Adaptive Dynamic Treatment Regimes
  }
  \author{}
  \date{}
  \maketitle
  \vspace*{0.5in}
} \fi

\spacingset{1.2}

\counterwithin{figure}{section}
\newcommand{\argmax}[0]{\text{argmax}}
\newcommand{\bI}[0]{\mathbf{I}}
\newcommand{\bV}[0]{\mathbf{V}}
\newcommand{\bv}[0]{\mathbf{v}}
\newcommand{\bW}[0]{\mathbf{W}}
\newcommand{\bw}[0]{\mathbf{w}}
\newcommand{\bH}[0]{\mathbf{H}}
\newcommand{\bh}[0]{\mathbf{h}}
\newcommand{\bX}[0]{\mathbf{X}}
\newcommand{\bE}[0]{\mathbf{E}}
\newcommand{\bY}[0]{\mathbf{Y}}
\newcommand{\iid}[0]{\overset{iid}\sim}
\newcommand{\Bern}[0]{\text{Bernoulli}}
\newcommand{\ind}[0]{\overset{ind}\sim}
\newcommand{\indep}{\perp \!\!\! \perp}
\newcommand{\eps}[0]{\epsilon}
\newcommand{\MC}[0]{\text{MC}}

\section{Proof of Theoretical Results}
\label{append:proofs}
In the following, denote $\theta_0$ the parameter that identifies the true preference model, $\pi^{opt}$ the true optimal policy within the class of deterministic policies denoted as $\Pi$. Denote $\hat\theta_n$ its estimate obtained by maximizing the data log posterior. 
Denote $P$ the true probability measure that corresponds to the observed data. 
In our setting, we define the Q-function as: 
\[
Q^{\pi}_k(\bH_k, A_k) = \E_{A_{k+1}, \dots, A_K \sim \pi, \bX_{k+1}, \bW_{k+1}, \dots, \bY \sim P_{\theta_0}}[\bE^\top\bY^*(A_1, \dots, A_k, A_{k+1}, \dots, A_K) | \bH_k, A_k].
\]
Recall $(A_1, \dots, A_{k-1}) \subset \bH_k$ so that all actions before $A_{k+1}$ are conditioned. 
Accordingly, denote 
\[V_k^{\pi^{opt}}(\bH_k) 
    = \max_{a_k \in \mathcal{A}_{k}(\bH_k)}Q_k^{\pi^{opt}}(\bH_k, a_k) 
    = \max_{a_k \in \mathcal{A}_{k}(\bH_k)}\mathbb{E}[\mathbf{E}^\top \mathbf{Y}^*|\mathbf{H}_k, a_k].\]
For any $\pi \in \Pi$, the class of deterministic policies,
\[
\widehat{V}_k^{\pi}(\bH_k) 
    = \widehat{Q}_k^{\pi}(\bH_k, \pi_k) 
    = \widehat{\E}[\widehat{V}_{k+1}^{\pi}(\bH_{k+1}) \mid \bH_{k}, \pi_k].
\]
Denote $\|\cdot\|$ the general norm, $\|\cdot\|_{L^\infty(P_{\theta_0})}$ the $L^\infty(P_{\theta_0})$ norm, and $\|\cdot\|_{P_{\theta_0}}$ the $L^2(P_{\theta_0})$ norm, so that for example $\|Q^{\pi^{opt}}(\bH_k, A_k)\|_{P_{\theta_0}} = \mathbb{E}_{A_{k} \sim \mu_k, \bH_k \sim P_{\theta_0}}[Q^{\pi^{opt}}(\bH_k, A_k))]$. We implicitly require $X \in L^2(P_{\theta_0})$ whenever we write $\|X\|_{P_{\theta_0}}$ in the assumption. Denote also $\mu_k(A_k|\bH_k)$ the behavior policy; that is, $\mu_k(A_k|\bH_k) = P(A_k | \bH_k)$.

\vspace*{1cm}
\subsection{LUQ-Learning}
\noindent
We begin with proving proposition 4.1 which justifies LUQ-Learning by showing that, under the true distribution, LUQ-Learning finds $\pi^{opt}$ the true optimal.

\vspace*{1cm}
\noindent
\textbf{Proof of Proposition 4.1:}
\begin{proof}
At $k = K$, we have that for any $\Bar{a}_K \in \otimes_{k=1}^K \mathcal{A}_{k}$,
\begin{align*}
    &\E[\bE^\top \bY^*(\Bar{a}_{K-1}, a_K) | \bX_1, \bW_1, A_1, \bX^*_2(A_1), \bW^*_2(A_1), \dots, \bW^*_K(A_1, \dots, A_{K-1})] \\
    = &\E[\bE^\top \bY(\Bar{a}_{K-1}, a_K) | \bX_1, \bW_1, A_1, \bX_2, \bW_2, \dots, \bW_K] \\
    \leq &\E[\bE^\top \bY(\Bar{a}_{K-1}, \pi^{opt}_K) | \bH_K] = V^{\pi^{opt}}_K(\bH_K).
\end{align*}
The first equality holds because, once $A_1$ conditioned on, $\bW^*_2(a_1) = \bW_2$ and $\bX^*_2(a_1) = \bX_2$, the observed data, and similarly for all $\{\bW^*_k, \bX^*_k\}_{k=3}^{K}$. Finally, the conditioning set includes all $(A_1, \dots, A_K)$, so $\bY^* = \bY$ by (A1) and (A2). The inequality follows from the algorithm that 
\[
\pi^{opt}_K = \argmax_{a_K} \E[\bE^\top \bY(\Bar{a}_{K-1}, a_K) | \bX_1, \bW_1, A_1, \bX_2, \bW_2, \dots, \bW_K]
\]
The last equality comes from the definition of $V^{\pi^{opt}}_K$.
Taking expectation on both sides gives 
\begin{align}
    \E[\bE^\top \bY(\Bar{a}_{K-1}, a_K) | \bH_1] \leq \E[\bE^\top \bY(\Bar{a}_{K-1}, \pi^{opt}_K) | \bH_1] \;\; \forall \Bar{a}_K \in \otimes_{k=1}^K \mathcal{A}_{k}.
    \label{eq:ineqK}
\end{align}
Similarly,
at $k = K - 1, \dots, 1$, for any $\Bar{a}_{k}$,
\begin{align*}
    &\E[\bE^\top \bY(\Bar{a}_{k-1}, a_k, \pi^{opt}_{k+1} \dots, \pi^{opt}_K) | \bX_1, \bW_1, \dots, \bW_k] \\
    \leq &\E[\bE^\top \bY(\Bar{a}_{k-1}, \pi^{opt}_k, \pi^{opt}_{k+1} \dots, \pi^{opt}_K) | \bH_k] = V^{\pi^{opt}}_k(\bH_k), \;\; \text{implying}
\end{align*}
\begin{align}
    \E_{a_{k+1}, \dots, a_K \sim \pi^{opt}}[\bE^\top \bY(\Bar{a}_{k-1}, a_k) | \bH_1] \leq \E[\bE^\top \bY(\Bar{a}_{k-1}, \pi^{opt}_k) | \bH_1], \;\; \forall \Bar{a}_k \in \otimes \mathcal{A}_{k}.
    \label{eq:ineqk}
\end{align}
Consequently, chaining inequalities \ref{eq:ineqK} and \ref{eq:ineqk}, we have
\begin{align*}
    \E[\bE^\top Y(\Bar{a}_K) | \bH_1] 
    \leq \cdots \leq \E[\bE^\top \bY(a_1, \pi^{opt}_2, \dots, \pi^{opt}_{K-1}, \pi^{opt}_K) | \bH_1]
    \leq \E[\bE^\top \bY(\pi^{opt}) | \bH_1],
\end{align*}
completing the proof.
\end{proof}

\vspace*{1cm}
\noindent
To prove Theorem 4.2 regarding convergence in probability of $V(\hat\pi^{opt}_n)$ to $V(\pi^{opt})$, we first prove the following lemmas.
\begin{lemma}
\label{lemma:3.1}
    Assume (A1)-(A5), under LUQ-Learning (Algorithm 1), for any $k = 2, \dots, K$,
    \[
    \widehat{Q}_{k-1}^{\pi^{opt}}(\bH_{k-1}, \pi^{opt}_{k-1})
    -
    \widehat{\E}\!\left[\widehat{V}^{\hat\pi}_{n,k}(\bH_k)\mid \bH_{k-1}, \hat\pi_{k-1}(\bH_{k-1})\right]
    \leq 0
    \quad a.s.
    \]
\end{lemma}
\begin{proof}
\begin{align*}
LHS = & \widehat{\E}[\widehat{V}^{\pi^{opt}_k, \dots, \pi^{opt}_K}(\bH_k) | \bH_{k-1}, \pi^{opt}_{k-1}]
    - \widehat{\E}[\widehat{V}^{\hat\pi_{k}, \dots, \hat\pi_{K}} (\bH_{k}) \mid \bH_{k-1}, \hat{\pi}_{k-1}] \\
    = &\widehat{\E}[\widehat{V}^{\pi^{opt}_k, \dots, \pi^{opt}_K}(\bH_k) | \bH_{k-1}, \pi^{opt}_{k-1}]
    - \widehat{\E}[\widehat{V}^{\pi^{opt}_k, \dots, \hat\pi_K}(\bH_k) | \bH_{k-1}, \pi^{opt}_{k-1}] \\
      & + \widehat{\E}[\widehat{V}^{\pi^{opt}_k, \dots, \hat\pi_K}(\bH_k) | \bH_{k-1}, \pi^{opt}_{k-1}]
    - \widehat{\E}[\widehat{V}^{\pi^{opt}_k, \dots, \hat\pi_{K-1}, \hat\pi_K}(\bH_k) | \bH_{k-1}, \pi^{opt}_{k-1}] \\
      & + \cdots - \cdots \\
      & + \widehat{\E}[\widehat{V}^{\hat\pi_k, \dots, \hat\pi_K}(\bH_k) | \bH_{k-1}, \pi^{opt}_{k-1}]
    - \widehat{\E}[\widehat{V}^{\hat\pi_{k}, \dots, \hat\pi_K} (\bH_{k}) \mid \bH_{k-1}, \hat{\pi}_{k-1}] \\
    = &\widehat{\E}_{A_{k-1}, \dots, A_{K-1} \sim \pi^{opt}}\left[\widehat{\E}[\bE^\top \bY | \bH_{K}, {\pi}^{opt}_K] - \widehat{\E}[\bE^\top \bY | \bH_{K}, \hat{\pi}_K] \mid \bH_{k-1}, A_{k-1} \right] \\
      & + \widehat{\E}_{A_{k-1}, \dots, A_{K-2} \sim \pi^{opt}}\left[\widehat{\E}[\widehat{V}^{\hat\pi_K}(\bH_K) | \bH_{K-1}, \pi^{opt}_{K-1}] - \widehat{\E}[\widehat{V}^{\hat\pi_K}(\bH_K) | \bH_{K-1}, \hat\pi_{K-1}] \mid \bH_{k-1}, A_{k-1} \right] \\
    &+ \dots +
    \widehat{\E}[\widehat{V}^{\hat\pi_{k}, \dots, \hat\pi_{K}} (\bH_{k}) \mid \bH_{k-1}, {\pi}^{opt}_{k-1}] 
     - \widehat{\E}[\widehat{V}^{\hat\pi_{k}, \dots, \hat\pi_{K}} (\bH_{k}) \mid \bH_{k-1}, \hat{\pi}_{k-1}]\\
    \leq &0,
\end{align*}
as each consecutive pair is non-positive a.s. by the definition of $\hat{\pi}_k$ for $k = 2, \dots, K$.
\end{proof}

\vspace*{0.5cm}
\begin{lemma}
\label{lemma:3.2}
{For any $k = 2, \dots, K$, $\pi \in \Pi$, the class of deterministic policy},
suppose $\|\widehat Q^{\pi}_{n,k}(\bH_k,A_k)-Q^{\pi}_k(\bH_k,A_k)\|_{P_{\theta_0}, A_k \sim \mu} = O_p(r_{k})$ for some deterministic sequence $r_k \geq 0$, then  under (A3),  
\begin{align}
    &\left\| \max_{A_k \in \mathcal{A}_{k}}|\widehat Q^{\pi}_{n,k}(\bH_k,A_k)-Q^{\pi}_k(\bH_k,A_k)|\right\|_{P_{\theta_0}} = O_p(r_k) \quad \text{ and }\\
    &\left\|\mathbb E\left[ \left| \max\nolimits_{A_k \in \mathcal{A}_{k}}\widehat Q^{\pi}_{n,k}(\bH_k,A_k)-\max\nolimits_{A_k \in \mathcal{A}_{k}}Q^{\pi}_k(\bH_k,A_k) \right| |\bH_{k-1},A_{k-1}\right]\right\|_{P_{\theta_0}} = O_p(r_k).
\end{align}
\end{lemma}
\begin{proof}
First observe that 
\begin{eqnarray*}
& &\left\|\max_{A_k \in \mathcal{A}_{k}(\bH_k)} \left|\widehat Q^{\pi}_{n,k}(\bH_k,A_k)-Q^{\pi}_k(\bH_k,A_k)\right|\right\|_{P_{\theta_0}}^2 \\
&=& \E_{\bH_k \sim P_{\theta_0}} \left[ \max_{A_k \in \mathcal{A}_{k}(\bH_k)} \left( \widehat{Q}^{\pi}_{n,k}(\bH_k,A_k)-Q^{\pi}_k(\bH_k,A_k) \right)^2 \right] \\
&\leq& \E_{\bH_k \sim P_{\theta_0}}\left[ \sum_{A_k \in \mathcal{A}_k(\bH_k)} \left( \widehat{Q}^{\pi}_{n,k}(\bH_k,A_k)-Q^{\pi}_k(\bH_k,A_k) \right)^2 \right]\\
&\leq& \frac{1}{c}\,\E_{\bH_k \sim P_{\theta_0}}\left[
         \sum_{A_k \in \mathcal{A}_k(\bH_k)}
         \mu_k(A_k\mid\bH_k)
         \left(\widehat Q^{\pi}_{n,k}(\bH_k,A_k)-Q^{\pi}_k(\bH_k,A_k) \right)^2
       \right]\\
&=& \frac{1}{c}\,\E_{\bH_k \sim P_{\theta_0}}\Bigl[
      \E_{A_k\sim\mu_k}\!\left[
        \left(\widehat Q^{\pi}_{n,k}(\bH_k,A_k)-Q^{\pi}_k(\bH_k,A_k)\right)^2
      \,\Big|\,\bH_k\right]
    \Bigr]\\
&=& \frac{1}{c}
    \times
    \mathbb{E}_{\bH_k\sim P_{\theta_0},\, A_k \sim \mu}\!\left[
      \left(\widehat Q^{\pi}_{n,k}(\bH_k, A_k)-Q^{\pi}_k(\bH_k, A_k)\right)^2
    \right],
\end{eqnarray*} 
where $1/{c} = O(1)$ from (A3). Additionally, 
\begin{eqnarray*}
\lefteqn{\left\|\max_{A_k \in \mathcal{A}_{k}}|\widehat Q^{\pi}_{n,k}(\bH_k,A_k)-Q^{\pi}_k(\bH_k,A_k)| \right\|_{P_{\theta_0}}^2}&&\\
&=&\mathbb E\left[\max\nolimits_{A_k} \left| \widehat Q^{\pi}_{n,k}(\bH_k,A_k)-Q^{\pi}_k(\bH_k,A_k)\right|^2\right]\\
&=& \mathbb E\left[\mathbb E\left(\max\nolimits_{A_k} \left| \widehat Q^{\pi}_{n,k}(\bH_k,A_k)-Q^{\pi}_k(\bH_k,A_k)\right|^2\big |\bH_{k-1},A_{k-1}\right)\right] \\
&\geq& \mathbb E\left[\mathbb E\left(\max\nolimits_{A_k}|\widehat Q^{\pi}_{n,k}(\bH_k,A_k)-Q^{\pi}_k(\bH_k,A_k)|\big |\bH_{k-1},A_{k-1}\right)^2\right] \\
&=& \Big\|\mathbb E\left(\max\nolimits_{A_k}|\widehat Q^{\pi}_{n,k}(\bH_k,A_k)-Q^{\pi}_k(\bH_k,A_k)|\big |\bH_{k-1},A_{k-1}\right)\Big\|_{P_{\theta_0}}^2 \\
&\geq& \Big\|\mathbb E\left( \left| \max\nolimits_{A_k}\widehat Q^{\pi}_{n,k}(\bH_k,A_k)-\max\nolimits_{A_k}Q^{\pi}_k(\bH_k,A_k) \right| \big |\bH_{k-1},A_{k-1}\right)\Big\|_{P_{\theta_0}}^2,
\end{eqnarray*}
where the first inequality by Jensen's inequality and the second inequality by $|\max_i a_i - \max_i b_i| \leq \max_i | a_i - b_i |$ for $a_i$, $b_i$ real numbers. 
Combining the two inequalities completes the proof.
\end{proof}

\vspace*{0.5cm}
\begin{lemma}
\label{lemma:3.3}
Assume (A1)-(A5). We show utilizing Lemma \ref{lemma:3.2}, that following LUQ-Learning (Algorithm 1), for any $1 \leq k \leq K$ and any $\pi \in \Pi$,
\begin{align}
    \left\| \max_{A_k \in \mathcal{A}_{k}(\bH_k)} \left| \widehat Q^{\pi}_{n,k}(\bH_k,A_k)-Q^{\pi}_k(\bH_k,A_k) \right| \right\|_{P_{\theta_0}} 
    =
    O_p(r_{E,n} + r_{Y,n} + \sum_{j= k+1}^K r_{V,j,n}).
\end{align}
\end{lemma}
\begin{proof}
We show $\left\| \widehat Q^{\pi}_{n,k}(\bH_k,A_k)-Q^{\pi}_k(\bH_k,A_k)\right\|_{P_{\theta_0}} = O_p(r_{E,n} + r_{Y,n} + \sum_{j={k+1}}^K r_{V,j,n})$ via induction. 
{At k = K}:
\begin{align*}
&\left\|\widehat Q_{n,K}(\bH_K,A_K)-Q_K(\bH_K,A_K)\right\|_{P_{\theta_0}}\\
=&\left\|\widehat{\mathbb E}_{\hat{\theta}_n}[\bE|\bH_K]^\top\widehat{\mathbb E}[\bY|\bH_K,A_K]-{\mathbb E}[\bE|\bH_K]^\top{\mathbb E}[\bY|\bH_K,A_K]\right\|_{P_{\theta_0}}\\
=& \left\|(\widehat{\mathbb E}_{\hat{\theta}_n}[\bE|\bH_K]-\mathbb E[\bE|\bH_K])^\top\widehat{\mathbb E}[\bY|\bH_K,A_K] + \mathbb E[\bE|\bH_K]^\top(\widehat{\mathbb E}(\bY|\bH_K,A_K)-\mathbb E(\bY|\bH_K,A_K))\right\|_{P_{\theta_0}} \\
\leq& \left\|\widehat{\mathbb E}_{\hat{\theta}_n}[\bE|\bH_K]-\mathbb E[\bE|\bH_K] \right\|_{P_{\theta_0}} \left\|\widehat{\mathbb E}[\bY|\bH_K,A_K]\right\|_{L^\infty(P_{\theta_0})} \\
 &+ \Big\|\mathbb E[\bE|\bH_K]\Big\|_{L^\infty(P_{\theta_0})} \left\|\widehat{\mathbb E}(\bY|\bH_K,A_K)-\mathbb E(\bY|\bH_K,A_K)\right\|_{P_{\theta_0}}\\
 = &O_p(r_{E,n} + r_{Y,n}),
\end{align*}
where the last inequality uses Minkowski's inequality. The last equality follows by (V1) and (V2) with $\|\mathbb E[\bE|\bH_K]\|_{L^\infty(P_{\theta_0})} \leq 1$ always as $\bE \in \Delta^{d-1}$. \\

We now complete the proof by induction. Define $\delta_k :=
\|\hat Q^\pi_{n,k}(\bH_k,A_k) - Q^\pi_k(\bH_k,A_k)\|_{P_{\theta_0}}$.
We show that if $\delta_{k+1} = O_p(r_{k+1})$ for some deterministic sequence
$r_{k+1}$, then $\delta_k = O_p(r_{V,k+1,n} + C\,r_{k+1})$,
where $C = \sqrt{1/c} < \infty$ is the constant from (A3):
\begin{align*}
&||\widehat Q^{\pi}_{n,k}(\bH_k, A_k) - Q^{\pi}_k (\bH_k, A_k)||_{P_{\theta_0}}\\
=&\left\|\widehat{\mathbb E}[\max_{A_{k+1}}\widehat Q^{\pi}_{n,k+1}(\bH_{k+1},A_{k+1})|\bH_k,A_k]-{\mathbb E}[\max_{A_{k+1}}Q^{\pi}_{k+1}(\bH_{k+1},A_{k+1})|\bH_k,A_k]\right\|_{P_{\theta_0}}\\
\leq& \left\|\widehat{\mathbb E}[\max_{A_{k+1}}\widehat Q^{\pi}_{n,k+1}(\bH_{k+1},A_{k+1})|\bH_k,A_k]-{\mathbb E}[\max_{A_{k+1}}\widehat Q^{\pi}_{n,k+1}(\bH_{k+1},A_{k+1})|\bH_k,A_k]\right\|_{P_{\theta_0}}\\
&+\left\|{\mathbb E}[\max\nolimits_{A_{k+1}}\widehat Q^{\pi}_{n,k+1}(\bH_{k+1},A_{k+1})|\bH_k,A_k]-\mathbb E[\max\nolimits_{A_{k+1}}Q^{\pi}_{k+1}(\bH_{k+1},A_{k+1})|\bH_k,A_k]\right\|_{P_{\theta_0}}\\
\leq& \left\|\widehat{\mathbb E}[\widehat{V}^{\pi}(\bH_{k+1})|\bH_k,A_k]-{\mathbb E}[\widehat V^{\pi}(\bH_{k+1})|\bH_k,A_k]\right\|_{P_{\theta_0}} \\
&+\left\|{\mathbb E}[ \left| \max_{A_{k+1}}\widehat Q^{\pi}_{n,k+1}(\bH_{k+1},A_{k+1}) - \max_{A_{k+1}}Q^{\pi}_{k+1}(\bH_{k+1},A_{k+1}) \right| |\bH_k,A_k]\right\|_{P_{\theta_0}}.
\end{align*}
The first term is $O_p(r_{V,k+1,n})$ directly by (V3).
For the second term, the second part of Lemma~\ref{lemma:3.2} gives
\begin{align*}
  &\left\|{\mathbb E}[ \left| \max\nolimits_{A_{k+1}}\widehat Q^{\pi}_{n,k+1}(\bH_{k+1},A_{k+1}) - \max\nolimits_{A_{k+1}}Q^{\pi}_{k+1}(\bH_{k+1},A_{k+1}) \right| |\bH_k,A_k]\right\|_{P_{\theta_0}} \\
  \;\leq\;
  &\left\|\max_{A_{k+1}}\left|\hat Q^{\pi}_{n,k+1}(\bH_{k+1},A_{k+1})
    - Q^{\pi}_{k+1}(\bH_{k+1},A_{k+1})\right|\right\|_{P_{\theta_0}}
  \;\leq\; C\,\delta_{k+1} = O_p(C\,r_{k+1}).
\end{align*}
Combining gives $\delta_k = O_p(r_{V,k+1,n} + C\,r_{k+1})$.

Initialising with the base case $\delta_K = O_p(r_{E,n} + r_{Y,n})$
and applying the recursion from $k = K-1$ down to $k = 1$:
\begin{align*}
\delta_{K-1} &= O_p\!\left(r_{V,K,n} + C\,\delta_K\right)
              = O_p\!\left(r_{V,K,n} + C(r_{E,n}+r_{Y,n})\right),\\
\delta_{K-2} &= O_p\!\left(r_{V,K-1,n} + C\,\delta_{K-1}\right)
              = O_p\!\left(r_{V,K-1,n} + C\,r_{V,K,n}
                + C^2(r_{E,n}+r_{Y,n})\right),\\
             &\;\;\vdots\\
\delta_k     &= O_p\!\left(\sum_{j=k+1}^{K} C^{j-k-1}\,r_{V,j,n}
                + C^{K-k}(r_{E,n}+r_{Y,n})\right).
\end{align*}
Since $K < \infty$ is fixed, all powers $C^{K-k} = O(1)$, so
\begin{equation}
  \delta_k
  = O_p\!\left(r_{E,n} + r_{Y,n} + \sum_{j=k+1}^{K} r_{V,j,n}\right)
  \quad\text{for all } 1 \leq k \leq K.
  \label{eq:deltak}
\end{equation}
Applying the first part of Lemma~\ref{lemma:3.2} to \eqref{eq:deltak} with the same constant $C$ gives
\[
  \left\|\max_{A_k\in\mathcal{A}_k(\bH_k)}
    \left|\hat Q^\pi_{n,k}(\bH_k,A_k) - Q^\pi_k(\bH_k,A_k)\right|
  \right\|_{P_{\theta_0}}
  \leq C\,\delta_k
  = O_p\!\left(r_{E,n} + r_{Y,n} + \sum_{j=k+1}^{K} r_{V,j,n}\right),
\]
completing the proof.
\end{proof}

\vspace*{1cm}
\noindent
\textbf{Proof of Theorem 4.2:}
\begin{proof}
For $k = K$,
\begin{align*}
      \|V_{K}(\pi^{opt}) - V_K(\hat{\pi}_n)\|_{P_{\theta_0}}
    &= \left\| \max_{A_K \in \mathcal{A}(\bH_k)} Q_K(\bH_K, A_K) - \mathbb{E}[\bE^\top \bY | \bH_K, \hat{\pi}_{n,K}(\bH_K)] \right\|_{P_{\theta_0}}\\
    &\leq \left\| Q_K(\bH_K, \pi^{opt}_K) - \widehat{\mathbb{E}}[\bE^\top \bY | \bH_K, {\pi}^{opt}_{K}(\bH_K)] \right\|_{P_{\theta_0}} \\
    \quad &+ \left\| \widehat{\mathbb{E}}[\bE^\top \bY | \bH_K, \hat{\pi}_{n,K}(\bH_K)] - \mathbb{E}[\bE^\top \bY | \bH_K, \hat{\pi}_{n,K}(\bH_K)] \right\|_{P_{\theta_0}} \\
    &= O_p(r_{E,n} + r_{Y,n}),
\end{align*}
as the two terms are both $O_p(r_{E,n} + r_{Y,n})$ following from Lemma~\ref{lemma:3.3}. And $ \left(\widehat{\mathbb{E}}[\bE^\top \bY | \bH_K, {\pi}^{opt}_{K}(\bH_K)] - \widehat{\mathbb{E}}[\bE^\top \bY | \bH_K, \hat{\pi}_{n,K}(\bH_K)]\right) \leq 0$ by definition of $\hat\pi_K$.\\

\noindent
For any $1 \leq k \leq K-1$,
\begin{align*}
    &\|V_{k}(\pi^{opt}) - V_k(\hat{\pi}_n)\|_{P_{\theta_0}} \\
    = &\left\| \max_{a_k \in \mathcal{A}(\bH_k)} Q^{\pi^{opt}}_k(\bH_k, a_k) - \E[V_{k+1}^{\hat{\pi}_n}(\bH_{k+1}) | \bH_k, \hat{\pi}_{n,k}(\bH_k)]\right\|_{P_{\theta_0}}\\
    = &\Bigg\| 
        \max_{a_k \in \mathcal{A}(\bH_k)}Q^{\pi^{opt}}_k(\bH_k, a_k) - \widehat{Q}^{\pi^{opt}}_k(\bH_k, \pi^{opt}_k)
        + \widehat{Q}^{\pi^{opt}}_k(\bH_k, \pi^{opt}_k) - \widehat{\E}[\widehat{V}^{\hat\pi}(\bH_{k+1}) \mid  \bH_k, \hat\pi_k(\bH_k)] \\
        & + \widehat{\E}[\widehat{V}^{\hat\pi}(\bH_{k+1}) \mid  \bH_k, \hat\pi_k(\bH_k)] - {\E}[\widehat{V}^{\hat\pi}(\bH_{k+1}) \mid  \bH_k, \hat\pi_k(\bH_k)] \\
        & + {\E}[\widehat{V}^{\hat\pi}(\bH_{k+1}) \mid  \bH_k, \hat\pi_k(\bH_k)] - {\E}[{V}^{\hat\pi}(\bH_{k+1}) \mid  \bH_k, \hat\pi_k(\bH_k)]
        \Bigg\|_{P_{\theta_0}} \\
    \leq &\underbrace{\Big\| 
        Q^{\pi^{opt}}_k(\bH_k, \pi^{opt}_k) - \widehat{Q}^{\pi^{opt}}_k(\bH_k, \pi^{opt}_k)\Big\|_{P_{\theta_0}}}_{A_n} \\
        & + \underbrace{\Big\| \widehat{\E}[\widehat{V}^{\hat\pi}(\bH_{k+1}) \mid  \bH_k, \hat\pi_k(\bH_k)] - {\E}[\widehat{V}^{\hat\pi}(\bH_{k+1}) \mid  \bH_k, \hat\pi_k(\bH_k)]\Big\|_{P_{\theta_0}}}_{B_n} \\
        &  + \underbrace{\Big\| {\E}[\widehat{V}^{\hat\pi}(\bH_{k+1}) \mid  \bH_k, \hat\pi_k(\bH_k)] - {\E}[{V}^{\hat\pi}(\bH_{k+1}) \mid  \bH_k, \hat\pi_k(\bH_k)] \Big\|_{P_{\theta_0}}}_{C_n},
\end{align*}
where the inequality follows by Minkowski's inequality, after dropping the cross term $\widehat{Q}^{\pi^{opt}}_k(\bH_k,\pi^{opt}_k) - \widehat{\E}[\widehat{V}^{\hat\pi}_{n,k+1}\mid\bH_k,\hat\pi_k]\leq 0$ a.s.\ by Lemma~\ref{lemma:3.1}.
By Lemma \ref{lemma:3.3} with $\pi = \pi^{opt}$, $A_n = O_p(r_{E,n} + r_{Y,n} + \sum_{j=k+1}^K r_{V,j,n})$. By (V3) and (A3),
\begin{align*}
B_n \leq \Big\| \max_{A_k \in \mathcal{A}_{k}(\bH_k)} \left| \widehat{\E}[\widehat{V}^{\hat\pi}(\bH_{k+1}) \mid  \bH_k, A_k] - {\E}[\widehat{V}^{\hat\pi}(\bH_{k+1}) \mid  \bH_k, A_k] \right| \Big\|_{P_{\theta_0}} 
= O_p(r_{V,k+1,n}).
\end{align*}
Finally, 
\begin{align*}
C_n
&\leq
\left\|\max_{A_k \in \mathcal{A}_k(\bH_{k})}\,\E\!\left[\max_{A_{k+1} \in \mathcal{A}_{k+1}(\bH_{k+1})}
  \bigl|\hat Q^{\hat{\pi}}_{n,k+1}(\bH_{k+1},A_{k+1})
  - Q^{\hat{\pi}}_{k+1}(\bH_{k+1},A_{k+1})\bigr|\,\Big|\,\bH_k,A_k\right]
\right\|_{P_{\theta_0}}\\
&\leq
\left\|\max_{A_{k+1} \in \mathcal{A}_{k+1}(\bH_{k+1})}
  \bigl|\hat Q^{\hat{\pi}}_{n,k+1}(\bH_{k+1},A_{k+1})
  - Q^{\hat \pi}_{k+1}(\bH_{k+1},A_{k+1})\bigr|
\right\|_{P_{\theta_0}} \\
&= O_p(r_{E,n} + r_{Y,n} + \sum_{j=k+2}^K r_{V,j,n})
\end{align*}
by Jensen's inequality, the tower property, (A3), and Lemma~\ref{lemma:3.3} again with $\pi = \hat{\pi}$.
Take $k = 1$ and combining the rates completes the proof.
\end{proof}

\vspace*{1cm}
\noindent
We focus on proving Proposition 5.1, which shows that under the proposed model for the BEST study, $\hat\theta_n$ is consistent and asymptotically normal under regularity conditions, and that $\|\hat{\mathbb{E}}[\mathbf{E}|\mathbf{H}_2; \hat\theta_n] - {\mathbb{E}}[\mathbf{E}|\mathbf{H}_2; \theta_0]\|_{P(\theta_0)} \to 0$, providing justification for $V(\hat\pi^{opt}_n) - V(\pi^{opt}) \to_p 0$. We first show Lemma 1.4. which is a general M-estimation consistency and asymptotic normality result for $\hat{\theta}_n$.\\

\begin{lemma}
\label{lemma_theta}
Let $\hat\theta_n = \argmax_{\theta \in \Theta}\left\{\sum_{i=1}^n \log P(\bH_{i,K+1}; M_\theta) + \log \Lambda_\theta(\theta)\right\}$ be the estimator, where $\Lambda_\theta$ denotes the prior density on $\theta$. 
Then
$\hat\theta_n\to_p \theta_0$ provided: 
(C1) $\exists \theta_0 $ an interior point of compact $\Theta$ such that $P(\mathbf H_{K+1}|\mathbf E) = M_{\theta_0}(\mathbf H_{K+1}, \mathbf E)g(\mathbf H_{K+1})$, with $M_{\theta}$ the $\theta$-dependent likelihood part and $g$ some non-negative measurable function bounded from above; 
(C2) $M_\theta(\mathbf H_{K+1}, \mathbf E)$ is continuous in $\theta$ for a.s. $(\mathbf H_{K+1}, \mathbf E)$; 
(C3) $\forall \theta$, $|M_\theta(\mathbf H_{K+1}, \mathbf E)|\leq F_1(\mathbf H_{K+1}, \mathbf E)$ for some $F_1$ satisfying $\E_{\theta_0, \mathbf E, \mathbf H_{K+1}}[F_1(\mathbf H_{K+1}, \mathbf E)] < \infty$; 
(C4)$\exists\,c>0$ such that $P(\bH_{K+1};\,M_{\theta}) = \int_{\mathcal{E}} M_\theta(\bH_{K+1},\bE)\,g(\bH_{K+1})\,dP(\bE) > c$ a.s.\ in $\bH_{K+1}$, uniformly over $\theta\in\Theta$;
(C5) $P(\mathbf H_{K+1}; M_{\theta_0})\neq P(\mathbf H_{K+1};M_\theta)$ for all $\theta\neq\theta_0$;
(C6) The prior satisfies $\Lambda_\theta(\theta_0)>0$, and $\sup_{\theta\in\Theta}\log\Lambda_\theta(\theta)<\infty$.

Moreover, $\sqrt{n}(\hat\theta_n-\theta_0)\to_d \mathcal N (0,I(\theta_0)^{-1})$, provided that in addition to the above:
(N1) $I(\theta_0)$ is non-singular; 
(N2) For some fixed $\epsilon > 0$, $\forall \theta_1,\theta_2\in \mathcal{N}_\epsilon(\theta_0)=\{\theta \in \Theta:||\theta-\theta_0|| < \epsilon\}$, $|M_{\theta_1}(\mathbf H_{K+1}, \mathbf E)-M_{\theta_2}(\mathbf H_{K+1}, \mathbf E)|\leq F_2(\mathbf H_{K+1},\mathbf E)\|\theta_1-\theta_2\|$ for some measurable function $F_2$ with $\mathbb E_{\theta_0,\mathbf E}[F_{2}^2(\mathbf H_{K+1},\mathbf E)]<\infty$ a.s. in $\mathbf H_{K+1}$;
(N3) $M_{\theta}(\mathbf H_{K+1} , \mathbf E)$ is continuously differentiable in $\theta$ for a.s. $\mathbf E$, with $||\nabla_\theta M_{\theta}(\mathbf H_{K+1},\mathbf E)||_{L^\infty(P_{\theta_0})} < G(\mathbf H_{K+1},\mathbf E)$ for some measurable function $G$ satisfying $\mathbb E_{\theta_0,\mathbf E}[G^2(\mathbf H_{K+1},\mathbf E)]<\infty$ a.s. in $\mathbf H_{K+1}$;
(N4) $\Lambda_\theta$ is positive and continuously differentiable on $\mathcal{N}_\epsilon(\theta_0)$ for the same fixed $\epsilon>0$ as in (N2).
\end{lemma}

\vspace*{1cm}
\noindent
\textbf{Proof of Lemma 4.1:}
\begin{proof}
Denote the full MAP objective and its population counterpart by
\[
  \widetilde L_n(\theta)
  = L_n(\theta) + R(\theta),
  \qquad
  L(\theta) = \E\bigl[\log P(\bH_{K+1};\,M_\theta)\bigr],
  \text{ where }
\]
$L_n(\theta)=\sum_{i=1}^n\log P(\bH_{i,K+1};\,M_\theta)$,
$P(\bH_{i,K+1};\,M_\theta) =\int_{\mathcal{E}} M_{\theta}(\bH_{i,K+1},\bE)\,g(\bH_{i,K+1})\,dP(\bE_i)$, and $R(\theta)=\log\Lambda_\theta(\theta)$.
By (C6), $\sup_{\theta\in\Theta}R(\theta)<\infty$, so
$\frac{1}{n}\sup_{\theta\in\Theta}|R(\theta)|\to 0$. Therefore,
\begin{equation}
  \label{eq:prior_vanishes}
  \sup_{\theta\in\Theta}
  \Bigl|\tfrac{1}{n}\widetilde L_n(\theta) - \tfrac{1}{n}L_n(\theta)\Bigr|
  = \tfrac{1}{n}\sup_{\theta\in\Theta}|R(\theta)| \to 0,
\end{equation}
so the MAP $\tilde{L}_n$ and likelihood objectives $L_n$ share the same population limit
$L(\theta)$.
By Theorem 5.7 of \citet{Vaart1998}, $\hat\theta_n\to_p \theta_0$ provided: (A1) $\widetilde L_n(\hat\theta_n)\geq \widetilde L_n(\theta_0)- o_p(1)$; (A2) $\sup_{\theta:d(\theta,\theta_0)\geq \eps}L(\theta)<L(\theta_0)$ for all $\eps>0$; and (A3) $\sup_{\theta\in\Theta}|\frac{1}{n}\widetilde{L}_n(\theta)-L(\theta)|\to_p 0$.  

Condition (A1) is satisfied by $\hat\theta_n$ being the maximizer of $\widetilde L_n(\theta)$. 
By (C1),
$
  P(\bH_{K+1}) = \int_{\mathcal{E}} P(\bH_{K+1}|\bE)\,dP(\bE)
  = \int_{\mathcal{E}} M_{\theta_0}(\bH_{K+1},\bE)\,g(\bH_{K+1})\,dP(\bE)
  = P(\bH_{K+1};\,M_{\theta_0}),
$
so the true data-generating distribution corresponds to model $M_{\theta_0}$.
By (C3), for any $\theta$,
$P(\bH_{K+1};\,M_\theta)
 = \int_{\mathcal{E}} M_\theta(\bH_{K+1},\bE)\,g(\bH_{K+1})\,dP(\bE)
 \leq \E_\bE[F_1(\bH_{K+1},\bE)] \cdot \sup g < \infty$
a.s., so $L(\theta)$ is finite. By (C2) and dominated convergence (using $F_1$ as envelope), $P(\bH_{K+1};\,M_\theta)$ is continuous in $\theta$ a.s., and by (C4), $\log P(\bH_{K+1};\,M_\theta) > \log c > -\infty$, so $L(\theta) = \E[\log P(\bH_{K+1};\,M_\theta)]$ is continuous in $\theta$ by another application of dominated convergence. By (C1) and (C5), Lemma~2.1 gives that $L(\theta)$ is uniquely maximized at $\theta_0$. Since $L(\theta)$ is continuous and $\Theta$ is compact, condition (A2) follows from Problem~5.27 of \citet{Vaart1998}.
Finally, we apply the Glivenko--Cantelli theorem to the class
$\mathcal{F} = \{\log P(\bH_{K+1};\,M_\theta) : \theta\in\Theta\}$ to verify (A3).
We first verify the envelope condition. By (C4), $P(\bH_{K+1};\,M_\theta) > c > 0$ uniformly over $\theta\in\Theta$ a.s., so $\log P(\bH_{K+1};\,M_\theta) > \log c > -\infty$ uniformly.
By (C3) and the boundedness of $g$ in (C1), $P(\bH_{K+1};\,M_\theta) \leq (\sup g)\,\E_\bE[F_1(\bH_{K+1},\bE)]$ a.s., so $\log P(\bH_{K+1};\,M_\theta)$ is bounded above by an integrable function under $P_{\theta_0}$. Hence $\mathcal{F}$ admits an integrable envelope under $P_{\theta_0}$. Combined with almost-sure continuity of $\log P(\bH_{K+1};\,M_\theta)$ in
$\theta$ established above and compactness of $\Theta$, Example~19.8 of \citet{Vaart1998} gives that $\mathcal{F}$ is a $P_{\theta_0}$-Glivenko--Cantelli class. So
$
  \sup_{\theta\in\Theta}
  \Bigl|\tfrac{1}{n}L_n(\theta) - L(\theta)\Bigr| \to_p 0.
$
Combining with \eqref{eq:prior_vanishes} via the triangle inequality,
\[
  \sup_{\theta\in\Theta}
  \Bigl|\tfrac{1}{n}\widetilde{L}_n(\theta) - L(\theta)\Bigr|
  \leq
  \sup_{\theta\in\Theta}\Bigl|\tfrac{1}{n}L_n(\theta) - L(\theta)\Bigr|
  + \tfrac{1}{n}\sup_{\theta\in\Theta}|R(\theta)|
  \to_p 0,
\]
verifying (A3).

By Theorem 5.39 of \citet{Vaart1998}, $\sqrt{n}(\hat\theta_n-\theta_0)\to_d\mathcal \mathcal{N}(0,I(\theta_0)^{-1})$ provided that: (B1) $\hat\theta_n\to_p \theta_0$; (B2) $I(\theta_0)$ is non-singular; (B3) $\log P(\bH_{K+1}; M_{\theta})$ is Lipschitz continuous in the neighborhood of $\theta_0$ with some Lipschitz constant $F_3(\bH_{K+1})$ square-integrable; and (B4) $P(\bH_{K+1}; M_{\theta})$ is Hellinger differentiable. In addition, we verify that $R(\theta)$ does not alter the asymptotic score equation.

(B1) is satisfied by conditions (C1)-(C5) from the reasoning above. 
(B2) is satisfied by assumption.
By (N2) and linearity of integration, 
$|P(\bH_{K+1};\,M_{\theta_1}) - P(\bH_{K+1};\,M_{\theta_2})| = \left|\int_{\mathcal{E}}(M_{\theta_1}-M_{\theta_2})\,g(\bH_{K+1})\,dP(\bE)\right| 
\leq (\sup g)\,\mathbb{E}_\bE[F_2(\bH_{K+1},\bE)]\,\|\theta_1-\theta_2\|.$ By (C4), $\frac{d}{dx} \log(x)$ with $x = M_\theta(\bH_{K+1})$ is upper bounded by $1/c$ on $\{x > c\}$. As the composition of Lipschitz continuous functions are also Lipschitz continuous with the Lipschitz constant being the product of those of the composing functions \citep{Shwartz2014}, $|\log P(\bH_{K+1}; M_{\theta_1}) - \log P(\bH_{K+1}; M_{\theta_2})|\leq \frac{1}{c}\mathbb{E}_\bE[F_2(\bH_{K+1},\bE)]\|\theta_1-\theta_2\|$ with $\frac{\sup g}{c}\,\mathbb{E}_{\theta_0,\bE}[F_2(\bH_{K+1},\bE)]<\infty$, where finiteness follows from $\mathbb{E}_{\theta_0,\bE}[F_2^2]<\infty$ by (N2) and Jensen's inequality. So condition (B3) holds. 

By (N3), we have $\mathbb E_{E}[G(\bH_{K+1}, \bE)]<\infty$ almost surely and thus by the Leibniz integral theorem, $\nabla_\theta P(\bH_{K+1}; M_\theta)= g(\bH_{K+1}) \mathbb{E}_\bE [\nabla_\theta M_\theta(\bH_{K+1},\bE)]$, and we have by the dominated convergence theorem that $\nabla_\theta P(\bH_{K+1}; M_\theta)$ is continuous. 
$
\nabla_\theta \sqrt{P(\bH_{K+1};\,M_\theta)}
= \frac{\nabla_\theta P(\bH_{K+1};\,M_\theta)}{2\sqrt{P(\bH_{K+1};\,M_\theta)}}
= \frac{g(\bH_{K+1})\,\mathbb{E}_\bE[\nabla_\theta M_\theta(\bH_{K+1},\bE)]}
       {2\sqrt{P(\bH_{K+1};\,M_\theta)}}.
$ 
By (C4) $P(\bH_{K+1}; M_\theta) > c > 0$ so the quantity is well-defined; then by (N3), $\nabla_\theta M_\theta(\bH_{K+1}, \bE)$ and $M_\theta(\bH_{K+1}, \bE)$ are both continuous, so $\nabla_\theta \sqrt{P(\bH_{K+1}; M_\theta)}$ is also continuous. 
Finally, under our assumptions, the Fisher information
\begin{align}
I(\theta)=\mathbb E_{\theta_0}\left[\frac{g(\bH_{K+1})^2}{P(\bH_{K+1}; M_\theta)^2}\mathbb{E}_\bE [\nabla_\theta M_\theta(\bH_{K+1}, \bE)]\mathbb{E}_\bE [\nabla_\theta M_\theta(\bH_{K+1}, \bE)]^\top\right]
\label{eq:fisher}
\end{align}
has each element bounded by $(\frac{C}{c})^2\mathbb{E}_\bE^2[G(\bH_{K+1},\bE)]\leq (\frac{C}{c})^2\mathbb{E}_\bE[G^2(\bH_{K+1},\bE)]$ where $C < \infty$ the upper bound of $g$ and $c > 0$, so $(\frac{C}{c})^2\mathbb E_{\bE,\theta_0}[G^2(\bH_{K+1},\bE)]<\infty$. 
By dominated convergence theorem once more, $I(\theta)$ is continuous. As $\sqrt{P(\bH_{K+1}; M_\theta)}$ is continuously differentiable and $I(\theta)$ is continuous, we have by Lemma 7.6 of \citet{Vaart1998} that $P(\bH_{K+1}; M_\theta)$ is Hellinger differentiable, satisfying condition (B4). 

Finally, since $\hat\theta_n \to_p \theta_0$ and $\theta_0$ is an interior point of
$\Theta$ by (C1), $\hat\theta_n$ lies in the interior of $\Theta$ with
probability tending to 1. On this event, $\hat\theta_n$ is an interior
maximizer of $\widetilde{L}_n(\theta)$, so the first-order condition 
$
  \nabla_\theta L_n(\hat\theta_n) + \nabla_\theta R(\hat\theta_n) = 0
$
holds.
By (N4), $\Lambda_\theta$ is continuously differentiable on
$\mathcal{N}_\eps(\theta_0)$, so $\nabla_\theta\log\Lambda_\theta$ exists and
is bounded on this neighborhood. Hence
$\nabla_\theta R(\hat\theta_n) = \nabla_\theta\log\Lambda_\theta(\hat\theta_n)
= O(1)$. Since $\nabla_\theta L_n(\theta_0) = O_p(\sqrt{n})$ by the central limit
theorem, 
$
  \frac{1}{\sqrt{n}}\nabla_\theta L_n(\hat\theta_n)
  + \frac{1}{\sqrt{n}}\nabla_\theta R(\hat\theta_n) = 0,
$ where $
  \frac{1}{\sqrt{n}}\nabla_\theta R(\hat\theta_n) = O(n^{-1/2}) = o_p(1).
$
Therefore the regularization term contributes $o_p(1)$ to the normalized score
equation, and a standard Taylor expansion argument around $\theta_0$ yields
$\sqrt{n}(\hat\theta_n-\theta_0)\to_d\mathcal{N}(0,\,I(\theta_0)^{-1})$,
with the Fisher information $I(\theta_0)$ defined the same as equation \ref{eq:fisher} evaluated at $\theta = \theta_0$.
\end{proof}

Most of the above conditions can be verified directly using the proposed model $M_\theta(\mathbf{H}_{K+1},\mathbf{E})$, without worrying about the integral $P(\mathbf{H}_{K+1}; \theta) = \int M_\theta(\mathbf{H}_{K+1},\mathbf{E})dP(\mathbf{E})$ whose close form is often difficult to obtain.
Condition (C4) is related to $P(\mathbf H_{K+1};M_\theta)$, but can usually be easily verified. For example, if preference $\mathbf W$ lies in a compact space, then combined with $\Theta$ compact, one can derive a lower bound for $\min_{\mathbf E}M_\theta(\mathbf H_{K+1}, \mathbf{E})$, then show that the lower bound is strictly away from 0 on some non-trivial sets in $\mathcal{E}$.
Conditions (C5) and (N1), however, cannot be easily reduced to the corresponding conditions on $M_\theta(\mathbf{H}_{K+1},\mathbf{E})$, though it is standard to assume non-singularity and identifiability when deriving theoretical results for latent variable models (\cite{McCullagh1989} and \cite{Breslow1993, Bianconcini2014, Butler2018}, respectively). 

\vspace*{1cm}
\noindent
\textbf{Proof of Proposition 5.1:}
\begin{proof}
To show that $\hat\theta_n \to_{p}\theta_0$, by Lemma 4.1, we require 
(C1) $P(\bH_{K+1}|\bE)=M_{\theta_0}(\bH_{K+1},\bE)g(\bH_{K+1})$ for some interior point $\theta_0\in\Theta$ compact and $g$ bounded from above; (C2) $M_\theta(\bH_{K+1},\bE)$ continuous in $\theta$; 
(C3) $\forall \theta$, $|M_\theta(\bH_{3}, \bE)|\leq F_1(\bH_{3}, \bE)$ for some $F_1$ integrable; 
(C4) $\exists c>0$ such that the measure induced by model $M_{\theta}$, $P(\bH_{3}; M_{\theta}) > c$ a.s. in $\bH_{3}$; 
(C5) $P(\bH_{3}; M_{\theta_0})\neq P(\bH_{3};M_\theta)$ for all $\theta\neq\theta_0$.
(C6) The prior satisfies $\Lambda_\theta(\theta_0)>0$, and $\sup_{\theta\in\Theta}\log\Lambda_\theta(\theta)<\infty$.

Identifiability assumptions (C1) and (C5) are assumed, and $\theta = (\alpha, \beta, \lambda) \in \Bar{\mathbb{R}}^d$ which is closed and bounded and thus we have $\Theta$ compact. 
Also, the proposed model $P(\bW^B_1|\bV,\beta)$, $P(\bW_1^{R}|\bV,\lambda)$, $P(\mathbf{W}^{Sat}_2|\bE^\top\bX_2,\alpha)$, $P(\bW^B_2|\bV,\beta)$, $P(\bW_2^{R}|\bV,\lambda)$ and $P(\mathbf{W}^{Sat}_3|\bE^\top\bY,\alpha)$ detailed in Section 5.1 in the manuscript are all continuous w.r.t. the parameters. Let $M_\theta(\bH_{3},\bE)$ be the product of these terms and note that $P(\bH_{3}; M_\theta)= \mathbb{E}_\bE[M_\theta(\bH_{3},\bE)]g(\bH_{3})$ and $P_\theta(\bH_{3}|\bE)=M_\theta(\bH_{3},\bE)g(\bH_{3})$ As the product of continuous functions is continuous, (C2) is satisfied. Moreover, $(\bW^B_k,\bW^R_k,\bW^{Sat}_{t+1})_{1\leq  t \leq 2}$ all categorical implying $M_\theta(\bH_{3},\bE) \leq 1$ pointwise, so (C3) is satisfied. (C6) is also satisfied, as the normal prior $\Lambda_\theta$  is strictly positive at $\theta_0$ and $\sup_{\theta\in\Theta}\log\Lambda_\theta(\theta) < \infty$ by compactness of $\Theta$ and continuity of $\log\Lambda_\theta$.

It remains to show (C4). We have assumed $\exists C < \infty$ such that $||\theta||_{L^\infty(P_{\theta_0})}\leq C$ and some small $\eps>0$ such that $\alpha_{k,\cdot,1},\lambda_k,\alpha_{k,j+1,0}-\alpha_{k,j,0}\geq\eps$, for all $\theta \in \Theta$. Then it can be seen that for all $\mathbf{V} \in\mathbb{R}^2$, $\min_{\theta,\bH_{3}\in\Theta\times\mathcal H_{3}}P(W^{B}_{k,j}|\bV,\theta) \geq \min\{\sigma(-C-C\sum_{j=1,2}V_j), 1-\sigma(C+C\sum_{j=1,2}V_j)\} = 1 - \sigma[C(\sum_{j=1,2}V_j + 1)], \forall (t,k)\in\{1,2\}\times\{1,\dots,12\}$; $\min_{\theta,\bH_{3}\in\Theta\times\mathcal H_{3}}P(\bW_t^R|\bE^R,\theta)\geq \exp(-3C)/6 \stackrel{\text{denote}}{=} C_1 >0$; and $\min_{\theta,\bH_{3}\in\Theta\times\mathcal H_{3}}P(\bW^{Sat}_{k}|\bE^\top\bX_{k})\geq \min\{1-\sigma(C),\sigma(\eps-10 C),\sigma(C)-\sigma(C-\eps)\} \stackrel{\text{denote}}{=} C_2>0$ for $k = 2,3$. Combined with the assumption that $\min_{\bH_{3}\in\mathcal H_{3}} g(\bH_{3}) > c > 0$, we have that
$P(\bH_3;M_\theta) 
\geq c K^2_1 K^2_2 \int_{\mathbb{R}^2}\left[1 - \sigma(C(\sum_{j=1,2}V_j+1))\right]^{24}dP(\bV) 
\geq c K^2_1 K^2_2 \{ 1 - \int_{\mathbb{R}} \sigma\left[C(Z + 1)\right] dP(Z)\}^{24} 
> 0$, 
where $Z \sim \mathcal{N}(0, 2)$ as $\bV \sim N_2(0, \mathbf{I})$; the second inequality follows from Jensen's inequality and the last inequality follows from $\sigma(x) < 1$ for any finite $x$, $C<\infty$, and $Z$ continuous so its value equaling infinity is of measure zero. Thus (C4) is also satisfied.

We now show $||\widehat{\mathbb E}[\bE|\bH_2;\hat{\theta}_n]-\mathbb E[\bE|\bH_2;\theta_0]||_{P_{\theta_0}} \to 0$. 
Let $\bH_{3,D}=(\bW^B_1,\bW^B_2,\bW_1^R,\bW_2^R,\\\bW_2^{Sat},\bW_3^{Sat})$ be the components of $\bH_{K+1}$ dependent on $\bE$ and $\bH_{3,I}=(\bX_1,\bX_2,A_1,A_2,\bY)$ be the components conditionally independent of $\bE$. Note that $\bH_{3}=\bH_{3,D}\cup \bH_{3,I}$, and $M_\theta(\bH_{K+1},\bE)$ is a function of only $\bE$ and $\bH_{3,D}$. 
At any $\theta \in \Theta$, $\widehat\E[\bE|\bH_2;\theta] = \frac{1/N_{sim}\sum_{b=1} \bE^{(b)} P(\bH_2 | \bE^{(b)}; \theta)}{1/N_{sim}\sum_{b=1} P(\bH_2 | \bE^{(b)}; \theta)}$. Applying the strong law of large numbers for both the numerator and denominator and the continuous mapping theorem gives $\widehat\E[\bE|\bH_2;\theta] \to_{a.s.} \E[\bE|\bH_2;\theta]$ as $N_{sim} \to \infty$. Combined with the assumption that $\|\E[\bE | \bH_2; \theta]\|_{L^\infty(P_{\theta_0})} < \infty$, we have that 
$\| \widehat\E[\bE|\bH_2;\theta] - \E[\bE|\bH_2;\theta]\|_{P_{\theta_0}} \to 0$.
Also, $\mathbb E_{\theta}[\bE|\bH_2]=\frac{\int_{\mathcal E}\bE P_{\theta}(\bH_2|\bE)dP(\bE)}{\int_{\mathcal E}P_{\theta}(\bH_2|\bE)dP(\bE)}=\frac{\int_{\mathcal E}\bE M_{\theta}(\bH_2,\bE)dP(\bE)}{\int_{\mathcal E}M_{\theta}(\bH_2, \bE)dP(\bE)}$, $M_{\theta}(\bH_2, \bE)=\sum_{\bH_{3,D}\in\mathcal H_{3,D}(\bH_{2,D})}M_\theta(\bH_{3}, \bE)$, $\mathcal H_{3,D}(\bh_{2,D})$ is the set of $\bH_{3,D}$ where $\bH_{2,D}=\bh_{2,D}$ and $\bH_{2,D}=(\bW^B_1,\bW^B_2,\bW_1^R,\bW_2^R,\bW^{Sat}_2)$. This is a finite sum, and each element $M_\theta(\bH_{3},\bE)$ of this sum is continuous in $\theta$. Therefore, $M_{\theta}(\bH_2, \bE)$ is continuous in $\theta$. As $M_\theta(\bH_2, \bE) < 1$, we have that both the numerator and the denominator are continuous in $\theta$ by the dominated convergence theorem, and that $\E[M_\theta(\bH_2, \bE)] > 0$ for any $\theta$ a.s. in $\bH_2$. Thus $\E_\theta[\bE|\bH_2]$ continuous in $\theta$, and by the continuous mapping theorem, $\mathbb E_{\hat\theta_n}[\bE|\bH_2] \to_p \mathbb E_{\theta}[\bE|\bH_2]$ as $n \to \infty$. Note that $||\widehat{\mathbb E}[\bE|\bH_2]-\mathbb E[\bE|\bH_2]||_{P_{\theta_0}}^2=\sum_{\bH_2\in\mathcal H_{2,D}}\left[\left(\mathbb E_{\hat\theta_n}[\bE|\bH_{2,D}]-\mathbb E[\bE|\bH_{2,D}]\right)^2\right] P(\bH_{2,D})$ is a finite sum. Thus $\|\E[\bE|\bH_2;\hat\theta_n] - \E[\bE|\bH_2;\theta_0]\|_{P_{\theta_0}} = o(1)$, implying 
$\|\widehat{\mathbb E}[\bE|\bH_2;\hat{\theta}_n]-\mathbb E[\bE|\bH_2;\theta_0]\|_{P_{\theta_0}} 
\leq \|\widehat{\mathbb E}[\bE|\bH_2;\hat{\theta}_n]-\mathbb{E}[\bE|\bH_2; \hat{\theta}_n]\|_{P_{\theta_0}} 
+ \|{\mathbb E}[\bE|\bH_2;\hat{\theta}_n]-\mathbb E[\bE|\bH_2;\theta_0]\|_{P_{\theta_0}} 
\to 0$ as $N_{sim}, n \to \infty$. 

To show that $\sqrt{n}(\hat\theta_n - \theta_0)\to_d \mathcal N (0,I(\theta_0)^{-1})$, by Lemma 4.1, as we have shown $\hat\theta_n = \theta + o_p(1)$, it remains to have: 
(N1) $I(\theta_0)$ non-singular; 
(N2) $\forall \theta_1,\theta_2\in \mathcal{N}_\eps(\theta_0)=\{\theta \in \Theta:||\theta-\theta_0||_2<\eps\}$, with any $\eps > 0$, $|M_{\theta_1}(\bH_{3}, \bE)-M_{\theta_2}(\bH_{3}, \bE)|\leq F_2(\bH_{3},\bE)\|\theta_1-\theta_2\|_2$ for some measurable function $F_2$ satisfying $\mathbb E_{\theta_0,\bE}[F_{2}^2(\bH_{3},\bE)]<\infty$ a.s. in $\bH_{3}$;
(N3) $M_{\theta}(\bH_{3} , \bE)$ is continuously differentiable in $\theta$ for a.s. $\bE$ with $||\nabla_\theta M_{\theta_1}(\bH_{3},\bE)||_{L^\infty(P_{\theta_0})} < G(\bH_{3},\bE)$ for some measureable function $G$ satisfying $\mathbb E_{\theta_0,\bE}[G^2(\bH_{3},\bE)]<\infty$ a.s. in $\bH_{3}$; and
(N4) $\Lambda_\theta$ is positive and continuously differentiable on $\mathcal{N}_\epsilon(\theta_0)$ for the same fixed $\epsilon>0$.

(N1) is satisfied by our assumption. 
(N4) is satisfied since the normal density is strictly positive and
infinitely differentiable on all of $\mathbb{R}^{|\theta|}$, and in
particular on $\mathcal{N}_\epsilon(\theta_0)$ for any fixed $\epsilon > 0$.
To prove the remaining conditions, we need to derive the gradient of the log-likelihood. We can derive the relevant quantities in closed form on the basis of our proposed data generating process. Specifically:
\small
\begin{align*}
P(W^B_{1,j}|\bV,\beta)=&\sigma(\beta_{1,j,0}+\beta_{1,j,1}^\top\bV)^{W^B_{1,j}}(1-\sigma(\beta_{1,j,0}+\beta_{1,j,1}^\top\bV))^{1-W^B_{1,j}},\\
\nabla_{\beta_{1,j,0}}P(W^B_{1,j}|\bV,\beta)=&P(W_{1,j}^B | \mathbf{V}, \beta) (W_{1,j}^B - \sigma(\beta_{1,j,0} + \beta_{1,j,1}^\top \mathbf{V})),\\
\nabla_{\beta_{1,j,1}} P(W_{1,j}^B | \mathbf{V}, \beta) =
&P(W_{1,j}^B | \mathbf{V}, \beta) (W_{1,j}^B - \sigma(\beta_{1,j,0} + \beta_{1,j,1}^\top \mathbf{V})) \mathbf{V}.\\
\nabla_{\lambda_1}P(\bW_1^R|\bV,\lambda)=& \frac{\left[\sum_{\bv\in\textit{Perm}}\exp(-\lambda_1 T(\bv,\bE^R))(T(\bv,\bE^R)-T(\bW_1^R,\bE^R)\right]}{\left(\sum_{\bv\in\textit{Perm}}\exp(-\lambda_1 T(\bv,\bE^R))\right)^2}\exp(-\lambda_1 T(\bW_1^R,\bE^R)), \\
P(\bW^{Sat}_2 | \bE^\top\bX_2, \alpha)=&\sigma(\alpha_{2,\bW^{Sat}_2,0}-\alpha_{2,\cdot,1}\bE^\top\bX_2)^{1-I(\bW^{Sat}_2=7)}\\
&-I(\bW^{Sat}_2\neq 1)\sigma(\alpha_{2,\bW^{Sat}_2-1,0}-\alpha_{2,\cdot,1}\bE^\top\bX_2), \\
\nabla_{\alpha_{2,j,0}}P(\bW^{Sat}_2 | \bE^\top\bX_2, \alpha)=& I(\bW^{Sat}_2=j)\sigma'(\alpha_{2,j,0}-\alpha_{2,\cdot,1}\bE^\top\bX_2)-I(\bW^{Sat}_2= j-1 )\sigma'(\alpha_{2,j,0}-\alpha_{2,\cdot,1}\bE^\top\bX_2)\\
=& \left[I(\bW^{Sat}_2 = j)- I(\bW^{Sat}_2 = j-1)\right]\sigma'(\alpha_{2,j,0}-\alpha_{2,\cdot,1}\bE^\top\bX_2), \\
\nabla_{\alpha_{2,\cdot,1}}P(\bW^{Sat}_2 | \bE^\top\bX_2, \alpha)
=& -\Big[I(\bW^{Sat}_2\neq 7)\sigma'(\alpha_{2,\bW^{Sat}_2,0}-\alpha_{2,\cdot,1}\bE^\top\bX_2) \\
&+I(\bW^{Sat}_2\neq 1)\sigma'(\alpha_{2,\bW^{Sat}_2-1,0}-\alpha_{2,\cdot,1}\bE^\top\bX_2)\Big](\bE^\top\bX_2).
\end{align*}
\normalsize
Similarly,
\small
\begin{align*}
\nabla_{\beta_{2,j,0}}P(W^{B}_{2,j}|\bV,\beta)=&P(W_{2,j}^B | \mathbf{V}, \beta) (W_{2,j}^B - \sigma(\beta_{2,j,0} + \beta_{2,j,1}^\top \mathbf{V})), \\
\nabla_{\beta_{2,j,1}}P(W^B_{2,j}|\bV,\beta)=&P(W_{2,j}^B | \mathbf{V}, \beta) (W_{2,j}^B - \sigma(\beta_{2,j,0} + \beta_{2,j,1}^\top \mathbf{V})) \mathbf{V}, \\
\nabla_{\lambda_2}P(\bW_{2}^R|\bV,\lambda)=& \frac{\left[\sum_{\bv\in\textit{Perm}}\exp(-\lambda_2 T(\bv,\bE^R))(T(\bv,\bE^R)-T(\bW_2^R,\bE^R)\right]}{\left(\sum_{\bv\in\textit{Perm}}\exp(-\lambda_2 T(\bv,\bE^R))\right)^2}\exp(-\lambda_2 T(\bW_2^R,\bE^R)), \\
\nabla_{\alpha_{3,j,0}}P(\bW^{Sat}_3 | \bE^\top\bY, \alpha)=& \left[I(\bW^{Sat}_3 = j)- I(\bW^{Sat}_3 = j-1)\right]\sigma'(\alpha_{3,j,0}-\alpha_{3,\cdot,1}\bE^\top\bY),\\
\nabla_{\alpha_{3,\cdot,1}}P(\bW^{Sat}_3 | \bE^\top\bY, \alpha)=&  - \Big[ I(\bW^{Sat}_3 \neq 7)\sigma'(\alpha_{3,\bW^{Sat}_3,0}-\alpha_{3,\cdot,1}\bE^\top\bY)\\
&+I(\bW^{Sat}_3\neq 1)\sigma'(\alpha_{3,\bW^{Sat}_3-1,0}-\alpha_{3,\cdot,1}\bE^\top\bY)\Big](\bE^\top\bY). 
\end{align*}
\normalsize

\noindent
Denote $\beta_{k,j,\cdot}=(\beta_{k,j,0},\beta_{k,j,1}^\top)^\top$, $\bV^*=(1,\bV^\top)^\top$, and $\bW_k^{B(-j)}=(\bW^B_{k,1},\dots,\bW^B_{k,j-1},\bW^B_{k,j+1},\\\dots,\bW^B_{k,12})^\top$. We have:
\small
\begin{align*}
M_\theta(\bH_{3}, \bV)=& P_\theta(\bW^B_1|\bV)P_\theta(\bW^B_2|\bV)P_\theta(\bW_1^R|\bV)P(\bW_2^R|\bV)P_\theta(\bW^{Sat}_2|\bE^\top\bX_2)P_\theta(\bW^{Sat}_3|\bE^\top\bY), \text{ so}
\end{align*}
{For any $1\leq j \leq 12, \; 1\leq k \leq 2$}, 
\begin{align*}
\nabla_{\beta_{k,j,\cdot}}M_\theta(\bH_{3}, \bV)=& P_\theta(\bW_1^{R}|\bV)P_\theta(\bW^{Sat}_2|\bE^\top\bX_2)P_\theta(\bW_2^{R}|\bV)P_\theta(\bW^{Sat}_3|\bE^\top\bY)\\ &\times P_\theta(\bW^B_{3-k}|\bV)P_\theta(\bW_k^{B,(-j)}|\bV)P_\theta(W_{k,j}^B | \mathbf{V}) (W_{k,j}^B - \sigma(\beta_{1,j,\cdot}^\top \mathbf{V}^*)) \mathbf{V}^*, \\
\nabla_{\lambda_k}M_\theta(\bH_{3}, \bV)=& P_\theta(\bW^B_1|\bV)P_\theta(\bW^{Sat}_2|\bE^\top\bX_2)P_\theta(\bW^B_2|\bV)P_\theta(\bW^{Sat}_3|\bE^\top\bY)P_\theta(\bW_{3-k}^{R}|\bV)\\ &\times\exp(-\lambda_k T(\bW_k^R,\bE^R))\frac{\left[\sum_{\bv\in\textit{Perm}}\exp(-\lambda_k T(\bv,\bE^R))(T(\bv,\bE^R)-T(\bW_k^R,\bE^R)\right]}{\left(\sum_{\bv\in\textit{Perm}}\exp(-\lambda_k T(\bv,\bE^R))\right)^2}.
\end{align*}
{And for any $1 \leq j \leq 6, \;  2 \leq k \leq 3$}, 
\begin{align*}
\nabla_{\alpha_{k,j,0}}M_\theta(\bH_{3}, \bV)=& P_\theta(\bW^B_1|\bV)P_\theta(\bW_1^{R}|\bV)P_\theta(\bW^B_2|\bV)P_\theta(\bW_2^{R}|\bV)\\
&\times P_\theta(\bW^{Sat}_{k}|\bE^\top\bX_2)^{I(k=2)}P_\theta(\bW^{Sat}_{k}|\bE^\top\bY)^{I(k=3)}
\left(I(\bW^{Sat}_{k} = j)- I(\bW^{Sat}_{k} = j-1)\right) \\
&\times \{\sigma'(\alpha_{k,j,0}-\alpha_{k,\cdot,1}\bE^\top\bX_{k})\}^{I(k=2)} \{\sigma'(\alpha_{k,j,0}-\alpha_{k,\cdot,1}\bE^\top\bY)\}^{I(k=3)},  \\
\nabla_{\alpha_{k,\cdot,1}}M_\theta(\bH_{3}, \bV)=
&P_\theta(\bW^B_1|\bV)P_\theta(\bW_1^{R}|\bV)P_\theta(\bW^B_2|\bV)P_\theta(\bW_2^{R}|\bV)\\
&\times P_\theta(\bW^{Sat}_{k}|\bE^\top\bX_2)^{I(k=2)}P_\theta(\bW^{Sat}_{k}|\bE^\top\bY)^{I(k=3)}\\
&\times \Big[- I(\bW^{Sat}_{k} \neq 7)\sigma'(\alpha_{k,\bW^{Sat}_{k},0}-\alpha_{k,\cdot,1}(\bE^\top\bX_2)^{I(k=2)}(\bE^\top\bY)^{I(k=3)})\\
&+ I(\bW^{Sat}_{k}\neq 1)\sigma'(\alpha_{k,\bW^{Sat}_{k}-1,0}-\alpha_{k,\cdot,1}(\bE^\top\bX_2)^{I(k=2)}(\bE^\top\bY)^{I(k=3)})\Big]\\
&\times (\bE^\top\bX_2)^{I(k=2)}(\bE^\top\bY)^{I(k=3)}.
\end{align*}
\normalsize

We can see that $\nabla_\theta M_\theta(\bH_{3}, \bV)$ is continuous in $\theta$ and so is $\nabla_\theta P_\theta(\bH_{3}|\bV)=g(\bH_{3})\nabla_\theta M_\theta(\bH_{3}, \bV)$. We now derive an upper bound for $||\nabla_\theta M_\theta(\bH_{3}, \bV)||_{L^\infty(P_{\theta_0})}$.  As $|P(\bW_k^R | \bV ;\theta)| \leq 1$,
$P(\bW^{Sat}_{2}| \bE^\top\bX_2 ;\theta)| \leq 1$, $P(\bW^{Sat}_{3}| \bE^\top\bY ;\theta)| \leq 1$ and $|P(\bW^{B}_k ; \bV ;\theta)|\leq 1$, $\forall \theta\in\Theta$, $k = 1,2$, and $|\sigma'(x)|\leq 1$, $\forall x\in\mathbb{R}$, we have that $|\nabla_{\beta_{k,j,\cdot}} M_\theta(\bH_{3}, \bV)|\leq \max_j |\bV^*_j|$. Further, $|\exp(-\lambda_k T(\bW_k^R,\bE^R)|<1$ and $|T(\bv,\bE^R)|$, and $|T(\bW_k^R,\bE^R)|\leq 3$ for $\bv\in\textit{Perm}$, $\exp(-\lambda_k T(\bv,\bE^R))=1$ for some $\bv \in \textit{Perm}$ and $|\textit{Perm}|=6$,  we have that $|\nabla_{\lambda_k} M_\theta(\bH_{3}, \bV)|\leq 18$. As $|\bE^\top\bX|\leq10$, $|\nabla_{\alpha_{k,j,0}} M_\theta(\bH_{3}, \bV)|\leq 1$ and $|\nabla_{\alpha_{k,\cdot,1}} M_\theta(\bH_{3}, \bV)|\leq 20$, we 
then have that $||\nabla_\theta M_\theta(\bH_{3}, \bV)||_{L^\infty(P_{\theta_0})} \leq \max\{\max_{j}|\bV^*_j|, 20\} = \max\{Z, 20\}$, with $Z \sim \mathcal{N}(0,1)$, so $\E_{Z}[\max\{Z^2, 400\}] < \infty$ and (N3) is satisfied.

It remains to show (N2). By the mean value theorem, an everywhere-differentiable function $f:\mathcal X\to\mathbb{R}$ with bounded first derivatives will be Lipschitz continuous over $\mathcal X$  with Lipschitz constant $L$ upper-bounded as $\sup_{x\in\mathcal X} |f'(x)|$ \citep{Shwartz2014}. In the following, we use superscript $(1)$ and $(2)$ to denote two parameters in the neighborhood of $\theta_0$. First, as $P(W^B_{k,j}|\bV,\theta)=W^B_{k,j}\sigma(\beta_{k,j,0}+\beta_{k,j,1}^{T}\bV)+(1-W^B_{k,j})(1-\sigma(\beta_{k,j,0}+\beta_{k,j,1}^{T}\bV))$, we have $\forall 1 \leq k \leq 2, \;\; 1 \leq j \leq 12$:
\small
\begin{align*}
&|P(W^B_{k,j}|\bV,\theta^{(1)})-P(W^B_{k,j}|\bV,\theta^{(2)})|\\
\leq& W^B_{k,j}\left|\sigma(\beta_{k,j,\cdot}^{(1)T}\bV^*)-\sigma(\beta_{k,j,\cdot}^{(2)T}\bV^*)\right| +(1-W^B_{k,j})\left|\sigma(\beta_{k,j,\cdot}^{(2)T}\bV^*)-\sigma(\beta_{k,j,\cdot}^{(1)T}\bV^*)\right|\\
\leq& W^B_{k,j}|\beta_{k,j,\cdot}^{(1)T}\bV^*-\beta_{k,j,\cdot}^{(2)T}\bV^*|+(1-W^B_{k,j})|\beta_{k,j,\cdot}^{(2)T}\bV^*-\beta_{k,j,\cdot}^{(1)T}\bV^*| \\
=& |(\beta_{k,j,\cdot}^{(2)}-\beta_{k,j,\cdot}^{(1)})^\top\bV^*|\\
\leq& ||\bV^*||_2||\beta_{k,j,\cdot}^{(2)}-\beta_{k,j,\cdot}^{(1)}||_2 \leq ||\bV^*||_2 ||\theta^{(2)}-\theta^{(1)}||_2.
\end{align*}
\normalsize
The first inequality follows from the triangle inequality; The second follows from the fact that the sigmoid function is everywhere-differentiable and $|\sigma'(x)|\leq 1,\forall x\in\mathbb{R}$, making it Lipschitz with constant $L=1$; And the third inequality follows from the Cauchy-Schwartz inequality. Moreover, $\forall 2 \leq k \leq 3, \;\; 1 \leq j \leq 6$:
\small
\begin{align*}
P(\bW^{Sat}_{k}|\bE^\top\bX_{k},\theta)
=&I(\bW^{Sat}_{k}=7)+I(\bW^{Sat}_{k}\neq 7)\sigma(\alpha_{k,\bW^{Sat}_{k},0}-\alpha_{k,\cdot,1}\bE^\top\bX_{k})\\
&-I(\bW^{Sat}_{k}\neq 1)\sigma(\alpha_{k,\bW^{Sat}_{k}-1,0}-\alpha_{k,\cdot,1}\bE^\top\bX_{k}), \;\text{so} \\
|P(\bW^{Sat}_{k}|\bE^\top\bX_{k},\theta^{(1)})&-P(\bW^{Sat}_{k}|\bE^\top\bX_{k},\theta^{(2)})|\\
\leq& I(\bW^{Sat}_{k}\neq 7)\left|\sigma(\alpha_{k,\bW^{Sat}_{k},0}^{(1)}-\alpha_{k,\cdot,1}^{(1)}\bE^\top\bX_{k})-\sigma(\alpha_{k,\bW^{Sat}_{k},0}^{(2)}-\alpha_{k,\cdot,1}^{(2)}\bE^\top\bX_{k})\right|\\&+I(\bW^{Sat}_{k}\neq 1)\left|\sigma(\alpha_{k,\bW^{Sat}_{k}-1,0}^{(1)}-\alpha_{k,\cdot,1}^{(1)}\bE^\top\bX_{k})-\sigma(\alpha_{k,\bW^{Sat}_{k}-1,0}^{(2)}-\alpha_{k,\cdot,1}^{(2)}\bE^\top\bX_{k})\right| \\
\leq&|(\alpha_{k,\bW^{Sat}_{k},0}^{(1)}-\alpha_{k,\bW^{Sat}_{k},0}^{(2)})+(\alpha_{k,\cdot,1}^{(2)}-\alpha_{k,\cdot,1}^{(1)})\bE^\top\bX_{k}|\\&+|(\alpha_{k,\bW^{Sat}_{k}-1,0}^{(1)}-\alpha_{k,\bW^{Sat}_{k}-1,0}^{(2)})+(\alpha_{k,\cdot,1}^{(2)}-\alpha_{k,\cdot,1}^{(1)})\bE^\top\bX_{k}| \\
\leq& |\alpha_{k,\bW^{Sat}_{k},0}^{(1)}-\alpha_{k,\bW^{Sat}_{k},0}^{(2)}|+|\alpha_{k,\cdot,1}^{(1)}-\alpha_{k,\cdot,1}^{(2)}|\bE^\top\bX_{k}\\
&+|\alpha_{k,\bW^{Sat}_{k}-1,0}^{(1)}-\alpha_{k,\bW^{Sat}_{k}-1,0}^{(2)}|+|\alpha_{k,\cdot,1}^{(1)}-\alpha_{k,\cdot,1}^{(2)}|\bE^\top\bX_{k} \\
\leq& 2||\alpha_{k,\cdot,0}^{(1)}-\alpha_{k,\cdot,0}^{(2)}||_2+20|\alpha_{k,\cdot,1}^{(1)}-\alpha_{k,\cdot,1}^{(2)}|\\
\leq& 22\left\|\theta^{(2)}-\theta^{(1)}\right\|_2.
\end{align*}
\normalsize
The first inequality uses the triangle inequality for absolute values. The second inequality uses the fact that the sigmoid function has Lipschitz constant $L=1$. The third inequality uses the triangle inequality again. The fourth inequality uses $|\bE^\top\bX_{k}|\leq 10, |\bE^\top\bY|\leq 10$. 

Note that $f:[0,\infty)\to[0,1]$ defined by $f(x)=\exp(-x)$ is Lipschitz with constant $L=\sup_{x\in[0,\infty)}\{f'(x)\}\leq 1$. Moreover, note that the set $\mathcal T= \{T(\bv,\bE^R): \bv\in\textit{Perm}\}$ is equivalent for all $\bE^R\in\textit{Perm}$. Finally, observe that $|\exp(-\lambda_k T(\bW_1^R,\bE^R))| \leq 1, | \{\sum_{\bv\in \textit{Perm}}\exp(-\lambda_k T(\bv,\bE^R))\}^{-1} |\leq 1$,  $T(x,y)\in\{0,1,2,3\}$, and $|\mathcal T|=6$. Putting these all together, we have that $\forall 1 \leq k \leq 2$,
\small
\begin{align*}
\nabla_{\lambda_k}\left({\sum_{\bv\in\textit{Perm}}\exp(-\lambda_kT(\bv,\bE^R))}\right)^{-1}= \frac{{\sum_{T\in \mathcal T}\exp(-\lambda_kT)T}}{\left(\sum_{T\in\mathcal T}\exp(-\lambda_k T)\right)^2}\leq 18
\end{align*}
\normalsize

Thus $f:[0,\infty)\to\mathbb{R}$ defined as $f(x)=1/\sum_{T\in\mathcal T}\exp(-xT)$ is Lipschitz with constant $L\leq 18$. Then:
\small
\begin{align*}
P(\bW_k^R|\bV,\lambda_k)=&\frac{\exp(-\lambda_k T(\bW_k^R,\bE^R))}{\sum_{\bv\in\textit{Perm}}\exp(-\lambda_k T(\bv,\bE^R))}\;\;\mbox{and}
\end{align*}
\begin{align*}
&|P(\bW_k^R|\bV,\lambda_k^{(2)})-P(\bW_k^R|\bV,\lambda_k^{(1)})|\\
=& \left|\frac{\exp(-\lambda_k^{(2)} T(\bW_k^R,\bE^R))}{\sum_{\bv\in\textit{Perm}}\exp(-\lambda_k^{(2)} T(\bv,\bE^R))}-\frac{\exp(-\lambda_k^{(1)} T(\bW_k^R,\bE^R))}{\sum_{\bv\in\textit{Perm}}\exp(-\lambda_k^{(1)} T(\bv,\bE^R))}\right|\\
\leq& \left|\frac{\exp(-\lambda_k^{(2)} T(\bW_k^R,\bE^R))}{\sum_{\bv\in\textit{Perm}}\exp(-\lambda_k^{(2)} T(\bv,\bE^R))}-\frac{\exp(-\lambda_k^{(1)} T(\bW_k^R,\bE^R))}{\sum_{\bv\in\textit{Perm}}\exp(-\lambda_k^{(2)} T(\bv,\bE^R))}\right|\\&+
\left|\frac{\exp(-\lambda_k^{(2)} T(\bW_k^R,\bE^R))}{\sum_{\bv\in\textit{Perm}}\exp(-\lambda_k^{(2)} T(\bv,\bE^R))}-\frac{\exp(-\lambda_k^{(1)} T(\bW_k^R,\bE^R))}{\sum_{\bv\in\textit{Perm}}\exp(-\lambda_k^{(1)} T(\bv,\bE^R))}\right| \\
\leq& \frac{}{}\left|\exp(-\lambda_k^{(2)} T(\bW_k^R,\bE^R))-\exp(-\lambda_k^{(1)} T(\bW_k^R,\bE^R))\right|\\&+
\left|\frac{1}{\sum_{\bv\in\textit{Perm}}\exp(-\lambda_k^{(2)} T(\bv,\bE^R))}-\frac{1}{\sum_{\bv\in\textit{Perm}}\exp(-\lambda_k^{(1)} T(\bv,\bE^R))}\right| \\
\leq& |\lambda_k^{(2)}T(\bW_k^R,\bE^R)-\lambda_k^{(1)}T(\bW_k^R,\bE^R)|+18|\lambda_k^{(2)}-\lambda_k^{(1)}|\\
\leq& 21 |\lambda_k^{(2)}-\lambda_k^{(1)}| \\
\leq& 21 \|\theta^{(2)}-\theta^{(1)}\|_2.
\end{align*}
\normalsize
As the product of Lipschitz continuous functions is also Lipschitz continuous with the Lipschitz constant being the sum of those of the functions being multiplied \citep{Shwartz2014}, $M_\theta(\bH_{3}, \bV)$ is Lipschitz continuous in $\theta\in\Theta$ with Lipschitz constant $L(\bV^*) \leq 24||\bV^*||_2+43$, satisfying $\E_{\theta_0, \bV}[ (24||\bV^*||_2+43)^2] < \infty$, so condition (N2) is satisfied, concluding the proof. 
\end{proof}

\vspace*{1cm}

\subsection{Uncertainty-aware Variants}

\noindent
\textbf{Proof of Proposition 4.3:}

\begin{proof}
Denote $\tilde{\pi}$ an arbitrary deterministic policy. 
Consider $\bar{\pi}^{(j)}$ defined as 
\[
\bar{\pi}^{(j)}_k =
\begin{cases}
\pi_k^{\mathrm{opt}}, & \text{if } k < j, \\
\tilde{\pi}_k,        & \text{if } k \ge j .
\end{cases}
\] 
for $j = 1, \dots, K + 1$. So $\bar{\pi}^{(1)} = \tilde{\pi}$ follows $\tilde{\pi}$ at all stages and $\bar{\pi}^{(K+1)} = \pi^{\mathrm{opt}}$ follows $\pi^{\mathrm{opt}}$ at all stages.
\begin{align*}
V_1\!\left(\bar{\pi}^{(k+1)}\right) - V_1\!\left(\bar{\pi}^{(k)}\right)
&= \mathbb{E}_{\mathbf{H}_k \sim \pi^{\mathrm{opt}}_{1:k-1}}
\Big[
Q_k^{\pi^{\mathrm{opt}}}\!\big(\mathbf{H}_k,\pi_k^{\mathrm{opt}}(\mathbf{H}_k)\big)
-
Q_k^{\pi^{\mathrm{opt}}}\!\big(\mathbf{H}_k,\tilde{\pi}_k(\mathbf{H}_k)\big)
\Big] \nonumber \\
&= \mathbb{E}
\Big[
\bar{\mathbf{E}}_k^{\top}\, \mathbf{y}_k(\pi_k^{\mathrm{opt}})
-
\bar{\mathbf{E}}_k^{\top}\, \mathbf{y}_k(\tilde{\pi}_k)
\Big]
\qquad \text{(by A5)} \nonumber \\
&= \mathbb{E}\!\left[ \Delta_k(\tilde{\pi}_k) \right],
\label{eq:telescoping}
\end{align*}
where
$
\Delta_k(\tilde{\pi}_k)
:=
\bar{\mathbf{E}}_k^{\top}\, \mathbf{y}_k(\pi_k^{\mathrm{opt}})
-
\bar{\mathbf{E}}_k^{\top}\, \mathbf{y}_k(\tilde{\pi}_k)
\geq 
0
$
by the definition of $\pi^{opt}_k$.
Thus 
\begin{align*}
V_1(\pi^{\mathrm{opt}}) - V_1(\tilde{\pi})
=
V_1\!\left(\bar{\pi}^{(K+1)}\right)
-
V_1\!\left(\bar{\pi}^{(1)}\right)
=
\sum_{k=1}^{K}
\Big[
V_1\!\left(\bar{\pi}^{(k+1)}\right)
-
V_1\!\left(\bar{\pi}^{(k)}\right)
\Big]
=
\sum_{k=1}^{K} \mathbb{E}\!\left[ \Delta_k(\tilde{\pi}_k) \right],
\end{align*}
and the result directly follows.
\end{proof}

\vspace*{1cm}
\noindent
\textbf{Proof of Corollary 4.4:}
\begin{proof}
Let $\pi_k^{\mathrm{ind}}(a_k \mid \mathbf{H}_k)$ denote the action distribution
induced by the estimated posterior:
\[
\pi_k^{\mathrm{ind}}(a_k \mid \mathbf{H}_k)
=
\mathbb{P}\!\left(
\arg\max_{a_k'}
\mathbf{E}^{\top}\mathbf{y}_k(a_k') = a_k
\;\middle|\;
\mathbf{E} \sim \hat{P}(\mathbf{E}\mid \mathbf{H}_k)
\right).
\]
Define random variable $Z := \mathbf{E}^{\top}\mathbf{d}_k
= \mathbf{E}^{\top}\mathbf{y}_k(\tilde{\pi}_k)
- \mathbf{E}^{\top}\mathbf{y}_k(\pi_k^{\mathrm{opt}})$,
where $\mathbf{E} \sim P(\mathbf{E}\mid \mathbf{H}_k)$.
Then 
\begin{align}
\mathbb{E}[Z \mid \mathbf{H}_k]
&=
\bar{\mathbf{E}}_k^{\top}\mathbf{y}_k(\tilde{\pi}_k)
-
\bar{\mathbf{E}}_k^{\top}\mathbf{y}_k(\pi_k^{\mathrm{opt}})
= -\Delta_k(\tilde{\pi}_k)
< 0,
\label{eq:Zmean}
\\
\mathrm{Var}(Z \mid \mathbf{H}_k)
&=
\mathbf{d}_k^{\top}
\mathrm{Cov}(\mathbf{E}\mid \mathbf{H}_k)
\mathbf{d}_k
=
\mathbf{d}_k^{\top}\Sigma_k\mathbf{d}_k .
\label{eq:Zvar}
\end{align}
Denote $\tilde{\pi}_k$ any deterministic policy different from $\pi^{opt}$. Then 
\begin{align}
\pi_k^{\mathrm{ind}}\!\big(\tilde{\pi}_k \mid \mathbf{H}_k\big)
=
\mathbb{P}(Z>0 \mid \mathbf{H}_k)
=
\mathbb{P}\!\left(
Z-\mathbb{E}[Z\mid \mathbf{H}_k] > \Delta_k
\;\middle|\; \mathbf{H}_k
\right).
\label{eq:Zprob}
\end{align}
By the boundedness assumption for $\mathbf{Y}$, $\|\mathbf{d}_k\| \leq 2 \sup_{a_k}\|\mathbf{y}(a_k)\| < \infty$. Also, $\mathbf{E} \in \Delta^{d-1}$ implies $\mathbf{E}$ bounded almost surely thus all entries of $\Sigma_k$ are bounded. These two together implies $Var(Z\mid \mathbf{H}_k) < \infty$.
Note that if $\Sigma_k$ is singular with $\mathbf{d}_k$ in the null space, in other words $\Sigma_k \mathbf{d}_k = 0$, then $Var(Z\mid \mathbf{H}_k) = 0$ and the condition that $\pi_k^{\mathrm{ind}}(\tilde{\pi}_k \mid \mathbf{H}_k) \ge \alpha_k > 0$ for some $\tilde{\pi}_k \neq \pi^{opt}$ is violated.
Apply then Cantelli's inequality to $\ref{eq:Zprob}$ gives 
\begin{align*}
\pi_k^{\mathrm{ind}}\!\big(\tilde{\pi}_k \mid \mathbf{H}_k\big)
\le
\alpha_k
\le 
\frac{\mathbf{d}_k^{\top}\Sigma_k \mathbf{d}_k}
{\mathbf{d}_k^{\top}\Sigma_k \mathbf{d}_k + \Delta_k^2}.
\end{align*}
Rearranging the inequality, we have that $\alpha_k \Delta_k^2
\le
(1-\alpha_k)\mathbf{d}_k^{\top}\Sigma_k \mathbf{d}_k$ and 
\[
\Delta_k
\le
\sqrt{\frac{1-\alpha_k}{\alpha_k}}
\sqrt{\mathbf{d}_k^{\top}\Sigma_k \mathbf{d}_k},
\]
which is the condition in Proposition 4.3 with $\phi_k=\sqrt{(1-\alpha_k)/\alpha_k}$ and $\mathbf{c}_k=\mathbf{d}_k$. 
Thus Proposition 4.3 gives that 
$
V_1(\pi^{\mathrm{opt}}) - V_1(\tilde{\pi})
\le
\sum_{k=1}^{K}
\mathbb{E}\!\left[
\sqrt{\frac{1-\alpha_k}{\alpha_k}}
\sqrt{\mathbf{d}_k^{\top}\Sigma_k \mathbf{d}_k}
\right].
$
Suppose we distribute $\epsilon$ uniformly over the stages so that each stage contribute at most $\epsilon / K$, then for 
$
\sigma_k := \E[\sqrt{\mathbf{d}_k^{\top} \Sigma_k \mathbf{d}_k}],
$
$
\sqrt{\frac{1-\alpha_k}{\alpha_k}} \sigma_k \leq \epsilon / K 
$
implies that the selected $\tilde{\pi}_k$ must satisfy 
$\alpha_k \ge \alpha_k^{*}
:= \frac{K^2 \sigma_k^2}{K^2 \sigma_k^2 + \epsilon^2}.$
\end{proof}

\vspace*{1cm}
\noindent
\textbf{Proof of Corollary 4.5:}
\begin{proof}
Let $\hat{\pi}^{\mathrm{rob}}$ be the robust policy with uncertainty radius
$\delta>0$, defined by
\begin{align*}
\hat{\pi}_k^{\mathrm{rob}}(\mathbf{H}_k)
&=
\arg\max_{a_k \in \mathcal{A}_{k}}
\;
\min_{\mathbf{E}\in \mathcal{U}_k(\mathbf{H}_k)}
\mathbf{E}^{\top}\mathbf{y}_k(a_k), \\
\mathcal{U}_k(\mathbf{H}_k)
&=
\Big\{
\mathbf{E}\in \Delta^{d-1} :
(\mathbf{E}-\bar{\mathbf{E}}_k)^{\top}
\Sigma_k^{-1}
(\mathbf{E}-\bar{\mathbf{E}}_k)
\le \delta^2
\Big\}.
\end{align*}
The Lagrangian for the inner problem can be expressed as 
\[
\mathcal{L}(x,\lambda)
= x^\top v
+ \lambda \bigl(x^\top \Sigma_k^{-1} x - \delta^2\bigr), \quad \lambda \ge 0,
\]
where $x := \mathbf{E} - \bar{\mathbf{E}}_k$, $v := \mathbf{y}_k(a_k)$. Solving from the KKT conditions 
$\nabla_x \mathcal{L}(x,\lambda) = v + 2\lambda \Sigma_k^{-1} x = 0$
and
$x^\top \Sigma_k^{-1} x = \delta^2, \lambda > 0$
gives 
$\mathbf{E}^* = \bar{\mathbf{E}}_k + x^* = \bar{\mathbf{E}}_k - \delta \frac{\Sigma_k v}{\sqrt{v^\top \Sigma_k v}}.$
$
\mathbf{E}^* \in \Delta^{d-1}
$
as 
$(\mathbf{E}_k - \bar{\mathbf{E}}_k)^\top \mathbf{1} = 0$ implies $\Sigma_k \mathbf{1}
=\mathbb{E}\!\left[(\mathbf{E}_k - \bar{\mathbf{E}}_k)(\mathbf{E}_k - \bar{\mathbf{E}}_k)^\top \mathbf{1}\right] = 0$ and
$\sum_j \mathbf{E}^*_j = \sum_j \bar{\mathbf{E}}_j - \delta \frac{\Sigma_k \mathbf{1} v}{\sqrt{v^\top \Sigma_k v}} = 1 - 0 = 1$. Thus the robust policy solves 
\[
\hat{\pi}_k^{\mathrm{rob}}
=
\arg\max_{a_k}
\left[
\bar{\mathbf{E}}_k^\top \mathbf{y}_k(a_k)
-
\delta
\sqrt{
\mathbf{y}_k(a_k)^\top
\Sigma_k
\mathbf{y}_k(a_k)
}
\right].
\]
Since $\hat{\pi}_k^{\mathrm{rob}}$ is the maximizer, its penalized objective dominates that of $\pi_k^{\mathrm{opt}}$:
\begin{align*}
\bar{\mathbf{E}}_k^\top \mathbf{y}_k(\hat{\pi}_k^{\mathrm{rob}})
- \delta
\sqrt{
\mathbf{y}_k(\hat{\pi}_k^{\mathrm{rob}})^\top
\Sigma_k
\mathbf{y}_k(\hat{\pi}_k^{\mathrm{rob}})
}
&\ge
\bar{\mathbf{E}}_k^\top \mathbf{y}_k(\pi_k^{\mathrm{opt}})
- \delta
\sqrt{
\mathbf{y}_k(\pi_k^{\mathrm{opt}})^\top
\Sigma_k
\mathbf{y}_k(\pi_k^{\mathrm{opt}})
}.
\end{align*}
Rearranging and define
$
\Delta_k^{\mathrm{rob}}:=\bar{\mathbf{E}}_k^\top \mathbf{y}_k(\pi_k^{\mathrm{opt}})-\bar{\mathbf{E}}_k^\top \mathbf{y}_k(\hat{\pi}_k^{\mathrm{rob}}),
$
so that
\begin{align}
\Delta_k^{\mathrm{rob}}
&\le
\delta
\left[
\sqrt{\mathbf{y}_k(\hat{\pi}_k^{\mathrm{rob}})^\top \Sigma_k \mathbf{y}_k(\hat{\pi}_k^{\mathrm{rob}})}
-
\sqrt{\mathbf{y}_k(\pi_k^{\mathrm{opt}})^\top \Sigma_k \mathbf{y}_k(\pi_k^{\mathrm{opt}})}
\right] \notag \\
&\le
\delta
\sqrt{\mathbf{y}_k(\hat{\pi}_k^{\mathrm{rob}})^\top \Sigma_k \mathbf{y}_k(\hat{\pi}_k^{\mathrm{rob}})
},
\label{eq:12}
\end{align}
where the last inequality drops the non-positive term
$
-\delta \sqrt{\mathbf{y}_k(\pi_k^{\mathrm{opt}})^\top \Sigma_k \mathbf{y}_k(\pi_k^{\mathrm{opt}})} \le 0.
$
This is the condition in Proposition 4.3 with $\phi_k = \delta$ and
$c_k = \mathbf{y}_k(\hat{\pi}_k^{\mathrm{rob}})$.
Suppose (V2) in Theorem 4.2 satisfies, then $\exists M < \infty$ such that $\hat{\mathbf{y}}_k(a_k)^\top \Sigma_k \hat{\mathbf{y}}_k(a_k)
\;\le\;
\lambda_{\max}(\Sigma_k)\,\|\hat{\mathbf{y}}_k(a_k)\|^2
\;\le\;
\lambda_{\max}(\Sigma_k) M^2.$ 
Thus Proposition 4.3 gives that 
$
V_1(\pi^{\mathrm{opt}})
-
V_1(\hat{\pi}^{\mathrm{rob}})
\;\le\;
\delta \, M \, \sum_{j=1}^K \sqrt{\lambda_{\max}(\Sigma_j)},
$
implying that for $\hat{\pi}^{rob}$ to be $\epsilon$-optimal, one need
$\delta \leq \delta^* = \epsilon / (M \sum_{j=1}^K \sqrt{\lambda_{max}(\Sigma_j)})$. 

\end{proof}

\section{Latent Variable Modeling}
\label{append:latentmodeling}

\subsection{Model Selection}
\label{append:modelselection}

When estimating $\mathbb{E}[\mathbf{E} \mid \mathbf{H}_k]$ via the parametric route, one must specify a model for $P(\mathbf{H}_{K+1} \mid \mathbf{E})$ with parameter vector $\theta$ and a distribution $P(\mathbf{E})$. Since neither is known in advance, we recommend proposing a finite set of diverse candidate models $M_1(\mathbf{H}_{K+1} \mid \mathbf{E}, \theta_1), \dots, M_P(\mathbf{H}_{K+1} \mid \mathbf{E}, \theta_P)$ and a finite collection of candidate distributions for $P(\mathbf{E})$, making only the weaker assumption that at least one pair is correctly specified.

\textbf{Selection criterion.} For any fixed $P(\mathbf{E})$, partition the data as $\mathcal{D} = \mathcal{D}_T \cup \mathcal{D}_V$, fit each candidate on $\mathcal{D}_T$, and evaluate on $\mathcal{D}_V$ via the observed log-likelihood
\begin{equation}
  \label{eq:obsll}
  \ell_p = \sum_{\mathbf{H}_{K+1} \in \mathcal{D}_V}
    \log \int_{\mathcal{E}} M_p\!\left(\mathbf{H}_{K+1} \mid \mathbf{E},
    \hat{\theta}_p\right) dP(\mathbf{E}),
\end{equation}
or its BIC analogue $\ell_p - \tfrac{1}{2}\,\mathrm{Card}(\theta_p)\log n$, which penalizes model complexity. Select $\hat{p} = \arg\max_{1 \le p \le P} \ell_p$. In practice, when the candidate class for $P(\mathbf{H}_{K+1} \mid \mathbf{E})$ is sufficiently rich, a reliable estimate of $\theta$ can typically be obtained without exhaustively enumerating candidate distributions for $P(\mathbf{E})$. Furthermore, since $P(\mathbf{H}_{K+1} \mid \mathbf{E}) = f(\mathbf{H}_{K+1} \mid \mathbf{E})\,g(\mathbf{H}_{K+1})$ where $g(\mathbf{H}_{K+1})$ is unknown and does not depend on $\theta$, it is sufficient to propose and select among models for $f(\mathbf{H}_{K+1} \mid \mathbf{E})$ alone. Lemma~\ref{lemma:modelselect} provides theoretical justification; the proof is a direct application of Lemma~5.35 of 
\citet{Vaart1998}.

\textbf{Conjugate pairs.} When the selected models for $P(\mathbf{H}_{K+1} \mid \mathbf{E})$ and $P(\mathbf{E})$ form a conjugate pair, the posterior $P(\mathbf{E} \mid \mathbf{H}_k)$ and the integral in \eqref{eq:obsll} admit closed-form expressions, bypassing Monte Carlo integration entirely and substantially reducing computational cost. Practitioners should prefer conjugate specifications when available; for example, a Dirichlet prior on $\mathbf{E}$ paired with a Categorical model for discrete preference responses, or a Gaussian prior paired with a linear-Gaussian preference model.

\textbf{Diagnostics.} Model selection via \eqref{eq:obsll} is a relative comparison — it identifies the best-fitting model among candidates but does not assess absolute adequacy. We recommend complementing selection with posterior predictive checks: simulate replicated preference responses 
$\mathbf{W}_k^{rep}$ by drawing $\mathbf{E}^{(b)} \sim \Lambda_E$ and 
$\mathbf{W}_k^{rep} \sim M_{\hat{p}}(\cdot \mid \mathbf{E}^{(b)}, 
\hat{\theta}_{\hat{p}})$, then compare distributional summaries of 
$\mathbf{W}_k^{rep}$ to the observed $\mathbf{W}_k$. Natural test statistics include item-level response frequencies for binary questionnaires, empirical rank concordance rates, and marginal distributions of satisfaction scores. Systematic discrepancies indicate model misfit that may not be detected by likelihood comparison alone and should prompt consideration of richer model classes.

\begin{lemma}
\label{lemma:modelselect}
Suppose $\mathcal{M} = \{f_\theta : \theta \in \Theta = \{\theta_1, \dots, 
\theta_P\}\}$ is a finite class of models for $f(\mathbf{H}_{K+1} \mid 
\mathbf{E})$, where $F_{\theta_p}(\mathbf{H}_{K+1}) = f_p(\mathbf{H}_{K+1} 
\mid \mathbf{E}, \theta_p)\,g(\mathbf{H}_{K+1})$, $1 \le p \le P$, define 
valid probability measures. Suppose $\exists\,\theta_0 \in \Theta$ such that 
$F_{\theta_0}(\mathbf{H}_{K+1} \mid \mathbf{E}) = f(\mathbf{H}_{K+1} \mid 
\mathbf{E})$ and $F_{\theta_p}(\mathbf{H}_{K+1}) \neq F_{\theta_0}
(\mathbf{H}_{K+1})$ for all $\theta_p \neq \theta_0$. Then 
$\mathbb{E}_{\theta_0}[\log(dF_{\theta_0}/dF_\theta)]$ attains its unique 
maximum at $\theta = \theta_0$.
\end{lemma}

\subsection{Specification of Priors}
\label{append:priors}

Maximum A Posteriori (MAP) estimation requires selecting priors for $\theta$. Two regularity conditions from Lemma~\ref{lemma_theta} govern valid prior choices: (C6) requires 
$\Lambda_\theta(\theta_0) > 0$ and $\sup_{\theta \in \Theta} \log 
\Lambda_\theta(\theta) < \infty$, ensuring the prior does not dominate the likelihood asymptotically; (N4) requires $\Lambda_\theta$ to be continuously differentiable near $\theta_0$, ensuring the prior contributes only $o(1)$ to the normalized score equation and does not affect consistency or asymptotic normality. Both conditions are satisfied by any proper, continuously differentiable prior supported on a compact $\Theta$.

\textbf{Prior shape and regularization.} Prior choice determines the 
regularization structure on $\theta$. A normal prior $\Lambda_\theta(\theta) \propto \exp(-\|\theta\|_2^2/2\sigma^2)$ corresponds to $L_2$ regularization and shrinks all components toward zero uniformly, which is appropriate when parameters are on comparable scales. A Laplace prior corresponds to $L_1$ regularization and promotes sparsity, which may be preferable when many binary questionnaire loadings $\beta_{k,j,1}$ are expected to be near zero. For sign-constrained parameters such as $\alpha_{k,\cdot,1} > 0$ and $\lambda_k > 0$, priors supported on the positive half-line — such as log-normal or half-normal — directly encode the constraint without requiring separate barrier penalties, and are preferable to truncated normal priors when the true parameter may be close to zero. In our BEST simulation, we use normal priors combined with a smooth barrier approximation $\frac{1}{10}\sum_m \exp(-20\,c_m(\theta))$ to enforce boundary constraints, which implicitly places a strongly informative prior near the constraint boundary.

\textbf{Prior sensitivity.} We recommend assessing robustness by refitting under two to three prior scales and comparing the resulting posterior mean preferences $\widehat{\mathbb{E}}[\mathbf{E} \mid \mathbf{H}_k; \hat\theta_n]$ and estimated policy values $V(\hat\pi^{opt}_n)$. Substantial variation in policy value across priors suggests the likelihood is weakly informative relative to the prior, which may indicate model overparameterization or insufficient preference data relative to $|\theta|$. Remedies include reducing the number of binary questionnaire items entering the likelihood, pooling stage-specific parameters $\beta_{k,j,1}$ across stages with a hierarchical prior, or collecting additional preference data. We refer readers to \citet{gelman_bayesian_2014} for a comprehensive treatment of prior 
specification and sensitivity analysis.

\section{Application to the BEST Trial, further results}
\subsection{Computation details}
\label{sec:computation_details}
In this subsection we provide additional computation details for the simulation study described in Section 5.1 and 5.2 of the manuscript and is connected to the result shown in Table 1 of the manuscript:

Monte Carlo (MC) integration is used for computing the posterior mean preference with $N_{sim} = 2000$. We considered sample sizes ranging from 300 to 4800, including the target sample size of 600. Given that $|\theta| = 88$, a sample size of 300 represents a stringent small-sample setting. For each scenario, we ran 400 replicates using different random seeds. In each replicate, parameters were sampled and used to generate training data. Testing data were generated independently using a different seed and matched in size to the training data. In all reported tables regarding $V(\widehat{\pi}_n^{opt})$, $\widehat{\pi}_n^{opt}$ is estimated from the training data while its value $V(\widehat{\pi}_n^{opt})$ is computed on the testing data of the same size.

At each $N$, all algorithms share the same training and testing data. Random Forest (RF) with 500 trees is used to fit $\mathbb{E}[\mathbf{Y}|\mathbf{H}_K, A_K]$ and $\mathbb{E}[\widehat{V}^{\widehat{\pi}^{opt}_n}_{k+1}|\mathbf{H}_k, A_k]$ for all algorithms. RF is chosen to remain flexible while handling small training sample sizes relative to the state and action space. Training was done using R packages \textit{caret} and \textit{ranger} with hyperparameters \textit{mtry} and \textit{minimum node size} selected using a grid search via 5-fold cross-validation. The candidate values for \textit{mtry} are set to center around $\left\lfloor \sqrt{\text{number of predictors}} \right\rfloor$ and that for minimum node size are set to be 5, 10, and 25, following common practice \citep{breiman2001, Probst2018HyperparametersAT}.

We run preference model estimation on a GPU using TensorFlow \citep{tf2015}. Reverse-mode automatic differentiation \citep{Geron2019} is used to compute $\nabla_\theta\log P(\theta|\mathcal D)$, and L-BFGS algorithm \citep{Liu1989} is used for optimization with 5 random starting points, and the estimate that yields the largest observed log-likelihood is selected. We also performed 200 simple gradient descent steps with a learning rate of $10^{-4}$ prior to applying L-BFGS to improve stability. To constrain $\hat\theta_n$ to be within $\Theta = \{\alpha, \beta, \gamma, \lambda \in \mathbb{R}: \alpha_{k, \cdot, 1} > 0, \alpha_{k,1,0} < \alpha_{k,2,0} < \cdots < \alpha_{k,6,0}, 1 \leq k \leq 2, \lambda > 0\}$, a penalty of $\frac{1}{10}
\sum_{m}\exp\!\big(-20\, c_m(\theta)\big),$ is added to the minimization objective where $c = (c_1, \dots, c_m)$ a vector of linear combinations of coordinates of $\theta$ that are assumed to be positive. This penalty provides a smooth barrier approximation to the constrained problem, in which the objective is set to $+\infty$ whenever $\theta \notin \Theta$.

\subsection{Data Generating Process for the by K Case}

This subsection provides the detailed data generating process that investigates the effect of trajectory length on algorithm performance, whose result is presented in Table 2 and Figure 2 of the manuscript.

We consider the following data generating process, which is more general than the previous setup that is tailored towards the BEST trial. We generate treatment assignment uniformly over $\mathcal{A} = \{a_1, \dots, a_4\}$ at all stages, considered only the binary questionnaires and reported satisfaction after treatment, and remove the opposing effect of $A_K$ on $Y_2$ and $Y_3$. This would reduce the difference between LUQ-Learning and the naive approach. The data trajectory is now $(\mathbf{X}_1, \mathbf{W}_1, A_1, \dots, \mathbf{Y}, \mathbf{W}_{K+1})$, where $\mathbf{W}_1 = (\mathbf{W}^B_1), \mathbf{W}_k = (\mathbf{W}^B_k, \mathbf{W}^{Sat}_k), k \in \{2, \dots, K\},$ and $\mathbf{W}_{K+1} = (\mathbf{W}^{Sat}_{K+1})$. Training and testing sample size are fixed to be both 600 or both 2400, with optimization approaches the same as before.

{\footnotesize
\begin{align*}
    & \mathbf{V} \sim \mathcal N_{2}(0,\mathbf I), \quad 
    \mathbf E=\text{SoftMax}(( \mathbf{V}, 1)) = \frac{( \exp(\mathbf{V}), 1)}{\sum_{j=1}^2 \exp(V_j)+1}, \\
&\text{At k = 1:}\\
    &\quad {X}_{1j} \sim \text{Bin}\left(n = 10, p = 0.5\right), \quad (1 \leq j \leq 3), \\
    &\quad W^{B}_{1j} \sim \text{Bern}\left( p = \sigma(\beta_{1,j,0} + \beta_{1,j,1}^\top \mathbf V) \right), \quad (1 \leq j \leq 2), \\
    &\quad A_1 \sim \text{Unif}(\mathcal{A}_{1}), \quad \mathcal{A}_{1} = \{a_1, \dots, a_4\}. \\
& \text{At $k = 2, \dots, K$:}\\
    &\quad X_{kj} \sim \text{Bin}\left(n = 10, p = g(A_{k-1}, X_{k-1,j}) \right), \quad (1 \leq j \leq 3), \\
    &\quad g(A_{k-1}, X_{k-1,j}) = \sigma\left[\sum_{a_l \in \mathcal{A}_{k-1}} \gamma_{k,j,l,0} I(A_{k-1} = a_l) + \frac{X_{k-1,j} - \widehat{\mathbb{E}}[X_{k-1,j}]}{\widehat{\mathrm{SD}}(X_{k-1,j})} \sum_{a_l \in \mathcal{A}_{k-1}} \gamma_{k,j,l,1} I(A_{k-1} = a_l) \right],\\
    &\quad W^{B}_{kj} \sim \text{Bern}\left( p = \sigma(\beta_{k,j,0} + \beta_{k,j,1}^\top \mathbf V) \right), \quad (1 \leq j \leq 2), \\
    &\quad P(\mathbf{W}^{Sat}_k \leq j | \mathbf{Y}, \mathbf{E}) = \sigma\left(\alpha_{k,j,0} - \alpha_{k,\cdot,1} \mathbf{E}^\top \mathbf{X}_k\right), \quad (1 \leq j \leq 2),\\
    &\quad A_k \sim \text{Unif}(\mathcal{A}_{k}), \quad \mathcal{A}_{k} = \{a_1, \dots, a_4\}. \\
&\text{At k = K+1:}\\
    &\quad Y_{j} \sim \text{Bin}\left(n = 10, p = g(A_{K}, X_{K,j}) \right), \quad (1 \leq j \leq 3), \\
    &\quad g(A_{K}, X_{K,j}) = \sigma\left[\sum_{a_l \in \mathcal{A}_{K}} \gamma_{k,j,l,0} I(A_{K} = a_l) + \frac{X_{K,j} - \widehat{\mathbb{E}}[X_{K,j}]}{\widehat{\mathrm{SD}}(X_{K,j})} \sum_{a_l \in \mathcal{A}_{K}} \gamma_{k,j,l,1} I(A_{K} = a_l) \right],\\
    &\quad \mathbf{W}^{Sat}_{K+1} \text{ generated by the same means as when k = 2, ..., K.}
\end{align*} 
}
\normalsize
Parameters are generated as follows. We let parameters at the next stage be positively correlated to the previous stage with some additive random noise. In this setting, for $K \geq 2$, $\theta(K) = (\alpha(K), \beta(K))$, where $\alpha(K) = (\alpha_{k,j,0}, \alpha_{k,\cdot,1})_{k=2,j=1}^{k=K, j = 2}$, $\beta(K) = (\beta_{k,j,0}, \beta_{k,j,1})_{k=1,j=1}^{k=K, j=2}$, so $|\theta(K)| = 7K-3$.
\footnotesize
\begin{align*}
&\text{At k = 1:}\\
    &\quad \beta_{1,j,0} = 0, \quad \beta_{1,j,1} \sim \mathcal{N}_2(0, \mathbf{I}_2), \quad (1 \leq j \leq 2).\\
&\text{At k = 2}\\
    &\quad \alpha_{2,j,0} = 0.75 j, \quad \alpha_{2,\cdot,1} = 0.25 + 0.10 /K, \quad (1 \leq j \leq 2),\\
    &\quad \beta_{2,j,0} = 0, \quad \beta_{2,j,1} \sim \mathcal{N}_2(0, \mathbf{I}), \quad (1 \leq j \leq 2),\\
    &\quad \gamma_{2,j,l,0} \sim \mathcal{N}(0, 0.5^2), \quad \gamma_{2,j,l,1} \sim \mathcal{N}(0, 1), \quad (1 \leq j \leq 3, \;\; 1 \leq l \leq 4). \\
&\text{At k = 3 to K}\\
    &\quad \alpha_{k, j, 0} = \alpha_{1, j, 0} + 0.1\,(k-1)/(K-1), \quad 
    \alpha_{k, \cdot, 1} = 0.25 + 0.10 \,(k-1)/K, \quad (1 \leq j \leq 2),\\
    &\quad \beta_{k,j,0} = 0, \quad  \beta_{k, j, 1} = \sqrt{0.8} \beta_{k-1,j,1} + \sqrt{0.2} \epsilon_\beta, \quad \epsilon_\beta \sim \mathcal{N}(0, 1), \quad (1 \leq j \leq 2), \\
    &\quad \gamma_{k,j,l,0} = \sqrt{0.8} \gamma_{k-1,j,l,0} + \sqrt{0.2} \epsilon_{\gamma_0}, \quad \epsilon_{\gamma_0} \sim \mathcal{N}(0, 0.5^2), \\
    &\quad \gamma_{k,j,l,1} = \sqrt{0.8} \gamma_{k-1,j,l,1} + \sqrt{0.2} \epsilon_{\gamma_1}, \quad \epsilon_{\gamma_1} \sim \mathcal{N}(0, 1) \quad (1 \leq j \leq 3, \;\; 1 \leq l \leq 4).\\
&\text{At k = K + 1}\\
    &\quad \gamma_{k,j,l,0} = \sqrt{0.8} \gamma_{K,j,l,0} + \sqrt{0.2} \epsilon_{\gamma_0}, \quad \epsilon_{\gamma_0} \sim \mathcal{N}(0, 0.5^2), \\
    &\quad \gamma_{k,j,l,1} = \sqrt{0.8} \gamma_{K,j,l,1} + \sqrt{0.2} \epsilon_{\gamma_1}, \quad \epsilon_{\gamma_1} \sim \mathcal{N}(0, 1) \quad (1 \leq j \leq 3, \;\; 1 \leq l \leq 4).
\end{align*}
\normalsize

\subsection{Latent Model Estimation for the BEST Study}
\label{append:modelestimation}
\begin{figure}[ht]
\centering
\includegraphics[width=0.8\linewidth]{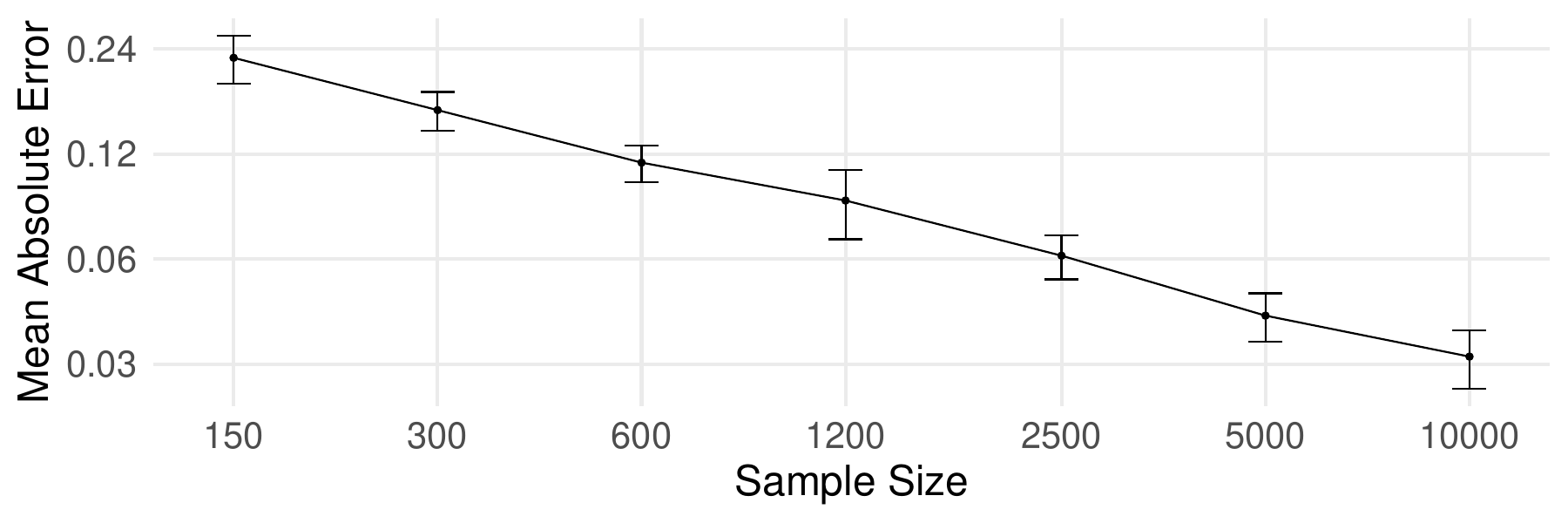}
\caption{Average with standard error bars of the mean absolute error $||\hat\theta_n -\theta_0||_1/\dim(\hat\theta_n)$ for our fitted model $\hat\theta_n$ across sample sizes. }
\label{fig:best_mae}
\end{figure}

Here we provide further details regarding estimation of $\theta$ for the preference model $P(\mathbf{H}_{K+1}|\mathbf{V})$ under the setting tailored towards the BEST study introduced in Section 5.1 in the manuscript. Denote our parametric model $P(\mathbf{H}_{K+1}|\mathbf{V},\theta)$. We calculate $\hat\theta_n $ and plot the mean absolute error $\dim(\theta)^{-1}||\hat\theta_n-\theta_0||_1$ for varying sample sizes in Figure \ref{fig:best_mae}. We can see that error declines with sample size at an approximately linear rate, verifying the results given in Lemma~\ref{lemma_theta}. We also note that the identifiability assumption made in Lemma~\ref{lemma_theta} is a necessary condition for consistency. Our algorithm converges to values close to the true parameter vector for large sample sizes across multiple seeds, supporting the identifiability of our proposed model. 

\begin{SCfigure}[][h]
\centering
\includegraphics[width=0.55\linewidth]{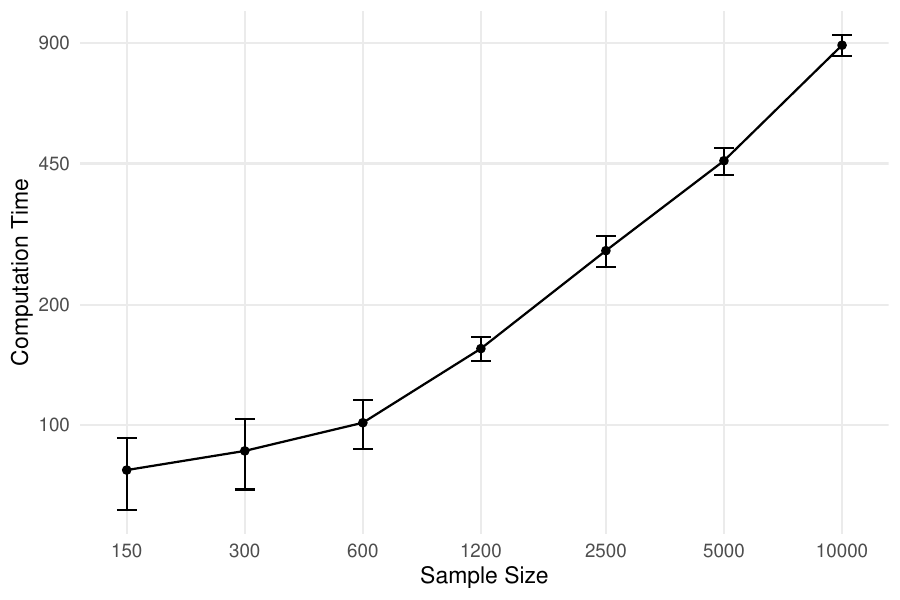}
\caption{Mean computation time (seconds) across 10 seeds is shown with standard error bars for various sample sizes. Optimization was performed with a single Tesla V100-SXM2 GPU, five 2.40 GHz Intel CPU cores, and 10GB of RAM.}
\label{figure:best_comp}
\vspace{-0.4cm}
\end{SCfigure}

Our optimization algorithm also performed well. Across sample sizes and seeds, we consistently observed $\log P(\widehat\theta|\mathcal D)\geq \log P(\theta_0|\mathcal D)$ and $||\nabla_\theta\log P(\widehat\theta|\mathcal D)||_{L^\infty(P_{\theta_0})}<10^{-7}$, indicating high convergence quality. Computation times for model fitting with varying sample sizes are reported in Figure \ref{figure:best_comp}. With $N=600$ simulated patients, which is a conservative estimate of the actual sample size for the BEST study, model-fitting took around $100$ seconds on average. Even with $10,000$ simulated patients, model fitting took under $900$ seconds on average. Computational performance can further be improved if needed by reducing the number of starting points and gradient descent iterations used as a warm-up for L-BFGS. These results demonstrate the efficiency and scalability of our optimization algorithm. 

While GPU computing and TensorFlow are more common in deep learning contexts, they provide practical advantages here. Automatic differentiation enables efficient computation of $\nabla_\theta \log P(\theta \mid \mathcal{D})$ without manual derivation of gradients, and GPU acceleration substantially reduces wall-clock time for the Monte Carlo integration required at each optimization step. Standard CPU-based alternatives such as adaptive Gauss-Hermite quadrature or EM algorithms \citep{Givens2012, GLIMMIX2018, Butler2018} would scale less favorably with $n$ and $|\theta|$ in our setting.

\sidecaptionvpos{figure}{c}

\subsection{Misspecification of $P(\mathbf{E})$}
\label{append:modelmisspec}
This subsection provides additional simulation result when $P(\bE)$ is mis-specified under the data-generating process designed for BEST described in Section 5.1 of the manuscript.

We generate the true latent preference as $\bE \sim Dirichlet(\alpha = c(1,1,1))$, with $\tilde{\bV} = \bE$; but we estimate $\theta$ assuming $\bV \sim \mathcal{N}_2(0, \mathbf{I})$, $\tilde{\bV} = (0, \bV)$, $\bE = SoftMax(\tilde{\bV})$. The evaluation data is again an independent data generation based on the truth. The estimation method, modeling choice for the Q functions, and baseline comparators remain the same as those described in the manuscript. Table (\ref{apptable:best_Eerr}) summarizes the estimated conditional preference when $P(\bE)$ is specified correctly and when it is not in various sample sizes. The sample sizes displayed are both the training and evaluation sample sizes. Results are summarized over 400 randomization seeds.

\begin{table}[H]
\centering
\captionsetup{margin=1.5cm}
\caption{Mean (SD) of $10 \times \mathrm{MAE}\!\left(\mathbb{E}[\mathbf{E}\mid\mathbf{H}_2;\theta_0]-\mathbb{E}[\mathbf{E}\mid\mathbf{H}_2;\hat\theta_n]\right)$ across Sample Sizes.}
\label{apptable:best_Eerr}
\begin{tabular}{lccccc}
\hline
 & N = 300 & N = 600 & N = 1200 & N = 2400 & N = 4800 \\
\hline
Correct $P(\mathbf{E})$ 
& 1.778 (0.31) 
& 1.258 (0.22) 
& 0.920 (0.15) 
& 0.669 (0.11) 
& 0.499 (0.07) \\

Mis-specified $P(\mathbf{E})$ 
& 6.743 (0.95) 
& 6.508 (0.96) 
& 6.372 (0.94) 
& 6.310 (0.94) 
& 6.267 (0.93) \\
\hline
\end{tabular}
\end{table}

We can see that, when the models for $P(\bW|\bX, \bE)$ are held to be the same, the mean absolute error (MAE) of $\mathbb{E}[\mathbf{E}|\mathbf{H}_2; \theta_0]- {\mathbb{E}}[\mathbf{E}|\mathbf{H}_2;\hat\theta_n]$ is consistently larger when $P(\bE)$ is mis-specified, and decrease more slowly with the increase of sample size compared to if correctly specified.

The effect of model mis-specification on the resulting value of the estimated DTRs can be seen by comparing Table~\ref{apptable:best_n_correct} and Table~\ref{apptable:best_n_mis}.
Table~\ref{apptable:best_n_correct} is the same as Table 1 in the main text, but is included here for ease of comparison.

Across both correctly specified and mis-specified settings, the performance ordering is largely preserved,
\[
\hat\pi_{\text{Known}} \succ
\hat\pi_{\text{LUQL}} \succ
\hat\pi_{\text{Naive}} \succ
\hat\pi_{\text{Wlast}},
\]
with the oracle benchmark $\hat\pi_{\text{Known}}$ uniformly achieving the highest value and LUQL consistently ranking second. Model mis-specification increases separation among methods. In particular, the performance gap between $\hat\pi_{\text{Known}}$ and $\hat\pi_{\text{LUQL}}$ widens under mis-specification, reflecting the additional estimation difficulty introduced by an incorrect working model. More substantially, the gap between $\hat\pi_{\text{LUQL}}$ and
$\hat\pi_{\text{Wlast}}$ becomes markedly larger when the model is mis-specified.

Although performance differences decrease with increasing sample size in both settings—consistent with variance reduction—the rate of contraction differs. Under correct specification, the gaps narrow relatively quickly and $\hat\pi_{\text{Wlast}}$ becomes competitive at
the largest sample size. In contrast, under mis-specification, the absolute gaps shrink more gradually and remain non-negligible even at the largest $N$ considered.

Together, the results suggest that (1) model selection via likelihood checks is important, and (2) LUQ-Learning remains preferable to naive and last-reported satisfaction approaches even under misspecification.

\begin{table}[H]
\centering
\captionsetup{margin=2cm}
\caption{Mean (SD) of $V(\hat{\pi}) - V(\pi_{obs})$ across Sample Sizes, $P(\mathbf{E})$ Correctly Specified.}
\label{apptable:best_n_correct}
\begin{tabular}{lccccc}
\hline
DTR & N = 300 & N = 600 & N = 1200 & N = 2400 & N = 4800 \\
\hline
$\hat\pi_{\text{Known}}$ 
& 0.169 (0.15) 
& 0.190 (0.13) 
& 0.202 (0.12) 
& 0.225 (0.13) 
& 0.303 (0.11) \\

$\hat\pi_{\text{LUQL}}$ 
& 0.155 (0.14) 
& 0.177 (0.13) 
& 0.189 (0.12) 
& 0.213 (0.12) 
& 0.288 (0.11) \\

$\hat\pi_{\text{Wlast}}$ 
& 0.076 (0.15) 
& 0.116 (0.14) 
& 0.153 (0.15) 
& 0.203 (0.16) 
& 0.341 (0.18) \\

$\hat\pi_{\text{Naive}}$ 
& 0.148 (0.14) 
& 0.171 (0.12) 
& 0.180 (0.12) 
& 0.203 (0.12) 
& 0.277 (0.11) \\
\hline
\end{tabular}
\end{table}

\begin{table}[H]
\centering
\captionsetup{margin=2cm}
\caption{Mean (SD) of $V(\hat{\pi}) - V(\pi_{obs})$ across Sample Sizes, $P(\mathbf{E})$ Mis-specified.}
\label{apptable:best_n_mis}
\begin{tabular}{lccccc}
\hline
DTR & N = 300 & N = 600 & N = 1200 & N = 2400 & N = 4800 \\
\hline
$\hat\pi_{\text{Known}}$ 
& 0.318 (0.19) 
& 0.376 (0.18) 
& 0.418 (0.18) 
& 0.457 (0.19) 
& 0.483 (0.18) \\

$\hat\pi_{\text{LUQL}}$ 
& 0.295 (0.19) 
& 0.343 (0.18) 
& 0.396 (0.18) 
& 0.426 (0.18) 
& 0.451 (0.17) \\

$\hat\pi_{\text{Wlast}}$ 
& 0.071 (0.14) 
& 0.108 (0.14) 
& 0.173 (0.15) 
& 0.236 (0.16) 
& 0.314 (0.18) \\

$\hat\pi_{\text{Naive}}$ 
& 0.283 (0.18) 
& 0.337 (0.18) 
& 0.382 (0.17) 
& 0.414 (0.18) 
& 0.440 (0.17) \\
\hline
\end{tabular}
\end{table}

\section{Application to the CATIE Trial}
\label{sec:catie}

To demonstrate the broad applicability of LUQ-Learning, we apply our method to the setting considered by \citet{Butler2018}. Their simulation setting was inspired by the first phase of the Clinical Antipsychotic Trials of Intervention Effectiveness (CATIE) trial \citep{Stroup2003}. Focusing on the first phase, this becomes a single decision point problem, with the data trajectory summarized as $(\textbf{X}_1, \textbf{W}_1, A_1, \textbf{Y}, \textbf{W}_2)$. The authors dichotomize the five treatment options into traditional and atypical antipsychotics, resulting in $\mathcal{A}_{1} = \{0, 1\}$. $\textbf{Y} \in \mathbb{R}^2$ comprises two continuous outcomes: efficacy measured using the Positive and Negative Syndromes Scale (PANSS) \citep{kay1987positive} and side effect burden measured as the sum of side effects and adverse events. $\mathbf{W}_1 \in \{0, 1\}^{10}$ are 10 Yes/No questions from the Drug Attitude Inventory \citep{Hogan1983}.  We adopt the same generative model as in \cite{Butler2018} with the addition of a log-linear model for $\mathbf{W}_2$, the reported treatment satisfaction collected at the end of study.
\footnotesize
\begin{align*}
& V\sim \mathcal{N}(0,1), \\
& \mathbf{E}=(\Phi(V), 1-\Phi(V)), \\
& \textbf{X}_1 \sim \mathcal{N}_{5}(0,\mathbf{I}), \\
& {W}_{1,j} \sim \text{Bernoulli}(p = \sigma(\beta_{j,0}+\beta_{j,1}V)), \quad (1 \leq j \leq 10), \\
& A_1 \sim \text{Bernoulli}(p = 0.5), \\
& Y_j = \mathbf{X}_{1*}^\top\gamma_{j,0} + A\mathbf{X}_{1*}^\top\gamma_{j,1}+\epsilon_j, \text{ where } \epsilon_j \sim \mathcal{N}(0,1), \quad (1 \leq j \leq 2), \\
& \mathbf{W}_2 \sim \text{Pois}\left(\lambda=\exp(\alpha_0+\alpha_1 \mathbf{E}^\top \mathbf{Y})\right).
\end{align*}
\normalsize
Here $\mathbf{X}_{1*}=(\mathbf{1},\mathbf{X}_1)$ and $\Phi(\cdot)$ is the standard normal cumulative distribution function. The parameters are generated as follows. $\gamma=(\gamma_{ij})_{i,j=1}^{2}$ was fixed as in \citet{Butler2018} to make outcomes $Y_1$ and $Y_2$ in a competing relationship. Here, $\theta = (\alpha, \beta)$, where $\alpha = (\alpha_0, \alpha_1), \beta = (\beta_{j,0}, \beta_{j,1})_{j=1}^{10}$, so $\mathrm{Card}(\theta) = 22$.
\footnotesize
\begin{align*}
    &\beta_{j,0}=0, \quad \beta_{j,1}\sim \mathcal{N}(0,1) \quad (1 \leq j \leq 10) \\
    &\alpha_0= - \alpha_1\min_{i}(\mathbf{E}_{i}^\top \mathbf{Y}_{i}) - 3, \quad \alpha_1= 6 /(\max_{i}(\mathbf{E}_{i}^\top \mathbf{Y}_{i})-\min_i(\mathbf{E}_{i}^\top \mathbf{Y}_{i}))\\
    &\gamma_{1,0} = (2.5, 0.2, 0.25, -0.7, -2.5, 2.4), \quad
    \gamma_{1,1} = (1.7, -2.3, 4.5, 6, -7.3, -1.6)\\
    &\gamma_{2,0} = 3 - 2\gamma_{1,0}, \quad \gamma_{2,1} = 3 - 2\gamma_{1,1}
\end{align*}
\normalsize

\begin{table}[H]
\centering
\captionsetup{margin=1.5cm}
\caption{Mean (SD) of $\text{MAE}(\hat{\mathbb{E}}[\mathbf{E}\mid\mathbf{H}_1;\hat\theta_n]- \hat{\mathbb{E}}[\mathbf{E}\mid\mathbf{H}_1;\theta_0])$ by Sample Sizes between LUQ-Learning and Butler's approach for Single Stage DTR, Summarized across 400 Seeds.}
\label{table:catie_Eerr}
\begin{tabular}{lcccc}
\hline
 & N = 100 & N = 200 & N = 500 & N = 1500 \\
\hline
LUQ-Learning 
& 0.050 (0.03) 
& 0.036 (0.03) 
& 0.026 (0.04) 
& 0.017 (0.04) \\

Butler's Method 
& 0.214 (0.17) 
& 0.209 (0.18) 
& 0.200 (0.19) 
& 0.196 (0.19) \\
\hline
\end{tabular}
\end{table}

\begin{table}[H]
\centering
\captionsetup{margin=2cm}
\caption{Mean (SD) of $V(\hat{\pi}) - V(\pi_{obs})$ by Sample Sizes, Summarized across 400 Seeds.}
\label{table:catie_n}
\begin{tabular}{lcccc}
\hline
DTR & N = 100 & N = 200 & N = 500 & N = 1500 \\
\hline
$\hat\pi_{\text{Known}}$ 
& 3.236 (1.06) 
& 3.830 (0.80) 
& 4.133 (0.48) 
& 4.324 (0.26) \\

$\hat\pi_{\text{LUQL}}$ 
& 2.813 (1.16) 
& 3.266 (0.84) 
& 3.473 (0.59) 
& 3.641 (0.46) \\

$\hat\pi_{\text{Butler}}$ 
& 2.074 (1.45) 
& 2.374 (1.32) 
& 2.369 (1.36) 
& 2.282 (1.52) \\

$\hat\pi_{\text{Wlast}}$ 
& 1.801 (1.36) 
& 2.193 (1.02) 
& 2.435 (0.63) 
& 2.628 (0.34) \\

$\hat\pi_{\text{Naive}}$ 
& 2.813 (1.06) 
& 3.024 (0.77) 
& 3.040 (0.48) 
& 3.072 (0.26) \\
\hline
\end{tabular}
\end{table}

We assume a correctly specified model for $P(\mathbf{W}_k|\mathbf{X}_k, \mathbf{E})$, $k = 1, 2$, and $P(\mathbf{W}_{K+1}|\mathbf{Y}, \mathbf{E})$. We estimated the preference model parameters $\theta = (\alpha, \beta)$ by maximizing the data log posterior and fitted the Q-functions using Random Forests with hyperparameters selected as described in Section \ref{sec:computation_details} above. Since the simulated datasets contain a single decision point and two outcomes, the methodology of \citet{Butler2018} is also applicable. In this setting, Butler's method reduces to LUQ-Learning with $\mathbf{W}_2$ excluded from the likelihood and an EM algorithm for parameter estimation. We consider sample sizes around $N = 200$, approximating the actual CATIE trial size. The policies $\hat\pi_{Wlast}$, $\hat\pi_{Naive}$, and $\hat\pi_{Known}$ are defined same as before.

Table \ref{table:catie_n} summarizes the results. LUQ-Learning yields more accurate estimates of expected preference weights (Table \ref{table:catie_Eerr}), leading to superior estimated DTR performance. In contrast, Butler's method exhibits high estimation error, making $\hat{\pi}_{Butler}$ less robust to small sample sizes.